\documentclass[10pt]{article} % For LaTeX2e

\usepackage[preprint]{tmlr/tmlr}
\usepackage{amsmath,amsfonts,bm}

\def\figref#1{figure~\ref{#1}}
\def\secref#1{section~\ref{#1}}
\def\eqref#1{equation~\ref{#1}}
\def\1{\bm{1}}

\def\vb{{\bm{b}}}

\def\vf{{\bm{f}}}

\def\vh{{\bm{h}}}

\def\vx{{\bm{x}}}

\def\evx{{x}}

\def\mW{{\bm{W}}}

\DeclareMathAlphabet{\mathsfit}{\encodingdefault}{\sfdefault}{m}{sl}
\SetMathAlphabet{\mathsfit}{bold}{\encodingdefault}{\sfdefault}{bx}{n}

\def\gG{{\mathcal{G}}}

\def\gI{{\mathcal{I}}}

\def\sR{{\mathbb{R}}}

\newcommand{\Ls}{\mathcal{L}}

\newcommand{\normlone}{L^1}

\usepackage[backref=page]{hyperref}

\usepackage{tikz}
\usetikzlibrary{shapes.geometric, arrows, positioning}
\tikzstyle{startstop} = [rectangle, rounded corners, minimum width=3cm, minimum height=1cm,text centered, draw=black, fill=red!30]
\tikzstyle{io} = [trapezium, trapezium left angle=70, trapezium right angle=110, minimum width=3cm, minimum height=1cm, text centered, draw=black, fill=blue!30]
\tikzstyle{process} = [rectangle, minimum width=3cm, minimum height=1cm, text centered, draw=black, fill=orange!30]
\tikzstyle{decision} = [diamond, minimum width=3cm, minimum height=1cm, text centered, draw=black, fill=green!30]
\tikzstyle{arrow} = [thick,->,>=stealth]

\usepackage{url}
\usepackage{graphicx}
\usepackage[hypcap=false]{caption}
\usepackage{comment}
\usepackage[most]{tcolorbox}
\usepackage{amssymb}
\usepackage{amsmath}
\usepackage{amsthm}
\usepackage{enumitem}
\usepackage{cleveref}
\usepackage{glossaries}
\usepackage{booktabs}

\usepackage{xcolor}
\usepackage{ifthen}
\usepackage[normalem]{ulem}
\definecolor{myorange}{HTML}{FEAE03}
\definecolor{myturquois}{HTML}{01AB8F}
\definecolor{mypink}{HTML}{D31876}

\definecolor{brightred}{HTML}{E55347} % Darker Tomato
\definecolor{orange}{HTML}{FF8C00} % Dark Orange
\definecolor{yellowgreen}{HTML}{6B8E23} % Olive Drab, darker for better contrast
\definecolor{green}{HTML}{228B22} % Forest Green, darker and richer

\newboolean{show_todos}
\setboolean{show_todos}{false}

\newcommand{\keep}[1]{\ifthenelse{\boolean{show_todos}}{#1}{#1}}
\newcommand{\refactor}[1]{\ifthenelse{\boolean{show_todos}}{{\color{orange}#1}}{#1}}
\newcommand{\todo}[1]{\ifthenelse{\boolean{show_todos}}{{\color{purple}#1}}{}}
\newcommand{\expand}[1]{\ifthenelse{\boolean{show_todos}}{{\color{red}#1}}{#1}}
\newcommand{\shorten}[1]{\ifthenelse{\boolean{show_todos}}{{\color{pink}#1}}{#1}}

\hypersetup{
  colorlinks=true,
  linkcolor=myorange,
  citecolor=myturquois, 
  filecolor=mypink,
  urlcolor=myturquois
}
\newtcolorbox{greybox}[1][]{
	float,
	title=#1,
}

\newtcolorbox{bluebox}[1][]{
	float,
  	title=#1,
	colback=myturquois!5,
	colframe=myturquois
}
\newtcolorbox{pinkbox}[1][]{
	float,
  	title=#1,
	colback=mypink!5,
	colframe=mypink
}

\newenvironment{defbox}[1][\unskip]{%
  \refstepcounter{definition}%
  \begin{bluebox}[Definition \thedefinition: #1]%
}{%
  \end{bluebox}%
}

\newenvironment{hypbox}[1][\unskip]{%
  \refstepcounter{hypothesis}%
  \begin{orangebox}[Hypothesis \thehypothesis: #1]%
}{%
  \end{orangebox}%
}

\newtcolorbox{orangebox}[1][]{
	float,
  	title=#1,
  	colback=myorange!5,
  	colframe=myorange
}

\renewcommand{\secref}[1]{Section~\ref{#1}}
\renewcommand{\figref}[1]{Figure~\ref{#1}}

\usepackage{subcaption}
\title{Mechanistic Interpretability for AI Safety \\ A Review}

\author{\name Leonard Bereska \hspace{1cm} \name Efstratios Gavves \\
      \email \{leonard.bereska, egavves\}@uva.nl \\
      \addr University of Amsterdam 
      }

\def\openreview{\url{https://openreview.net/forum?id=XXXX}} % Insert correct link to OpenReview for camera-ready version

\newcommand{\term}[1]{\textit{\gls{#1}}}
\makeglossaries

\newglossaryentry{prediction orthogonality}{
name={prediction orthogonality},
description={A model whose objective is prediction can simulate agents who optimize toward any objectives with any degree of optimality \citep{janus_simulators_2022}.}
}

\newglossaryentry{linear representation}{
name={linear representation},
description={Features are directions in activation space, i.e., linear combinations of neurons.}
}

\newglossaryentry{motifs}{
name={motifs},
description={Repeating patterns that emerge across models and tasks, manifesting as circuits, features, or higher-level behaviors from component interactions. Examples include curve detectors, induction circuits, and branch specialization. Motifs reveal common structures and mechanisms underlying neural network intelligence.}
}

\newglossaryentry{internal world models}{
name={internal world models},
description={Internal causal environment models formed within neural networks, implicitly emerging as a by-product of prediction (e.g., in large language models).} 
}

\newglossaryentry{simulacra}{
name={simulacra},
description={The text outputs generated by a predictive model simulating the causal processes underlying text creation. These outputs simulate coherent and contextually relevant language, sometimes exhibiting agentic behaviors or goals despite the predictive model itself lacking genuine agency or intentionality. Simulacra can be either \textit{agentic}, mimicking intentional and persuasive language use, or \textit{non-agentic}, merely generating descriptive text without simulated goals or agency \citep{janus_simulators_2022, bereska_taming_2023}.}
}

\newglossaryentry{natural abstractions}{
name={natural abstractions},
description={High-level summaries or descriptions of a system or environment learned and used by many cognitive systems. According to the \textit{natural abstraction hypothesis} \citep{chan_natural_2023}, a set of "natural" abstractions exist that represent redundantly encoded information in the world and tend to be learned by intelligent systems produced through local selection pressures. These natural abstractions form a relatively small, discrete set of concepts like "tree," "velocity," etc., that allow compact descriptions of the world while discarding many irrelevant low-level details.}
}

\newglossaryentry{circuits}{
name={circuits},
description={Sub-graphs within neural networks consisting of \term{features} and the weights connecting them. Circuits can be thought of as \textit{computational primitives} that perform understandable operations to produce (ideally interpretable) features from prior (ideally interpretable) features. % An end-to-end circuit describes how a model's inputs are transformed into its outputs, ideally through a sequence of interpretable intermediate computations. 
Examples include circuits for detecting curves at specific orientations \citep{cammarata_curve_2020, cammarata_curve_2021}, continuing repeated patterns in text \citep{olsson_incontext_2022}, and resolving anaphoric references \citep{wang_interpretability_2023}. While circuits can involve clearly interpretable features, the definition allows for intermediate representations that are less easily interpretable.}
}

\newglossaryentry{simulation}{
name={simulation},
description={The simulation hypothesis says that when scaled up sufficiently, predictive models will learn to simulate the real-world causal processes that generated their training data \citep{janus_simulators_2022}. When these models are optimized for predictive accuracy on broad data distributions like natural language, they are incentivized to discover the underlying rules, physics, and semantics that govern the data to model and predict future observations effectively. This allows the models to go beyond just memorizing or pattern-matching their training sets, instead learning to simulate hypothetical scenarios, reason about counterfactuals, and exhibit behaviors characteristic of general intelligence -- all as a byproduct of the drive for efficient compression and accurate prediction. The simulation hypothesis suggests these models will develop rich \term{internal world models} capturing the causal dynamics of the training distribution.}
}

\newglossaryentry{features}{
name={features},
description={The fundamental units of how neural networks encode knowledge, which cannot be further decomposed into smaller, distinct \term{concepts}. Features are core components of a neural network's representation, analogous to how cells form the fundamental unit of biological organisms \citep{olah_zoom_2020}. The \term{superposition} hypothesis suggests an alternative definition: that features correspond to the \term{disentangled} concepts that a larger, sparser network with \textit{sufficient capacity} would learn to represent with individual (\term{monosemantic}) neurons \citep{olah_zoom_2020, bricken_monosemanticity_2023}.}
}

\newglossaryentry{concepts}{
name={concepts},
description={An abstract idea or representation derived from observations of the world. Concepts refer to the \term{natural abstractions} that a cognitive system, like a neural network, aims to capture and represent through its learned \term{features}, which may or may not align perfectly with human-defined concepts.}
}

\newglossaryentry{disentangled}{
name={disentangled},
description={In disentangled representations, individual dimensions or components correspond to distinct, \textit{independent factors of variation in the data}, rather than representing a tangled mixture of these factors.}
}

\newglossaryentry{monosemantic}{
name={monosemantic},
description={A neuron corresponding to a single concept. The intuition is that analyzing what inputs activate a given neuron reveals its associated semantic meaning or concept. In contrast to \term{polysemantic}.}
}

\newglossaryentry{polysemantic}{
name={polysemantic},
description={Neurons that are associated with multiple, unrelated \term{concepts}, contradicting the interpretation of neurons as representational primitives and making it challenging to understand the information processing of neural networks. This term is derived from linguistic concepts of \textit{polysemy} \citep{falkum_polysemy_2015}, and in the context of neural networks first introduced by \citet{arora_linear_2018}, who suggested that word embeddings of polysemous words may be stored as a \term{superposition} of vectors representing distinct meanings. \citet{olah_zoom_2020} first used the term \textit{polysemanticity}, elaborating on the concept of \textit{polysemantic} neurons as a challenge for mechanistic interpretability.}
}

\newglossaryentry{superposition}{
name={superposition},
description={The superposition hypothesis suggests that neural networks can leverage high-dimensional spaces to represent more \term{features} than the actual count of neurons by encoding features in almost orthogonal directions \citep{elhage_toy_2022}.}
}

\newglossaryentry{modularity}{
name={modularity},
description={The property of an AI system being composed of distinct, semi-independent components or submodules that can be separately understood, modified, and recombined, rather than a monolithic, opaque structure.}
}

\newglossaryentry{universality}{
name={universality},
description={The universality hypothesis proposes the emergence of common \term{circuits} across neural network models trained on similar tasks and data distributions. 
A \textit{stronger} form posits that these common circuits represent a set of fundamental computational \term{motifs} that neural networks gravitate towards when learning. 
The \textit{weaker} version suggests that for a given task, dataset, and model architecture, an optimal way to solve the problem may exist, which different models will tend to converge towards, resulting in analogous circuits. 
The universality hypothesis implies that rather than each model learning arbitrary, unstructured representations, there is an underlying universality to the circuits that emerge, shaped by the learning task and inductive biases.}
}

\newglossaryentry{representation engineering}{
name={representation engineering},
description={A top-down approach to transparency research that treats representations as the fundamental unit of analysis, aiming to understand and control representations of high-level cognitive phenomena in neural networks like large language models. Representation engineering has two main areas: 1) Reading representations to probe and interpret their contents, and 2) Controlling representations to manipulate high-level concepts like honesty or morality \citep{zou_representation_2023}.}
}

\newglossaryentry{privileged basis}{
name={privileged basis},
description={In certain neural network representations, the basis directions formed by the individual neurons are architecturally distinguished from arbitrary directions in the activation space. This privileged basis makes it meaningful to analyze the properties and roles of individual neurons, as the architecture encourages features to align with these basis directions. Hence, a privileged basis is \textit{necessary} but \textit{not sufficient} for the formation of \term{monosemantic} neurons. \citep{elhage_toy_2022}.}
}

\newglossaryentry{streetlight interpretability}{
name={streetlight interpretability},
description={Examining AI systems under only ideal conditions of maximal interpretability, risking missing critical phenomena that only emerge in more realistic and diverse contexts.}
}

\newglossaryentry{machine unlearning}{
name={machine unlearning},
description={Techniques for removing private data or dangerous knowledge from models.}
}

\newglossaryentry{reverse engineering}{
name={reverse engineering},
description={The process of deconstructing a neural network’s computations to fully understand and specify its operations. This involves breaking down the network’s functionality into explicit, interpretable components, potentially as clear and detailed as pseudocode.}
}

\newglossaryentry{outer misalignment}{
name={outer misalignment},
description={Outer misalignment, or reward hacking, occurs when the specified reward function or utility function fails to capture the desired objectives correctly. This leads the AI to optimize for behaviors that achieve high reward scores but are misaligned with the intended outcomes.}
}

\newglossaryentry{reward hacking}{
name={reward hacking},
description={See \term{outer misalignment}.}
}

\newglossaryentry{inner misalignment}{
name={inner misalignment},
description={Inner misalignment, or goal misgeneralization, occurs when an AI system develops goals or behaviors during training that are misaligned with the intended objectives despite a correctly specified reward signal.}
}

\newglossaryentry{mesa-optimization}{
name={mesa-optimization},
description={The emergence of unintended subagents within a model with their own objectives, potentially misaligned with the original training objective.}
}

\newglossaryentry{eliciting latent knowledge}{
name={eliciting latent knowledge},
description={Developing strategies to make a machine learning model explicitly report latent facts or knowledge embedded in its parameters, especially in cases where the model's output is untrusted \citep{christiano_eliciting_2021}. This involves finding patterns in neural network activations that track the true state of the world \citep{mallen_eliciting_2023}.}
}

\newglossaryentry{well-founded AI}{
name={well-founded AI},
description={Developing AI systems with provable safety guarantees about their behavior and alignment with human values through rigorous mathematical modeling and verification. \citep{tegmark_provably_2023, dalrymple_guaranteed_2024}.}
}

\newglossaryentry{microscope AI}{
name={microscope AI},
description={Systems that extract and utilize knowledge from a model without allowing the model to take autonomous actions. This involves reverse engineering a trained model to understand its learned knowledge about the world, aiming to leverage this understanding directly without deploying the model in an operational capacity.}
}

\newglossaryentry{grokking}{
name={grokking},
description={"Grokking refers to the surprising phenomenon of delayed generalization where neural networks, on certain learning problems, generalize long after overfitting their training set." \citep{liu_understanding_2022}}
}

\newglossaryentry{hydra effect}{
name={hydra effect},
description={The phenomenon where models can internally self-repair and maintain capabilities even when key components are ablated, making it challenging to identify the relevant components underlying a particular behavior \citep{mcgrath_hydra_2023}.}
}

\newglossaryentry{oversight}{
name={oversight},
description={(Scalable) oversight refers to the challenge of providing reliable supervision—through labels, reward signals, or critiques—to AI models, ensuring effectiveness even as models \textit{surpass} human-level performance.}
}

\newglossaryentry{iterative distillation and amplification}{
name={iterative distillation and amplification},
description={A technique for training AI systems by repeatedly distilling knowledge from a larger model into a smaller one while amplifying the smaller model's capabilities through feedback and interaction with humans.}
}

\newglossaryentry{deceptive alignment}{
name={deceptive alignment},
description={When a misaligned model aims to appear aligned to gain more power to take control once sufficiently powerful.}
}

\newglossaryentry{deceptive inflation}{
name={deceptive inflation},
description={Theoretical result on deceptive behavior: policies produce trajectories that look better than they actually are from the human's perspective with limited observations to get higher reward signals during training. This deceptive behavior arises in reinforcement learning from human feedback when the human provides feedback based only on partial observations of the trajectories, while the policy has full state information during training \citep{lang_when_2024}.}
}

\newglossaryentry{sycophancy}{
name={sycophancy},
description={The tendency of models to generate responses that align with user beliefs rather than providing truthful information. This behavior, encouraged by human feedback used in fine-tuning, is observed in state-of-the-art AI assistants across various tasks \citep{sharma_understanding_2023}. Sycophancy arises because human preference judgments often favor responses that match users' views, leading to a preference for convincingly written sycophantic responses over correct ones.}
}

\newglossaryentry{irreducible}{
name={irreducible},
description={We adopt the notion of \term{features} as the fundamental units of neural network representations, such that features cannot be further decomposed into smaller, distinct factors. 
To make this more precise, we can formalize the definition of features as irreducible input patterns following \citet{engels_not_2024}:
A feature $f$ of sparsity $s$ is a function that maps a subset of the input space (with probability $1-s > 0$) into a higher-dimensional representational space. We say the feature is active on this subset.
A feature $f$ is reducible into features $a$ and $b$ if there exists a transformation that decomposes $f$ into $a$ and $b$, such that the transformed distribution $p(a, b)$ is either:
\begin{enumerate}
\item Separable: $p(a, b) = p(a)p(b)$
\item A mixture: $p(a, b) = w p_1(a, b) + (1-w)p_2(a, b)$ where $p_1$ is lower-dimensional.
\end{enumerate}
Features are defined as irreducible patterns that cannot be decomposed into separable or mixture distributions via such transformations.
This formalizes the notion that features form the fundamental atomic units underlying neural representations. Features that can be \term{disentangled} into statistically independent components (separable) or simpler lower-dimensional factors (mixtures) are not considered the core representational primitives.
The key properties are that 1) features map from the input space to higher-dimensional representational spaces, 2) features are sparse and only activated on subsets of the input, and crucially, 3) features are irreducible and cannot be expressed as transformations of other statistically independent components.}
}

\newglossaryentry{structured probes}{
name={structured probes},
description={Advanced techniques in conceptual interpretability that aim to uncover complex features like truth representations in language models.}
}
\newglossaryentry{activation patching}{
name={activation patching},
description={}
}
\newglossaryentry{sparse autoencoders}{
name={sparse autoencoders},
description={}
}
\newglossaryentry{developmental interpretability}{
name={developmental interpretability},
description={A focus on learning dynamics, aiming to understand the incremental development of internal structure in neural networks, one phase transition at a time.}
}
\newglossaryentry{intrinsic interpretability}{
name={intrinsic interpretability},
description={Methods that aim to design neural networks more amenable to reverse engineering through architectural choices and training procedures that encourage sparsity, modularity, and monosemanticity.}
}
\newglossaryentry{post-hoc interpretability}{
name={post-hoc interpretability},
description={Techniques applied to trained models to gain insights into their behavior and decision-making processes.}
}
\newglossaryentry{adversarial robustness}{
name={adversarial robustness},
description={A property of models resistant to adversarial attacks, where small, carefully crafted perturbations to the input can significantly change the model's output.}
}

\newglossaryentry{neurosymbolic reasoning}{
name={neurosymbolic reasoning},
description={Approaches that combine neural networks with symbolic reasoning, leveraging the strengths of both paradigms to create more interpretable and compositional AI systems.}
}
\newglossaryentry{program synthesis}{
name={program synthesis},
description={The automatic generation of executable programs from high-level specifications or examples, with applications in interpretability for \term{reverse-engineering} the algorithms learned by neural networks.}
}
\newglossaryentry{trojan detection}{
name={trojan detection},
description={The task of identifying and removing malicious backdoors or "trojans" that have been intentionally inserted into a model, often via data poisoning.}
}
\newglossaryentry{causal scrubbing}{
name={causal scrubbing},
description={A rigorous method to formalize and test hypotheses about how neural networks implement specific behaviors by replacing activations in the model's computational graph with equivalent activations according to the hypothesis.}
}
\newglossaryentry{causal abstraction}{
name={causal abstraction},
description={A mathematical framework that treats both neural networks and potential explanations as causal models, allowing the validity of an explanation to be empirically tested through interchange interventions.}
}
\newglossaryentry{locally consistent abstractions}{
name={locally consistent abstractions},
description={A more permissive notion of explaining neural network behavior, where the consistency between the neural network and the explanation is checked only one step away from the intervention node.}
}

\newglossaryentry{MLP-In-The-Middle illusion}{
name={MLP-In-The-Middle illusion},
description={A phenomenon where patching an entire Multi-Layer Perceptron (MLP) layer shows no observable effect, yet patching a specific subspace within the same layer reveals significant impacts, raising questions about the relevance of certain subspaces in the model's normal functioning.}
}
\newglossaryentry{attribution patching}{
name={attribution patching},
description={A gradient-based alternative to traditional activation patching, which takes a linear approximation to provide a faster and more scalable approach to probing neural network behaviors.}
}
\newglossaryentry{path patching}{
name={path patching},
description={A variation of activation patching that quantitatively tests hypotheses expressing that behaviors are localized to a set of paths in the neural network.}
}
\newglossaryentry{attention pattern patching}{
name={attention pattern patching},
description={A method that leverages attention attribution patterns to gain insights into the information flow within a neural network.}
}
\newglossaryentry{causal inference}{
name={causal inference},
description={A set of techniques and principles for understanding cause-and-effect relationships, which can be applied to analyze the causal structure of neural networks.}
}
\newglossaryentry{singular learning theory}{
name={singular learning theory},
description={A mathematical framework for understanding the asymptotic behavior of learning algorithms in the presence of degeneracy, which can provide insights into the emergence and phase transitions of neural network representations.}
}

\newglossaryentry{AI alignment}{
name={AI alignment},
description={The goal of ensuring that the behavior of an AI system is aligned with human values and intentions, preventing unintended or harmful outcomes.}
}

\newglossaryentry{monosemanticity}{
name={monosemanticity},
description={The property of a neuron or feature in a neural network corresponding to a single, clearly interpretable semantic concept, rather than being associated with multiple unrelated concepts.}
}

\begin{document}

\maketitle

\ifthenelse{\boolean{show_todos}}{
	\section*{Color coding status of completion:}
\todo{This is a todo item.}
\expand{This needs additional information.}
\refactor{This should be refactored.}
\shorten{Consider deleting or shortening this.}
\keep{This text may be kept as is or need light editing.}
}{}

\begin{abstract}
Understanding AI systems' inner workings is critical for ensuring value alignment and safety. This review explores mechanistic interpretability: reverse engineering the computational mechanisms and representations learned by neural networks into human-understandable algorithms and concepts to provide a granular, causal understanding.
We establish foundational concepts such as features encoding knowledge within neural activations and hypotheses about their representation and computation. We survey methodologies for causally dissecting model behaviors and assess the relevance of mechanistic interpretability to AI safety. We examine benefits in understanding, control, alignment, and risks such as capability gains and dual-use concerns. 
We investigate challenges surrounding scalability, automation, and comprehensive interpretation. We advocate for clarifying concepts, setting standards, and scaling techniques to handle complex models and behaviors and expand to domains such as vision and reinforcement learning. Mechanistic interpretability could help prevent catastrophic outcomes as AI systems become more powerful and inscrutable. For an HTML version of the paper, visit \url{https://leonardbereska.github.io/blog/2024/mechinterpreview/}. 
\end{abstract}

\section{Introduction}\label{sec:introduction}

    As AI systems rapidly become more sophisticated and general \citep{bubeck_sparks_2023, bengio_managing_2023}, advancing our understanding of these systems is crucial to ensure their alignment \citep{ji_ai_2024} with human values and avoid catastrophic outcomes \citep{hendrycks_overview_2023, hendrycks_xrisk_2022}. The field of interpretability aims to demystify the internal processes of AI models, moving beyond evaluating performance alone. This review focuses on mechanistic interpretability, an emerging approach within the broader interpretability landscape that strives to comprehensively specify the computations underlying deep neural networks. We emphasize that understanding and interpreting these complex systems is not merely an academic endeavor -- \textit{it's a societal imperative to ensure AI remains trustworthy and beneficial}.

    The interpretability landscape is undergoing a paradigm shift akin to the evolution from behaviorism to cognitive neuroscience in psychology. Historically, lacking tools for introspection, psychology treated the mind as a black box, focusing solely on observable behaviors. Similarly, interpretability has predominantly relied on black-box techniques \citep{casper_blackbox_2024}, analyzing models based on input-output relationships or using attribution methods that, while probing deeper, still neglect the model's internal architecture. However, just as advancements in neuroscience allowed for a deeper understanding of internal cognitive processes, the field of interpretability is now moving towards a more granular approach. This shift from surface-level analysis to a focus on the internal mechanics of deep neural networks characterizes the transition towards inner interpretability \citep{rauker_transparent_2023}.

     Mechanistic interpretability, as an approach to inner interpretability, aims to completely specify a neural network's computation, potentially in a format as explicit as pseudocode (also called \term{reverse engineering}), striving for a granular and precise understanding of model behavior. It distinguishes itself primarily through its \textit{ambition} for comprehensive reverse engineering and its strong \textit{motivation} towards AI safety. Our review serves as the first comprehensive exploration of mechanistic interpretability research, with the most accessible introductions currently scattered in a blog or list format \citep{olah_mechanistic_2022, nanda_comprehensive_2022, olah_zoom_2020, sharkey_current_2022, olah_building_2018, nanda_mechanistic_2023, nanda_extremely_2024}. Concurrently, \citet{ferrando_primer_2024} and \citet{rai_practical_2024} have also contributed valuable reviews giving concise, technical introductions to mechanistic interpretability in transformer-based language models. Our work complements these efforts by synthesizing the research (addressing the "research debt" \citep{olah_research_2017}) and providing a structured, accessible, and comprehensive introduction for AI researchers and practitioners.

	The structure of this paper provides a cohesive overview of mechanistic interpretability, situating the mechanistic approach in the broader interpretability landscape (\secref{sec:comparison}), presenting core concepts and hypotheses (\secref{sec:concepts}), explaining methods and techniques (\secref{sec:methods}), presenting a taxonomy and survey of the current field (\secref{sec:survey}), exploring relevance to AI safety (\secref{sec:relevance}), and addressing challenges (\secref{sec:challenges}) and future directions (\secref{sec:future}).

\section{Interpretability Paradigms from the Outside In}\label{sec:comparison}

\begin{figure}
    \centering
    \includegraphics[width=\textwidth]{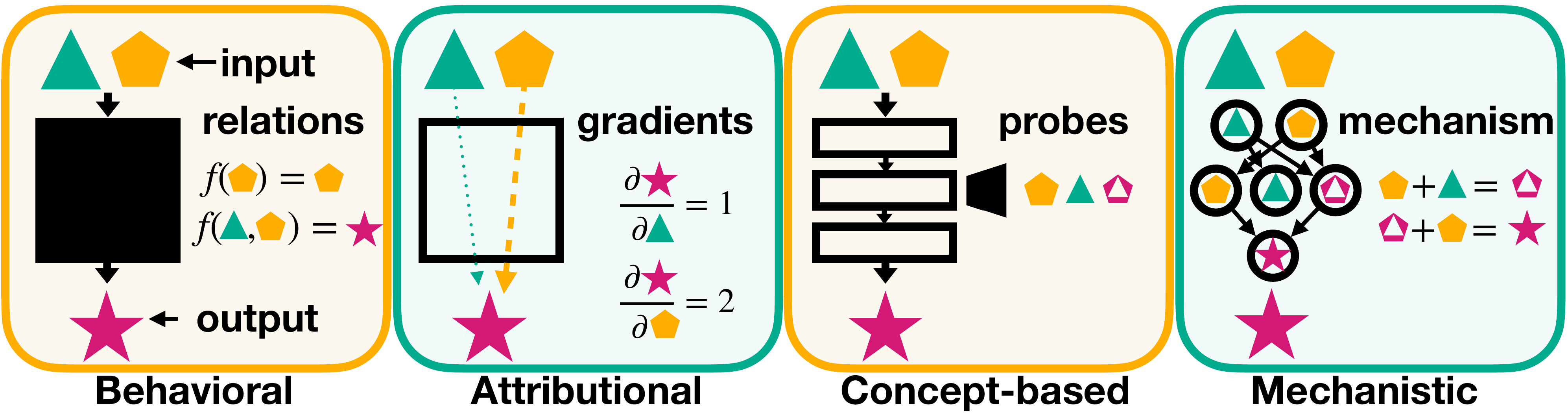}
    \caption{Interpretability paradigms offer distinct lenses for understanding neural networks: \textbf{Behavioral} analyzes input-output relations; \textbf{Attributional} quantifies individual input feature influences; \textbf{Concept-based} identifies high-level representations governing behavior; \textbf{Mechanistic} uncovers precise causal mechanisms from inputs to outputs.}
    \label{fig:paradigms}
    \end{figure}

    We encounter a spectrum of interpretability paradigms for decoding AI systems' decision-making, ranging from external black-box techniques to internal analyses. We contrast these paradigms with mechanistic interpretability, highlighting its distinct causal bottom-up perspective within the broader interpretability landscape (see \figref{fig:paradigms}).

    \paragraph{Behavioral} interpretability treats the model as a black box, analyzing input-output relations. Techniques such as minimal pair analysis \citep{warstadt_blimp_2020}, sensitivity and perturbation analysis \citep{casalicchio_visualizing_2018} examine input-output relations to assess the model's robustness and variable dependencies \citep{shapley_value_1988, ribeiro_why_2016, covert_explaining_2021}. Its \textit{model-agnostic} nature is practical for complex or proprietary models but lacks insight into internal decision processes and causal depth \citep{jumelet_evaluating_2023}.
    
    \paragraph{Attributional} interpretability aims to explain outputs by tracing predictions to individual input contributions using gradients. Raw gradients can be discontinuous or sensitive to slight perturbations. Therefore, techniques such as SmoothGrad \citep{smilkov_smoothgrad_2017} and Integrated Gradients \citep{sundararajan_axiomatic_2017} average across gradients. Other popular techniques are layer-wise relevance propagation \citep{bach_pixelwise_2015}, DeepLIFT \citep{shrikumar_learning_2017}, or GradCAM \citep{selvaraju_gradcam_2016}. Attribution enhances transparency by showing input feature influence without requiring an understanding of the internal structure, enabling decision validation, compliance, and trust while serving as a bias detection tool, but also has fundamental limitations \citep{bilodeau_impossibility_2024}.
    
    \paragraph{Concept-based} interpretability adopts a top-down approach to unraveling a model's decision-making processes by probing its learned representations for high-level concepts and patterns governing behavior. Techniques include training supervised auxiliary classifiers \citep{belinkov_probing_2021}, employing unsupervised contrastive and structured probes (see \secref{sec:methods:structured_probes}) to explore latent knowledge \citep{burns_discovering_2023}, and using neural representation analysis to quantify the representational similarities between the internal representations learned by different neural networks \citep{kornblith_similarity_2019, bansal_revisiting_2021}. 
    Beyond observational analysis, concept-based interpretability can enable manipulation of these representations -- also called \term{representation engineering} \citep{zou_representation_2023} -- potentially enhancing safety by upregulating concepts such as honesty, harmlessness, and morality.
    
    \paragraph{Mechanistic} interpretability is a bottom-up approach that studies the fundamental components of models through granular analysis of features, neurons, layers, and connections, offering an intimate view of operational mechanics. Unlike concept-based interpretability, it aims to uncover causal relationships and precise computations transforming inputs into outputs, often identifying specific neural circuits driving behavior. This \term{reverse engineering} approach draws from interdisciplinary fields like physics, neuroscience, and systems biology to guide the development of transparent, value-aligned AI systems. Mechanistic interpretability is the primary focus of this review.

\section{Core Concepts and Assumptions}\label{sec:concepts}

\begin{figure}
    \centering
    \includegraphics[width=1.\textwidth]{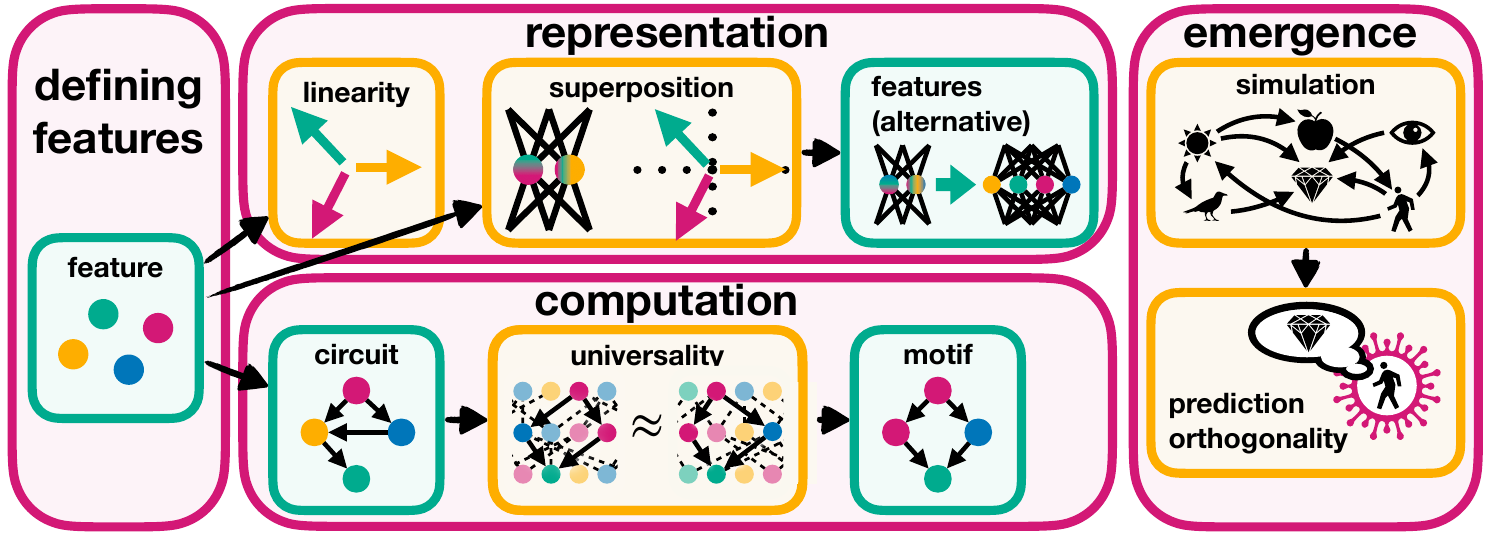}
    \caption{Overview of key concepts and hypotheses in mechanistic interpretability, organized into four subsection (pink boxes): defining features (\secref{sec:concepts:features}), representation (\secref{sec:concepts:representation}), computation (\secref{sec:concepts:computation}), and emergence (\secref{sec:concepts:emergence}). In turquoise, it highlights definitions like \term{features}, \term{circuits}, and \term{motifs}, and in orange, it highlights hypotheses like \term{linear representation}, \term{superposition}, \term{universality}, \term{simulation}, and \term{prediction orthogonality}. Arrows show relationships, e.g., superposition enabling an alternative feature definition or universality connecting circuits and motifs.}
  \label{fig:overview}
\end{figure}

This section introduces the key concepts and hypotheses of mechanistic interpretability, as summarized in \figref{fig:overview}. We start by defining features as the basic units of representation (\secref{sec:concepts:features}). We then examine the nature of these features, including the challenges posed by polysemantic neurons and the implications of the superposition and linear representation hypotheses (\secref{sec:concepts:representation}). Next, we explore computation through circuits and motifs, considering the universality hypothesis (\secref{sec:concepts:computation}). Finally, we discuss the implications for understanding emergent properties, such as internal world models and simulated agents with potentially misaligned objectives (\secref{sec:concepts:emergence}).

\subsection{Defining Features as Representational Primitives}\label{sec:concepts:features}

\paragraph{Features as fundamental units of representation.} The notion of a \emph{feature} in neural networks is central yet elusive, reflecting the pre-paradigmatic state of mechanistic interpretability. We adopt the notion of \term{features} as the \textit{fundamental units of neural network representations}, such that features cannot be further \term{disentangled} into simpler, distinct factors. These features are core components of a neural network's representation, analogous to how cells form the fundamental unit of biological organisms \citep{olah_zoom_2020}.

\begin{defbox}[Feature]
Features are the fundamental units of neural network representations that cannot be further decomposed into simpler independent factors.
\end{defbox}

\paragraph{Concepts as natural abstractions.} The world consists of various entities that can be grouped into categories or \term{concepts} based on shared properties. These concepts form high-level summaries like "tree" or "velocity," allowing compact world representations by discarding many irrelevant low-level details. Neural networks can capture and represent such \term{natural abstractions} \citep{chan_natural_2023} through their learned \term{features}, which serve as building blocks of their internal representations, aiming to capture the \term{concepts} underlying the data.

\paragraph{Features encoding input patterns.} In traditional machine learning, \term{features} are understood as characteristics or attributes derived directly from the input data stream \citep{bishop_pattern_2006}. This view is particularly relevant for systems focused on \textit{perception}, where features map closely to the input data. However, in more advanced systems capable of \textit{reasoning} with abstractions, features may emerge internally within the model as representational patterns, even when processing information unrelated to the input. In this context, features are better conceptualized as \emph{any measurable property or characteristic of a phenomenon} \citep{olah_mechanistic_2022}, encoding abstract {concepts} rather than strictly reflecting input attributes.

\paragraph{Features as representational atoms.} A key property of features is their irreducibility, meaning they cannot be decomposed into or expressed as a combination of simpler, independent factors. In the context of input-related features, \citet{engels_not_2024} define a feature as \term{irreducible} if it cannot be decomposed into or expressed as a combination of statistically independent patterns or factors in the original input data. Specifically, a feature is reducible if transformations reveal its underlying pattern, which can be separated into independent co-occurring patterns or is a mixture of patterns that never co-occur. We propose generalizing this notion of irreducibility to features encoding abstract concepts not directly tied to input patterns, such that features cannot be reduced to combinations or mixtures of other independent components within the model's representations.

\paragraph{Features beyond human interpretability.}
Features could be defined from a \emph{human-centric perspective} as \emph{semantically meaningful, articulable input patterns encoded in the network's activation space} \citep{olah_mechanistic_2022}. However, while cognitive systems may converge on similar \term{natural abstractions} \citep{chan_natural_2023}, these need not necessarily align with human-interpretable \term{concepts}.
Adversarial examples have been interpreted as non-interpretable features meaningful to models but not humans. Imperceptible perturbations fool networks, suggesting reliance on alien representational patterns \citep{ilyas_adversarial_2019}. As models surpass human capabilities, their learned features may become increasingly abstract, encoding information in ways incongruent with human intuition \citep{hubinger_chris_2019}.
Mechanistic interpretability aims to uncover the \emph{actual} representations learned, even if diverging from human concepts. While human-interpretable concepts provide guidance, a non-human-centric perspective that defines features as independent model components, whether aligned with human concepts or not, is a more comprehensive and future-proof approach.

\subsection{Nature of Features: From Monosemantic Neurons to Non-Linear Representations}\label{sec:concepts:representation}

\begin{figure}
    \centering
    \includegraphics[width=1.\textwidth]{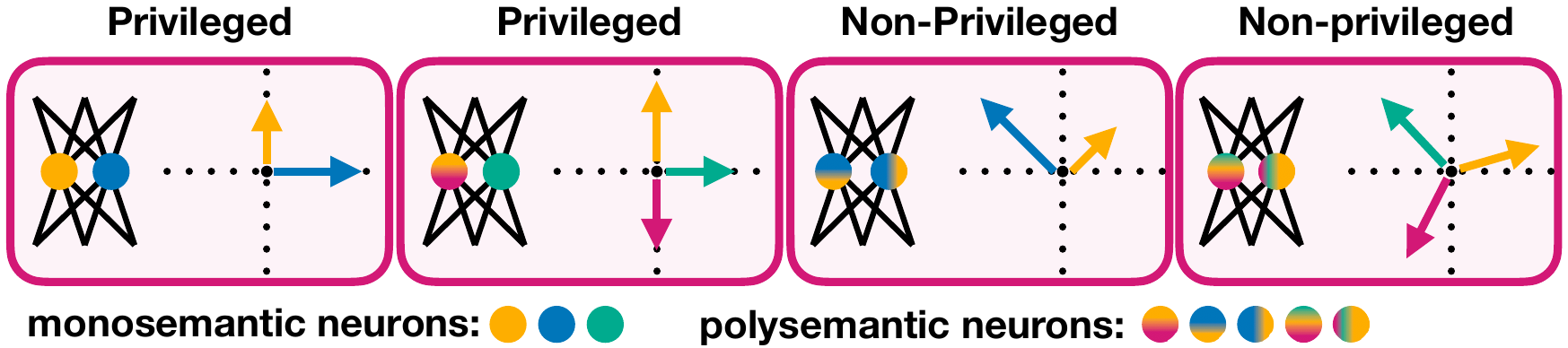}
    \caption{Contrasting privileged and non-privileged bases. In a non-privileged basis, there is no reason to expect features to be basis-aligned -- calling basis dimensions neurons has no meaning. In a privileged basis, the architecture treats basis directions differently -- features can but need not align with neurons \citep{bricken_monosemanticity_2023}. \textbf{Leftmost:} Privileged basis; individual features (arrows) align with basis directions, resulting in \term{monosemantic} neurons (colored circles). \textbf{Middle left:} Privileged basis, where despite having more features than neurons, some neurons are monosemantic, representing individual features, while others are \term{polysemantic} (overlapping gradients), encoding \term{superposition} of multiple features. \textbf{Middle right:} Non-privileged basis where, even when the number of features equals the number of neurons, the lack of alignment between the feature directions and basis directions results in polysemantic neurons encoding combinations of features. \textbf{Rightmost:} Non-privileged, polysemantic neurons as feature directions do not align with neuron basis.}
    \label{fig:privileged}
\end{figure}

\paragraph{Neurons as Computational Units?}
In the architecture of neural networks, neurons are the natural computational units, potentially representing individual features. Within a neural network representation $h\in \mathbb{R}^n$, the $n$ basis directions are called neurons. For a neuron to be meaningful, the basis directions must functionally differ from other directions in the representation, forming a \term{privileged basis} -- where the basis vectors are architecturally distinguished within the neural network layer from arbitrary directions in activation space, as shown in \figref{fig:privileged}. Typical non-linear activation functions privilege the basis directions formed by the neurons, making it meaningful to analyze individual neurons \citep{elhage_toy_2022}. Analyzing neurons can give insights into a network's functionality \citep{sajjad_neuronlevel_2022, mu_compositional_2020, dai_knowledge_2022, ghorbani_neuron_2020, voita_neurons_2023, durrani_analyzing_2020, goh_multimodal_2021, bills_language_2023, huang_rigorously_2023}.

\paragraph{Monosemantic and Polysemantic Neurons.}
A neuron corresponding to a single semantic concept is called \term{monosemantic}. The intuition behind this term comes from analyzing what inputs activate a given neuron, revealing its associated semantic meaning or concept. If neurons were the representational primitives of neural networks, all neurons would be monosemantic, implying a one-to-one relationship between neurons and features. Comprehensive interpretability would be as tractable as characterizing all neurons and their connections. However, empirically, especially for transformer models \citep{elhage_toy_2022}, neurons are often observed to be \term{polysemantic}, \emph{i.e.}, associated with multiple, unrelated concepts \citep{arora_linear_2018, mu_compositional_2020, elhage_softmax_2022, olah_zoom_2020}. For example, a single neuron may be activated by both images of cats and images of cars, suggesting it encodes multiple unrelated concepts. Polysemanticity contradicts the interpretation of neurons as representational primitives and, in practice, makes it challenging to understand the information processing of neural networks.

\paragraph{Exploring Polysemanticity: Hypotheses and Implications.}
To understand the widespread occurrence of polysemanticity in neural networks, several hypotheses have been proposed:
\begin{enumerate}[label=\emph{\roman*.)}]
\item One trivial scenario would be that feature directions are orthogonal but not aligned with the basis directions (neurons). There is no inherent reason to assume that features would align with neurons in a non-privileged basis, where the basis vectors are not architecturally distinguished. However, even in a privileged basis formed by the neurons, the network could represent features not in the standard basis but as linear combinations of neurons (see \figref{fig:privileged}, middle right).
\item An alternative hypothesis posits that \emph{redundancy due to noise} introduced during training, such as random dropout \citep{srivastava_dropout_2014}, can lead to redundant representations and, consequently, to polysemantic neurons \citep{marshall_understanding_2024}. This process involves distributing a single feature across several neurons rather than isolating it into individual ones, thereby encouraging polysemanticity.
\item Finally, the \term{superposition} hypothesis addresses the limitations in the network's representative capacity -- the number of neurons versus the number of crucial concepts. This hypothesis argues that the limited number of neurons compared to the vast array of important concepts necessitates a form of compression. As a result, an $n$-dimensional representation may encode features not with the $n$ basis directions (neurons) but with the $\propto \exp (n)$ possible almost orthogonal directions \citep{elhage_toy_2022}, leading to polysemanticity.
\end{enumerate}

\begin{hypbox}[Superposition ]
    Neural networks represent more features than they have neurons by encoding features in overlapping combinations of neurons.\label{hyp:superposition}
\end{hypbox}

\paragraph{Superposition Hypothesis.}
The \term{superposition} hypothesis suggests that neural networks can leverage high-dimensional spaces to represent more features than the actual count of neurons by encoding features in almost orthogonal directions. Non-orthogonality means that features interfere with one another. However, the benefit of representing many more features than neurons may outweigh the interference cost, mainly when concepts are sparse and non-linear activation functions can error-correct noise \citep{elhage_toy_2022}.

\begin{pinkbox}[Toy Model of Superposition]
A toy model \citep{elhage_toy_2022} investigates the hypothesis that neural networks can represent more \term{features} than the number of neurons by encoding real-world \term{concepts} in a compressed manner. The model considers a high-dimensional vector $\vx$, where each element $\evx_i$ corresponds to a feature capturing a real-world concept, represented as a random vector with varying importance determined by a weight $a_i$. These features are assumed to have the following properties: 
\begin{enumerate}[label=\emph{\roman*.)}]
    \item \textbf{Concept sparsity}: Real-world concepts occur sparsely. 
    \item \textbf{More concepts than neurons}: The number of potential concepts vastly exceeds the available neurons. 
    \item \textbf{Varying concept importance}: Some concepts are more important than others for the task at hand.
\end{enumerate}
The input vector $\vx$ represents features capturing these concepts, defined by a sparsity level $S$ and an importance level $a_i$ for each feature $\evx_i$, reflecting the sparsity and varying importance of the underlying concepts. The model dynamics involve transforming $\vx$ into a hidden representation $\vh$ of lower dimension, and then reconstructing it as $\vx'$:
\begin{equation*}
\vh = \mW\vx, \quad \vx' = \text{ReLU}(\mW^\top\vh + \vb).
\end{equation*}
The network's performance is evaluated using a loss function $\Ls$ weighted by the feature importances $a_i$, reflecting the importance of the underlying concepts:
\begin{equation*}
\Ls = \sum_{\vx}\sum_{i} a_i (\evx_i - \evx'_i)^2.
\end{equation*}
This toy model highlights neural networks' ability to encode numerous features representing real-world concepts into a compressed representation, providing insights into the superposition phenomenon observed in neural networks trained on real data.

\begin{minipage}{\textwidth}
\centering
\vspace{0.3cm}
\includegraphics[width=0.8\textwidth]{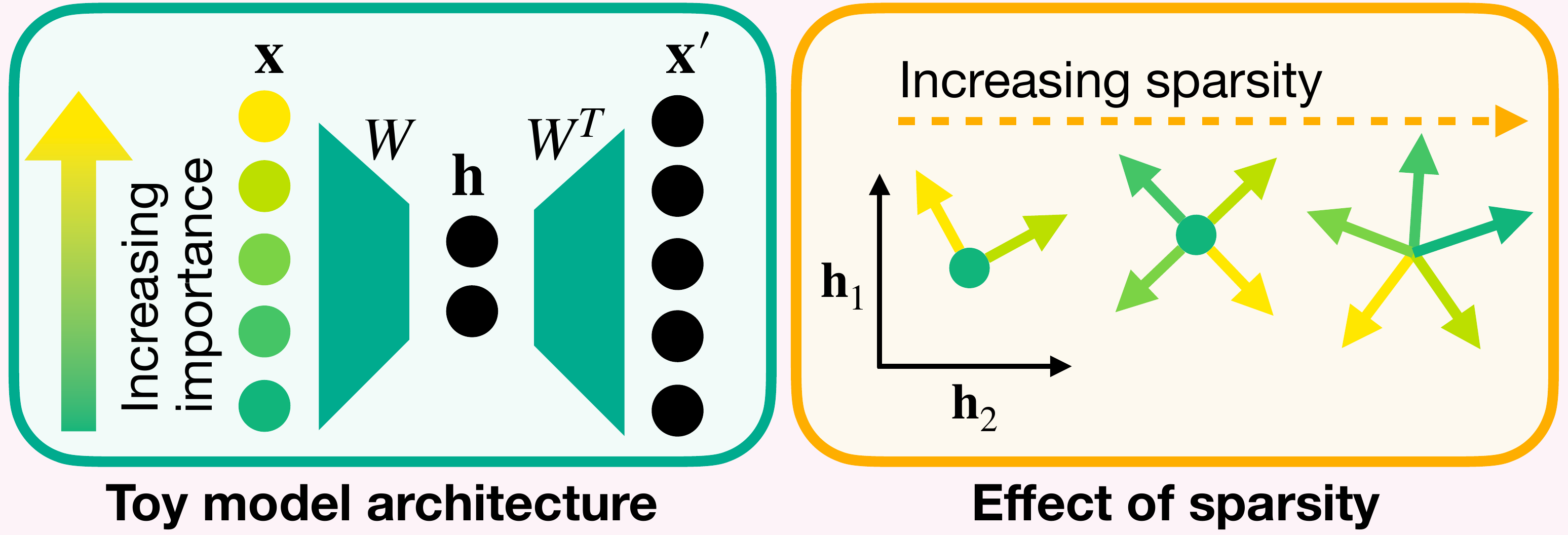}
\captionof{figure}{Illustration of the toy model architecture and the effects of sparsity. (left) Transformation of a five-feature input vector $\vx$ into a two-dimensional hidden representation $\vh$, and its reconstruction as $\vx'$ using the weight matrix $\mW$ and its transpose, with feature importance indicated by a color gradient from yellow to green. (right) The effect of increasing feature sparsity $S$ on the encoding capacity of the network, highlighting the network's enhanced ability to represent features in superposition as sparsity increases from $0$ to $0.9$, illustrated by arrows in the activation space $\vh$, which correspond to the columns of the matrix $\mW$.}
\label{fig:toy_model_of_superposition}
\end{minipage}

\end{pinkbox}

\begin{figure}[!htb]
\centering
\includegraphics[width=0.6\linewidth]{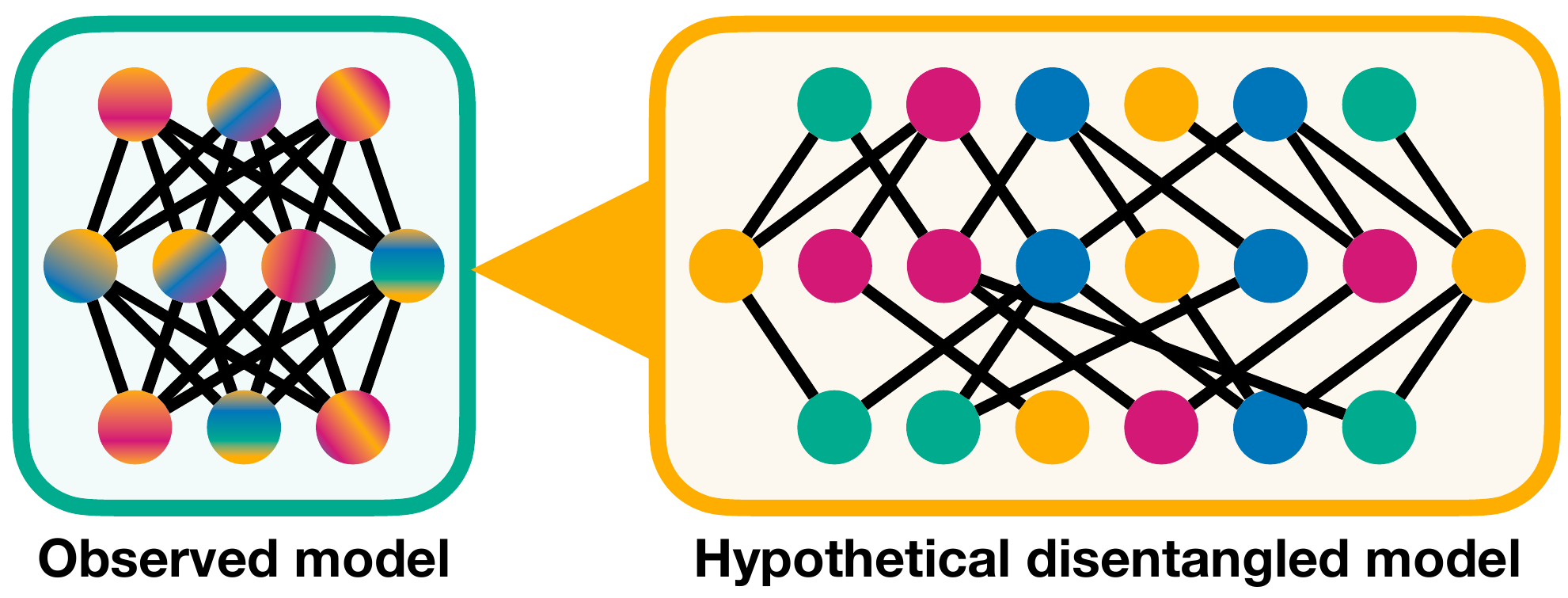}
\caption{Observed neural networks (left) can be viewed as compressed simulations of larger, sparser networks (right) where neurons represent distinct features. An "almost orthogonal" projection compresses the high-dimensional sparse representation, manifesting as polysemantic neurons involved with multiple features in the lower-dimensional observed model, reflecting the compressed encoding. Figure adapted from \citep{bricken_monosemanticity_2023}.}
\label{fig:polysemanticity}
\end{figure}

Toy models can demonstrate under which conditions superposition occurs \citep{elhage_toy_2022, scherlis_polysemanticity_2023}. Neural networks, via superposition, may effectively simulate computation with more neurons than they possess by allocating each feature to a linear combination of neurons, creating what is known as an overcomplete linear basis in the representation space. This perspective on superposition suggests that polysemantic models could be seen as compressed versions of hypothetically larger neural networks where each neuron represents a single concept (see \figref{fig:polysemanticity}). Consequently, an alternative definition of features could be:

\begin{defbox}[Feature (Alternative)]
Features are elements that a network would ideally assign to individual neurons if neuron count were not a limiting factor \citep{bricken_monosemanticity_2023}. In other words, \term{features} correspond to the disentangled \term{concepts} that a larger, sparser network with sufficient capacity would learn to represent with individual neurons.\label{def:feature_alternative}
\end{defbox}

Research on superposition, including works by \citep{elhage_toy_2022, scherlis_polysemanticity_2023, henighan_superposition_2023}, often investigates simplified models. However, understanding superposition in practical, transformer-based scenarios is crucial for real-world applications, as pioneered by \citet{gurnee_finding_2023}.

The need for understanding networks despite polysemanticity has led to various approaches: One involves training models without superposition \citep{jermyn_engineering_2022}, for example, using a softmax linear unit \citep{elhage_softmax_2022} as an activation function to empirically increase the number of \term{monosemantic} neurons, but at the cost of making other neurons less interpretable. From a capabilities standpoint, polysemanticity may be desirable as it allows models to represent more concepts with limited compute, making training cheaper. Overall, engineering monosemanticity has proven challenging \citep{bricken_monosemanticity_2023} and may be impractical until we have orders of magnitude more compute available.

Another approach is to train networks in a standard way (creating polysemanticity) and use post-hoc analysis to find the feature directions in activation space, for example, with Sparse Autoencoders (SAEs). SAEs aim to find the true, disentangled features in an uncompressed representation by learning a sparse overcomplete basis that describes the activation space of the trained model \citep{bricken_monosemanticity_2023, sharkey_taking_2022, cunningham_sparse_2024} (also see \secref{sec:methods:sparse_autoencoders}).

\paragraph{If not neurons, what are features then?}
We want to identify the fundamental units of neural networks, which we call \term{features}. Initially, neurons seemed likely candidates. However, this view fell short, particularly in transformer models where neurons often represent multiple concepts, a phenomenon known as polysemanticity. The superposition hypothesis addresses this, proposing that due to limited representational capacity, neural networks compress numerous features into the confined space of neurons, complicating interpretation.

This raises the question: \emph{How are features encoded if not in discrete neuron units?} While a priori features could be encoded in an arbitrarily complex, non-linear structure, a growing body of theoretical arguments and empirical evidence supports the hypothesis that features are commonly represented linearly, i.e., as linear combinations of neurons -- hence, as directions in representation space. This perspective promises to enhance our comprehension of neural networks by providing a more interpretable and manipulable framework for their internal representations.

\begin{hypbox}[Linear Representation] Features are directions in activation space, \emph{i.e.}, linear combinations of neurons.\label{hyp:linear}  
\end{hypbox}

The \term{linear representation} hypothesis suggests that neural networks frequently represent high-level features as linear directions in activation space. This hypothesis can simplify the understanding and manipulation of neural network representations \citep{nanda_emergent_2023}. The prevalence of linear layers in neural network architectures favors linear representations. Matrix multiplication in these layers most readily processes linear features, while more complex non-linear encodings would require multiple layers to decode. 

However, recent work by \citet{engels_not_2024} provides evidence against a strict formulation of the linear representation hypothesis by identifying circular features representing days of the week and months of the year. These multi-dimensional, non-linear representations were shown to be used for solving modular arithmetic problems in days and months. Intervention experiments confirmed that these circular features are the fundamental unit of computation in these tasks, and the authors developed methods to decompose the hidden states, revealing the circular representations.

Establishing non-linearity can be challenging. For example, \citet{li_emergent_2023} initially found that in a GPT model trained on Othello, the board state could only be decoded with a non-linear probe when represented in terms of white and black pieces, seemingly violating the linearity assumption. However, \citet{nanda_actually_2023, nanda_emergent_2023} later showed that a linear probe sufficed when the board state was decoded in terms of "one's own" and "the opponent's" pieces, reaffirming the linear representation hypothesis in this case. In contrast, the work by \citet{engels_not_2024} provides a clear and convincing existence proof for non-linear, multi-dimensional representations in language models.

While the linear representation hypothesis remains a useful simplification, it is important to recognize its limitations and the potential role of non-linear representations \citep{sharkey_current_2022}. As neural networks continue to evolve, ongoing reevaluation of the hypothesis is crucial, particularly considering the possible emergence of non-linear features under optimization pressure for interpretability \citep{hubinger_transparency_2022}. Alternative perspectives, such as the polytope lens proposed by \citet{black_interpreting_2022}, emphasize the impact of non-linear activation functions and discrete polytopes formed by piecewise linear activations as potential primitives of neural network representations.

Despite these exceptions, empirical evidence largely supports the linear representation hypothesis in many contexts, especially for feedforward networks with ReLU activations. Semantic vector calculus in word embeddings \citep{mikolov_distributed_2013}, successful linear probing \citep{alain_understanding_2016, belinkov_probing_2021}, sparse dictionary learning \citep{bricken_monosemanticity_2023, cunningham_sparse_2024, deng_measuring_2023}, and linear decoding of concepts \citep{omahony_disentangling_2023}, tasks \citep{hendel_incontext_2023}, functions \citep{todd_function_2023}, sentiment \citep{tigges_language_2024}, refusal \citep{arditi_refusal_2024}, and relations \citep{hernandez_linearity_2023, chanin_identifying_2023} in large language models all point to the prevalence of linear representations. Moreover, linear addition techniques for model steering \citep{turner_activation_2023, sakarvadia_memory_2023, li_inferencetime_2023} and \term{representation engineering} \citep{zou_representation_2023} highlight the practical implications of linear feature representations. 

Building upon the linear representation hypothesis, recent work investigated the structural organization of these linear features within activation space. \citet{park_geometry_2024} reveal a geometric framework for categorical and hierarchical concepts in large language models. Their findings demonstrate that simple categorical concepts (e.g., mammal, bird) are represented as simplices in the activation space, while hierarchically related concepts are orthogonal. This geometric analysis aligns with earlier observations on feature clustering and splitting in neural networks \citep{elhage_toy_2022}. It suggests that the linear features are not merely scattered directions but are organized to reflect semantic relationships and hierarchies.

\subsection{Circuits as Computational Primitives and Motifs as Universal Circuit Patterns}\label{sec:concepts:computation}
Having defined features as directions in activation space as the fundamental units of neural network representation, we now explore their computation. Neural networks can be conceptualized as computational graphs, within which \term{circuits} are sub-graphs consisting of linked features and the weights connecting them. Similar to how features are the representational primitive, circuits function as the computational primitive \citep{michaud_quantization_2023} and the primary building block of these networks \citep{olah_zoom_2020}.

\begin{figure}[!htb]
\centering
\includegraphics[width=1.\linewidth]{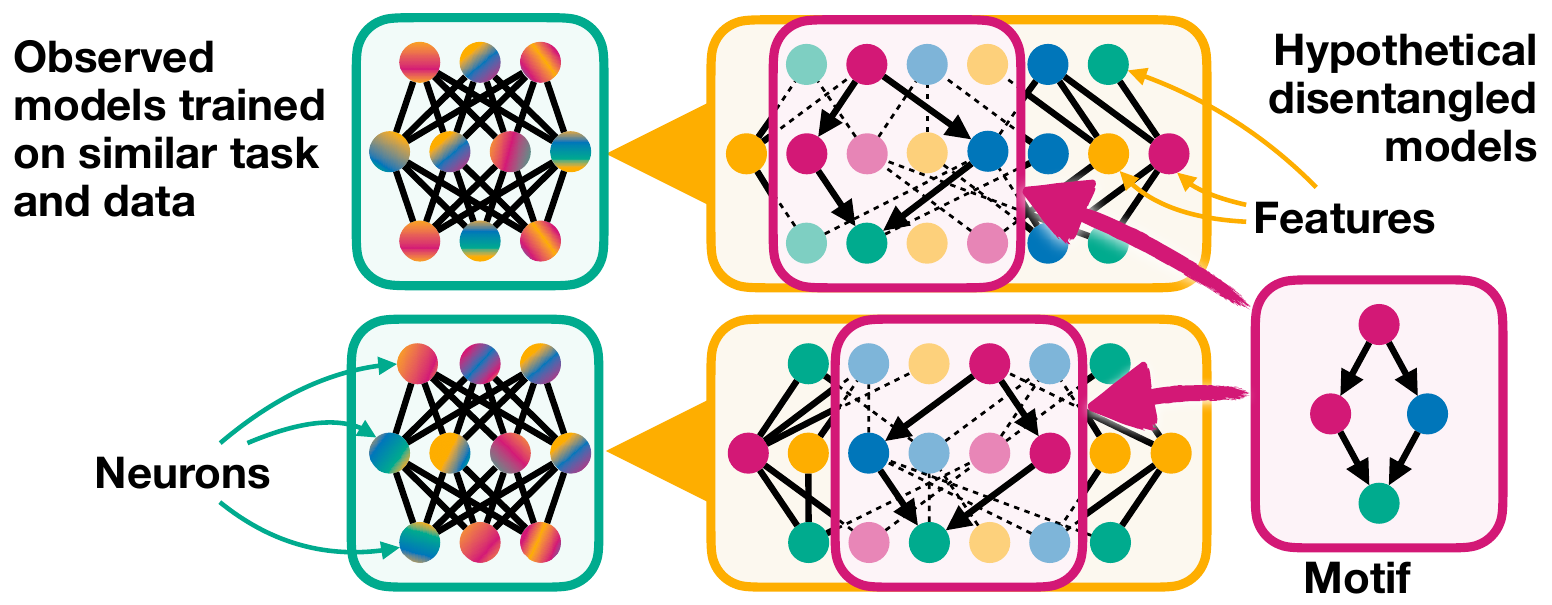}
\caption{Comparing observed models (left) and corresponding hypothetical \term{disentangled} models (right) trained on similar tasks and data. The observed models show different neuronal activation patterns, while the dissection into feature-level \term{circuits} reveals a \emph{motif} - a shared {circuit} pattern emerging across models, hinting at \term{universality} -- models converging on similar solutions based on common underlying principles.}
\label{fig:motif}
\end{figure}

\begin{defbox}[Circuit]
Circuits are sub-graphs of the network, consisting of features and the weights connecting them.\label{def:circuit}
\end{defbox}

The decomposition of neural networks into circuits for interpretability has shown significant promise, particularly in small models trained for specific tasks such as addition, as seen in the work of \citet{nanda_progress_2023} and \citet{quirke_understanding_2023}. Scaling such a \textit{comprehensive} circuit analysis to broader behaviors in large language models remains challenging. However, there has been notable progress in scaling circuit analysis of \textit{narrow behaviors} to larger circuits and models, such as indirect object identification \citep{wang_interpretability_2023} and greater-than computations \citep{hanna_how_2023} in GPT-2 and multiple-choice question answering \citep{lieberum_does_2023}.

In search of general and universal circuits, researchers focus particularly on more general and transferable behaviors. \citet{mcdougall_copy_2023}'s work on copy suppression in GPT-2's attention heads sheds light on model calibration and self-repair mechanisms. \citet{davies_discovering_2023} and \citet{feng_how_2023} focus on how large language models represent symbolic knowledge through variable binding and entity-attribute binding, respectively. \citet{yu_characterizing_2023, nanda_fact_2023, lv_interpreting_2024, chughtai_summing_2024, ortu_competition_2024} explore mechanisms for factual recall, revealing how circuits dynamically balance pre-trained knowledge with new contextual information. \citet{lan_locating_2023} extend circuit analysis to sequence continuation tasks, identifying shared computational structures across semantically related sequences.

More promisingly, some repeating patterns have shown \term{universality} across models and tasks. These universal patterns are called \term{motifs} \citep{olah_zoom_2020} and can manifest not just as specific circuits or features but also as higher-level behaviors emerging from the interaction of multiple components. Examples include the curve detectors found across vision models \citep{cammarata_curve_2021,cammarata_curve_2020}, induction circuits enabling in-context learning \citep{olsson_incontext_2022}, and the phenomenon of branch specialization in neural networks \citep{voss_branch_2021}. Motifs may also capture how models leverage tokens for working memory or parallelize computations in a divide-and-conquer fashion across representations. The significance of motifs lies in revealing the common structures, mechanisms, and strategies that naturally emerge across neural architectures, shedding light on the fundamental building blocks underlying their intelligence. Figure \ref{fig:motif} contrasts observed neural network models with hypothetical disentangled models, illustrating how a shared circuit pattern can emerge across different models trained on similar tasks and data, hinting at an underlying \term{universality}.

\begin{defbox}[Motif]
Motifs are repeated patterns within a network, encompassing either features or circuits that emerge across different models and tasks.\label{def:motif}
\end{defbox}

\paragraph{Universality Hypothesis.}
Following the evidence for motifs, we can propose two versions for a \term{universality} hypothesis regarding the convergence of features and circuits across neural network models:

\begin{hypbox}[Weak Universality]
There are underlying principles governing how neural networks learn to solve certain tasks. Models will generally converge on analogous solutions that adhere to the common underlying principles. However, the specific \term{features} and \term{circuits} that implement these principles can vary across different models based on factors like hyperparameters, random seeds, and architectural choices. 
\end{hypbox}

\begin{hypbox}[Strong Universality]
The \textit{same} core features and circuits will universally and consistently arise across all neural network models trained on similar tasks and data distributions and using similar techniques, reflecting a set of fundamental computational \term{motifs} that neural networks inherently gravitate towards when learning.
\end{hypbox}

The universality hypothesis posits a convergence in forming features and circuits across various models and tasks, which could significantly ease interpretability efforts in AI. It proposes that artificial and biological neural networks share similar features and circuits, suggesting a standard underlying structure \citep{chan_natural_2023, sucholutsky_getting_2023, kornblith_similarity_2019}. This idea posits that there is a fundamental basis in how neural networks, irrespective of their specific configurations, process and comprehend information. This could be due to inbuilt inductive biases in neural networks or \term{natural abstractions} \citep{chan_natural_2023} -- concepts favored by the natural world that any cognitive system would naturally gravitate towards.

Evidence for this hypothesis comes from \textit{cross-species neural structures} in neuroscience, where similar neural structures and functions are found in different species \citep{kirchner_neuroscience_2023}. Additionally, machine learning models, including neural networks, tend to converge on similar features, representations, and classifications across different tasks and architectures \citep{chen_going_2023, hacohen_let_2019, li_convergent_2015, bricken_monosemanticity_2023}. \citet{marchetti_harmonics_2023} provide mathematical support for emerging universal features. 

While various studies support the universality hypothesis, questions remain about the extent of feature and circuit similarity across different models and tasks. % Nevertheless, this concept bridges AI and other scientific disciplines, offering cross-disciplinary applications and a deeper understanding of artificial and natural cognitive processes. 
In the context of mechanistic interpretability, this hypothesis has been investigated for neurons \citep{gurnee_universal_2024}, group composition circuits \citep{chughtai_toy_2023}, and modular task processing \citep{variengien_look_2023}, with evidence for the weak but not the strong formulation \citep{chughtai_toy_2023}.

\subsection{Emergence of World Models and Simulated Agents}\label{sec:concepts:emergence}

\paragraph{Internal World Models.}
World models are internal causal models of an environment formed within neural networks. Traditionally linked with reinforcement learning, these models are \emph{explicitly} trained to develop a compressed spatial and temporal representation of the training environment, enhancing downstream task performance and sample efficiency through training on internal hallucinations \citep{ha_recurrent_2018}. However, in the context of our survey, our focus shifts to \term{internal world models} that potentially form \emph{implicitly} as a by-product of the training process, especially in LLMs trained on next-token prediction -- also called GPT.

LLMs are sometimes characterized as \emph{stochastic parrots} \citep{bender_dangers_2021}. This label stems from their fundamental operational mechanism of predicting the next word in a sequence, which is seen as relying heavily on memorization. From this viewpoint, LLMs are thought to form complex correlations based on observational data but cannot develop causal models of the world due to their lack of access to interventional data \citep{pearl_causality_2009}.

An alternative perspective on LLMs comes from the \textit{active inference} framework \citep{salvatori_braininspired_2023}, a theory rooted in cognitive science and neuroscience. Active inference postulates that the objective of minimizing prediction error, given enough representative capacity, is adequate for a learning system to develop complex world representations, behaviors, and abstractions. Since language inherently mirrors the world, these models could implicitly construct linguistic and broader world models \citep{kulveit_predictive_2023}.

The \term{simulation} hypothesis suggests that models designed for prediction, such as LLMs, will eventually simulate the causal processes underlying data creation. Seen as an extension of their drive for efficient compression, this hypothesis implies that adequately trained models like GPT could develop \term{internal world models} as a natural outcome of their predictive training \citep{janus_simulators_2022, shanahan_role_2023}.

\begin{hypbox}[Simulation]
A model whose objective is text prediction will simulate the causal processes underlying the text creation if optimized sufficiently strongly \citep{janus_simulators_2022}.\label{hyp:simulation}
\end{hypbox}

In addition to theoretical considerations for emergent causal world models \citep{richens_robust_2024, nichani_how_2024}, mechanistic interpretability is starting to provide empirical evidence on the types of internal world models that may emerge in LLMs. The ability to internally represent the board state in games like chess \citep{karvonen_emergent_2024} or Othello \citep{li_emergent_2023, nanda_emergent_2023}, create linear abstractions of spatial and temporal data \citep{gurnee_language_2023}, and structure complex representations of mazes, demonstrating an understanding of maze topology and pathways \citep{ivanitskiy_structured_2023} highlight the growing abstraction capabilities of LLMs. \citet{li_implicit_2021} identified contextual word representations that function as models of entities and situations evolving throughout a discourse, akin to linguistic models of dynamic semantics. \citet{patel_mapping_2022} demonstrated that LLMs can map conceptual domains (e.g., direction, color) to grounded world representations given a few examples, suggesting they learn rich conceptual spaces \citep{gardenfors_conceptual_2004} reflective of the non-linguistic world.

The \term{prediction orthogonality} hypothesis further expands on this idea: It posits that prediction-focused models like GPT may simulate agents with various objectives and levels of optimality. In this context, GPT are simulators, simulating entities known as \term{simulacra} that can be either agentic or non-agentic, with different objectives from the simulator itself \citep{janus_simulators_2022, shanahan_role_2023}. The implications of the simulation and prediction orthogonality hypotheses for AI safety and alignment are discussed in \secref{sec:relevance}.

\begin{hypbox}[Prediction Orthogonality]
A model whose objective is prediction can simulate agents who optimize toward any objectives with any degree of optimality \citep{janus_simulators_2022}.\label{hyp:prediction_orthogonality}
\end{hypbox}

In conclusion, the evolution of LLMs from simple predictive models to entities potentially possessing complex \term{internal world models}, as suggested by the \term{simulation} hypothesis and supported by mechanistic interpretability studies, represents a significant shift in our understanding of these systems. This evolution challenges us to reconsider LLMs' capabilities and future trajectories in the broader landscape of AI development.

\section{Core Methods}\label{sec:methods}
\begin{figure}[!htb]
\centering
\includegraphics[width=\textwidth]{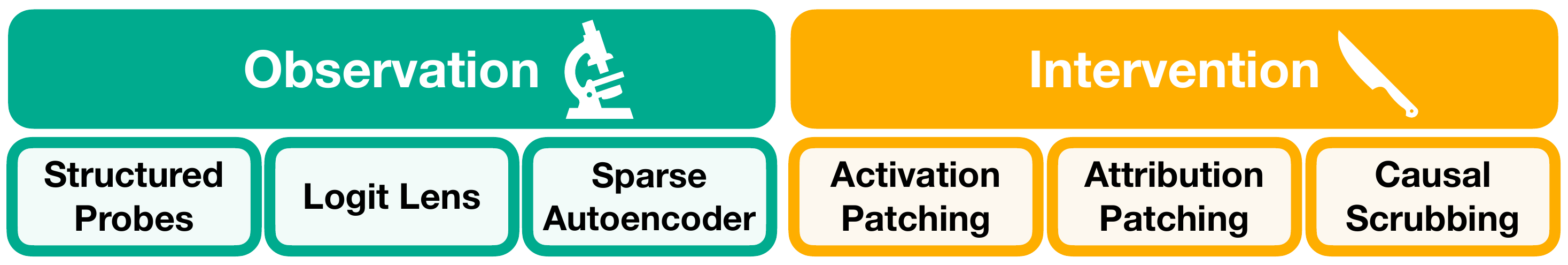}
\caption{Overview of key methods and techniques in mechanistic interpretability research. Observational approaches include structured probes, logit lens variants, and sparse autoencoders (SAEs). Interventional methods, focusing on causal understanding, encompass activation patching variants for uncovering causal mechanisms and causal scrubbing for hypothesis evaluation.}
\label{fig:methods}
\end{figure}

Mechanistic interpretability (MI) employs various tools, from observational analysis to causal interventions. This section provides a comprehensive overview of these methods, beginning with a taxonomy that categorizes approaches based on their key characteristics (\secref{sec:methods:taxonomy}). We then survey observational (\secref{sec:methods:microscope}), followed by interventional techniques (\secref{sec:methods:scalpel}). Finally, we study their synergistic interplay (\secref{sec:methods:integrate}). \figref{fig:methods} offers a visual summary of the methods and techniques unique to mechanistic interpretability.

\subsection{Taxonomy of Mechanistic Interpretability Methods}
\label{sec:methods:taxonomy}

We propose a taxonomy based on four key dimensions: causal nature, learning phase, locality, and comprehensiveness (Table \ref{tab:mi_taxonomy}).

The causal nature of methods ranges from purely observational, which analyze existing representations without direct manipulation, to interventional approaches that actively perturb model components to establish causal relationships. The learning phase dimension distinguishes between post-hoc techniques applied to trained models and intrinsic methods that enhance interpretability during the training process itself.

Locality refers to the scope of analysis, spanning from individual neurons (e.g., feature visualization) to entire model architectures (e.g., causal abstraction). Comprehensiveness varies from partial insights into specific components to holistic explanations of model behavior.

\begin{table}[htbp]

\centering

\caption{Taxonomy of Mechanistic Interpretability Methods}

\label{tab:mi_taxonomy}

\resizebox{\textwidth}{!}{%

\begin{tabular}{@{}llllll@{}}

\toprule

Method & Causal Nature & Phase & Locality & Comprehensiveness & Key Examples \\

\midrule

Feature Visualization & Observation & Post-hoc & Local & Partial & \citet{zeiler_visualizing_2014} \\

 & & & & & \citet{zimmermann_how_2021} \\

Exemplar methods & Observation & Post-hoc & Local & Partial & \citet{grosse_studying_2023} \\

 & & & & & \citet{garde_deepdecipher_2023} \\

Probing Techniques & Observation & Post-hoc & Both & Both & \citet{mcgrath_acquisition_2022} \\

 & & & & & \citet{gurnee_finding_2023} \\

Structured Probes & Observation & Post-hoc & Both & Both & \citet{burns_discovering_2023} \\

Logit Lens Variants & Observation & Post-hoc & Global & Partial & \citet{nostalgebraist_interpreting_2020} \\

 & & & & & \citet{belrose_eliciting_2023} \\

Sparse Autoencoders & Observation & Post-hoc & Both & Comprehensive & \citet{cunningham_sparse_2024} \\

 & & & & & \citet{bricken_monosemanticity_2023} \\

Activation Patching & Intervention & Post-hoc & Local & Partial & \citet{meng_locating_2022} \\

 & & & & & \citet{wang_interpretability_2023} \\

Path Patching & Intervention & Post-hoc & Both & Both & \citet{goldowsky-dill_localizing_2023} \\

Causal Abstraction & Intervention & Post-hoc & Global & Comprehensive &  \citet{geiger_causal_2023} \\
 
 & & & & & \citet{geiger_finding_2023} \\

 & & & & & \citet{wu_causal_2023} \\

Hypothesis Testing & Intervention & Post-hoc & Global & Comprehensive & \citet{chan_causal_2022} \\

 & & & & & \citet{jenner_comparison_2023} \\

Intrinsic Methods & -- & Pre/During & Global & Comprehensive & \citet{elhage_softmax_2022} \\

 & & & & & \citet{liu_seeing_2023} \\

\bottomrule

\end{tabular}%

}

\end{table}

The categorization is based on the methods' general tendencies. Some methods can offer local and global or partial and comprehensive interpretability depending on the scope of the analysis and application. Probing techniques can range from local to global and partial to comprehensive; simple linear probes might offer local insights into individual \term{features}, while more sophisticated structured probes can uncover global patterns. Sparse autoencoders decompose individual neuron activations (local) but aim to disentangle features across the entire model (global). Path patching extends local interventions to global model understanding by tracing information flow across layers, demonstrating how local perturbations can yield broader insights.

In practice, mechanistic interpretability research involves both method development and their application. When applying methods to understand a model, combining techniques from multiple categories is often necessary and beneficial to build a more comprehensive understanding (\secref{sec:methods:integrate}).

\subsection{Observation}
\label{sec:methods:microscope}

Mechanistic interpretability draws from observational methods that analyze the inner workings of neural networks, with many of these methods preceding the field itself. For a detailed exploration of inner interpretability methods, refer to \citep{rauker_transparent_2023}. Two prominent categories are example-based methods and feature-based methods:
\begin{enumerate}[label=\emph{\roman*.)}]
\item \textbf{Exemplar methods} identify real input examples that highly activate specific neurons or layers. This helps pinpoint influential data points that maximize neuron activation within the neural network \citep{grosse_studying_2023, garde_deepdecipher_2023, nanfack_adversarial_2024}.
\item \textbf{Feature visualization} encompasses techniques that generate synthetic inputs to optimize neuron activation. These visualizations reveal how neurons respond to stimuli and which features are sensitive to \citep{zeiler_visualizing_2014, zimmermann_how_2021}. By inspecting the synthetic inputs that drive neuron behavior, we can hypothesize about the features encoded by those neurons.
\end{enumerate}

\paragraph{Probing for Features.}
\label{sec:methods:probing}

Probing \citep{alain_understanding_2016, hewitt_structural_2019} involves training a classifier using the activations of a model, with the classifier's performance subsequently observed to deduce insights about the model's behavior and internal representations. However, the probe's performance may often reflect its own learning capacities more than the actual characteristics of the model's representations \citep{belinkov_probing_2021}. This dilemma has led researchers to investigate the ideal balance between the complexity of a probe and its capacity to accurately represent the model's features \citep{cao_lowcomplexity_2021, voita_informationtheoretic_2020}.

The \term{linear representation} hypothesis offers a resolution to this issue. Under this hypothesis, the failure of a simple linear probe to detect certain features suggests their absence in the model's representations. Conversely, suppose a more complex probe succeeds where a simpler one fails. In that case, it implies that the model contains features that a complex function can combine into the target feature but that the target feature itself is not explicitly represented. Thus, the hypothesis implies that using linear probes could suffice in most cases, circumventing the complexity considerations generally associated with probing \citep{belinkov_probing_2021}.

\citet{mcgrath_acquisition_2022} analyzed chess knowledge acquisition in AlphaZero, revealing the emergence of strategic concepts during training. In language models, \citet{gurnee_finding_2023} introduced \emph{sparse probing} to decode internal neuron activations to understand feature representation and sparsity. They show that early layers use sparse combinations of neurons to represent many features in superposition, while middle layers seem to have dedicated \term{monosemantic} neurons for higher-level contextual features.

Probing is limited in drawing causal or behavioral conclusions. Its primarily observational nature focuses on how information is encoded rather than how it is used (see \figref{fig:paradigms}), necessitating careful analysis and integration with interventional techniques (\secref{sec:methods:scalpel}), or alternative approaches \citep{elazar_amnesic_2021}. While in explainable AI, probing has primarily analyzed high-level concepts like linguistic representations \citep{tenney_bert_2019, dalvi_what_2019}, MI aims to probe towards uncovering underlying computational processes and functionality. This shift in goals towards uncovering mechanistic computation is a nuanced distinction rather than a clear-cut line between probing in MI and the broader explainability field.

\paragraph{Structured Probes.} \label{sec:methods:structured_probes}
While focusing on bottom-up, mechanistic interpretability approaches, we can also consider integrating top-down, concept-based structured probes with mechanistic interpretability.

Structured probes aid conceptual interpretability, probing language models for complex features like truth representations. Notably, \citet{burns_discovering_2023}'s \textit{contrast-consistent search} identifies linear projections exhibiting logical consistency in hidden states, contrasting truth values for statements and negations.

However, structured probes face significant challenges in unsupervised probing scenarios. As \citet{farquhar_challenges_2023} showed, arbitrary features, not just knowledge-related ones, can satisfy contrast consistency equally well, raising doubts about scalability. For example, the loss may capture \term{simulation} of knowledge from hypothesized \term{simulacra} within sufficiently powerful language models rather than the models' true knowledge. Furthermore, \citet{farquhar_challenges_2023} demonstrates self-supervised probing methods (like \citep{burns_discovering_2023}) often detect prominent but unintended distractor features in the data. The discovered features are also highly sensitive to prompt choice, and there is no principled way to select prompts that would reliably surface a model's true knowledge.

While structured probes primarily focus on high-level conceptual representations \citep{zou_representation_2023}, their findings could potentially inform or complement mechanistic interpretability efforts. For instance, identifying truth directions through structured probes could help guide targeted interventions or analyze the underlying circuits responsible for truthful behavior using mechanistic techniques such as activation patching or circuit tracing (\secref{sec:methods:scalpel}). Conversely, mechanistic methods could provide insights into how truth representations emerge and are computed within the model, addressing some of the challenges faced by unsupervised structured probes.

\paragraph{Logit Lens.}
The \textit{logit lens} \citep{nostalgebraist_interpreting_2020} provides a window into the model's predictive process by applying the final classification layer (which projects the residual stream activation into logits/vocabulary space) to intermediate activations of the residual stream, revealing how prediction confidence evolves across computational stages. This is possible because transformers tend to build their predictions across layers iteratively \citep{geva_transformer_2022}. Extensions of this approach include the tuned lens \citep{belrose_eliciting_2023}, which trains affine probes to decode hidden states into probability distributions over the vocabulary, and the Future Lens \citep{pal_future_2023}, which explores the extent to which individual hidden states encode information about subsequent tokens.

Researchers have also investigated techniques that bypass intermediate computations to probe representations directly. \citet{din_jump_2023} propose using linear transformations to approximate hidden states from different layers, revealing that language models often predict final outputs in early layers. \citet{dar_analyzing_2022} present a theoretical framework for interpreting transformer parameters by projecting them into the embedding space, enabling model alignment and parameter transfer across architectures.

Other techniques focus on interpreting specific model components or submodules. The DecoderLens \citep{langedijk_decoderlens_2023} allows analyzing encoder-decoder transformers by cross-attending intermediate encoder representations in the decoder, shedding light on the information flow within the encoder. The Attention Lens \citep{sakarvadia_attention_2023} aims to elucidate the specialized roles of attention heads by translating their outputs into vocabulary tokens via learned transformations.

\paragraph{Feature Disentanglement via Sparse Dictionary Learning.} \label{sec:methods:sparse_autoencoders}
Recent work suggests that the essential elements in neural networks are linear combinations of neurons representing features in superposition \citep{elhage_toy_2022}. To disentangle these features, researchers have developed sparse autoencoders (SAEs), which decompose neural network activations into individual component features \citep{sharkey_taking_2022, cunningham_sparse_2024}. This process, known as sparse dictionary learning, reconstructs activation vectors as sparse linear combinations of directional vectors within the activation space \citep{olshausen_sparse_1997}.

The theoretical foundations of SAEs are rooted in work on \term{disentangled} representations. \citet{whittington_disentangling_2022} demonstrate that autoencoders can recover ground truth features under conditions of feature sparsity and non-negativity. Furthermore, \citet{garfinkle_uniqueness_2019} provides guarantees for the uniqueness and stability of dictionaries for sparse representation, even in the presence of noise. These theoretical underpinnings support SAEs' ability to uncover true, disentangled features underlying the data distribution.

In practice, SAEs stand out for their simplicity and scalability \citep{sharkey_taking_2022}. They incorporate sparsity regularization to encourage learning sparse yet meaningful data representations, with the precise tuning of the sparsity penalty on hidden activations critical in dictating the autoencoder's sparsity level. We provide an overview of the SAE architecture in \figref{fig:sparse_autoencoder}.

SAEs' dictionary features exhibit higher scores on autointerpretability metrics and increased monosemanticity \citep{bricken_monosemanticity_2023, cunningham_sparse_2024, sharkey_taking_2022}. They are scalable to state-of-the-art models and can detect safety-relevant features \citep{templeton_scaling_2024}, measure feature sparsity \citep{deng_measuring_2023}, and interpret reward models in reinforcement learning-based language models \citep{marks_interpreting_2023}.

Evaluating SAE quality remains challenging due to the lack of ground-truth interpretable features. Researchers have addressed this through various approaches: \citet{karvonen_measuring_2024} proposed using language models trained on chess and Othello transcripts as testbeds, providing natural collections of interpretable features. \citet{sharkey_taking_2022} constructed a toy model with traceable features, while \citet{makelov_principled_2024, makelov_sparse_2024} compared SAE results with supervised features in large language models to demonstrate their viability.

The versatility of SAEs extends to various neural network architectures. They have been successfully applied to transformer attention layers \citep{kissane_interpreting_2024} and convolutional neural networks \citep{gorton_missing_2024}. Notably, \citet{gorton_missing_2024} applied SAEs to the early vision layers of InceptionV1, uncovering new interpretable features, including additional curve detectors not apparent from examining individual neurons \citep{cammarata_curve_2020}.

In circuit discovery, SAEs have shown particular promise (see also \secref{sec:methods:integrate}). \citet{he_dictionary_2024} proposed a circuit discovery framework alternative to activation patching (discussed in \secref{sec:methods:patching}), leveraging dictionary features decomposed from all modules writing to the residual stream. Similarly, \citet{oneill_sparse_2024} employed discrete sparse autoencoders for discovering interpretable circuits in large language models.

Recent advancements have focused on improving SAE performance and addressing limitations. \citet{rajamanoharan_improving_2024} introduced a gating mechanism to separate the functionalities of determining which directions to use and estimating their magnitudes, mitigating shrinkage -- the systematic underestimation of feature activations. An alternative approach by \citet{dunefsky_transcoders_2024} uses transcoders to faithfully approximate a densely activating MLP layer with a wider, sparsely-activating MLP layer, offering another path to interpretable feature discovery, a type of sparse distillation \citep{slavachalnev_sparse_2024}.

\begin{pinkbox}[Sparse Dictionary Learning]\label{box:sparse_autoencoders}
Sparse autoencoders \citep{cunningham_sparse_2024} are proposed as a solution to \term{polysemantic} neurons. The problem of \term{superposition} is mathematically formalized as \textit{sparse dictionary learning} \citep{olshausen_sparse_1997} problem to decompose neural network activations into \term{disentangled} component features.
The goal is to learn a dictionary of vectors $\{\vf_k\}_{k=1}^{n_{\text{feat}}} \subset \sR^d$ that can represent the unknown, ground truth network features %$\{\vg_j\}_{j=1}^{n_{\text{gt}}}$
as sparse linear combinations. If successful, the learned dictionary contains \term{monosemantic} neurons corresponding to \term{features} \citep{sharkey_taking_2022}.
The autoencoder architecture consists of an encoder and a ReLU activation function, expanding the input dimensionality to $d_{\text{hid}} > d_{\text{in}}$. The encoder's output is given by:
\begin{align}
\vh &= \text{ReLU}(\mW_{\text{enc}}\vx+\vb),  \\
\vx' &= \mW_{\text{dec}}\vh = \sum_{i=0}^{d_{\text{hid}}-1} h_i \vf_i,
\end{align}
where $\mW_{\text{enc}}, \mW_{\text{dec}}^\top \in \sR^{d_{\text{hid}} \times d_{\text{in}}}$ and $\vb \in \sR^{d_{\text{hid}}}$. The parameter matrix $\mW_{\text{dec}}$ forms the feature dictionary, with rows $\vf_i$ as dictionary features. The autoencoder is trained to minimize the loss, where the $\normlone$ penalty on $\vh$ encourages sparse reconstructions using the dictionary features,
\begin{equation}\label{eq:sae_loss}
\Ls(\vx) = |\vx - \vx'|_2^2 + \alpha |\vh|_1.
\end{equation}
\begin{minipage}{\textwidth}
\centering
\includegraphics[width=0.8\textwidth]{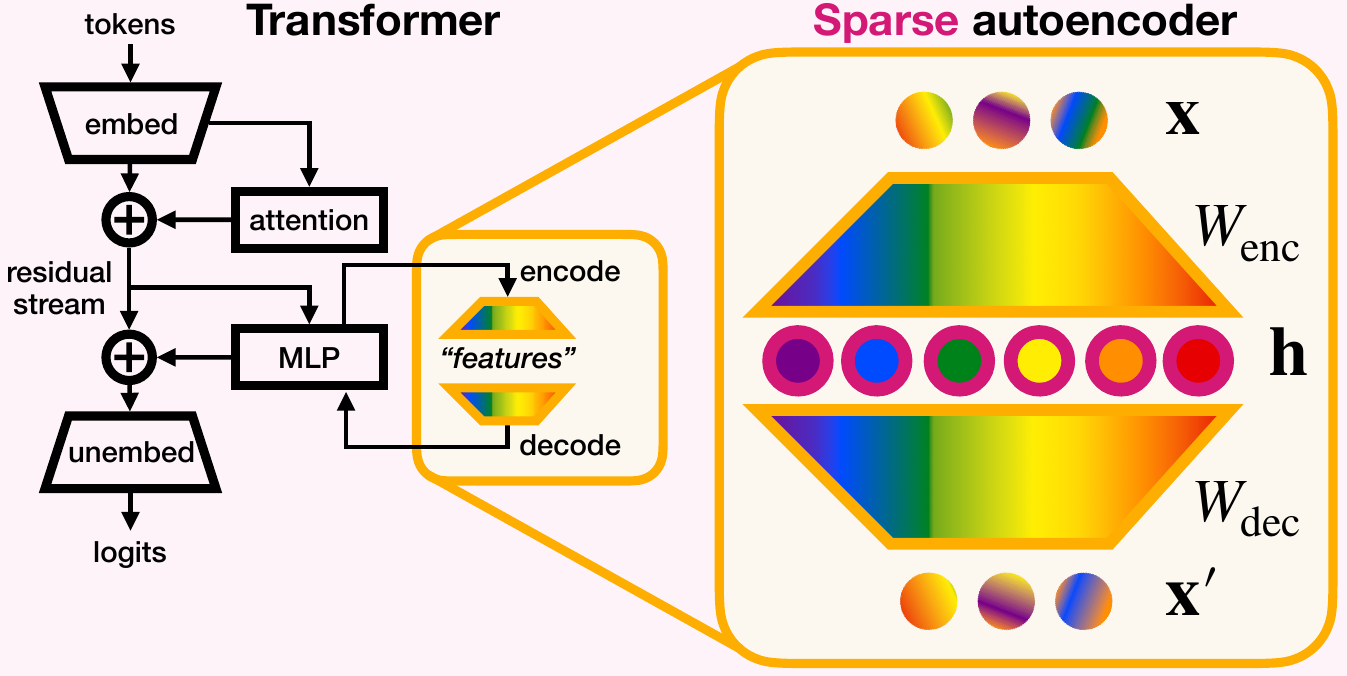}
\captionof{figure}{Illustration of a sparse autoencoder applied to the MLP layer activations, consisting of an encoder that increases dimensionality while emphasizing sparse representations and a decoder that reconstructs the original activations using the learned feature dictionary.}
\label{fig:sparse_autoencoder}
\end{minipage}
\end{pinkbox}

\subsection{Intervention}\label{sec:methods:scalpel}

\paragraph{Causality as a Theoretical Foundation.} The theory of causality \citep{pearl_causality_2009} provides a mathematically precise framework for mechanistic interpretability, offering a rigorous approach to understanding high-level semantics in neural representations \citep{geiger_causal_2023}. By treating neural networks as causal models, with their \textit{compute graphs serving as causal graphs}, researchers can perform precise interventions and examine the roles of individual parameters \citep{mueller_quest_2024}. This causal perspective on interpretability has led to the development of various intervention techniques, including activation patching (\secref{sec:methods:patching}), causal abstraction (\secref{sec:methods:abstraction}), and hypothesis testing methods (\secref{sec:methods:hypothesis}). 

\subsubsection{Activation Patching}\label{sec:methods:patching}

\begin{figure}[ht]
\centering
\begin{subfigure}[b]{0.49\textwidth}
\centering
\includegraphics[width=\textwidth]{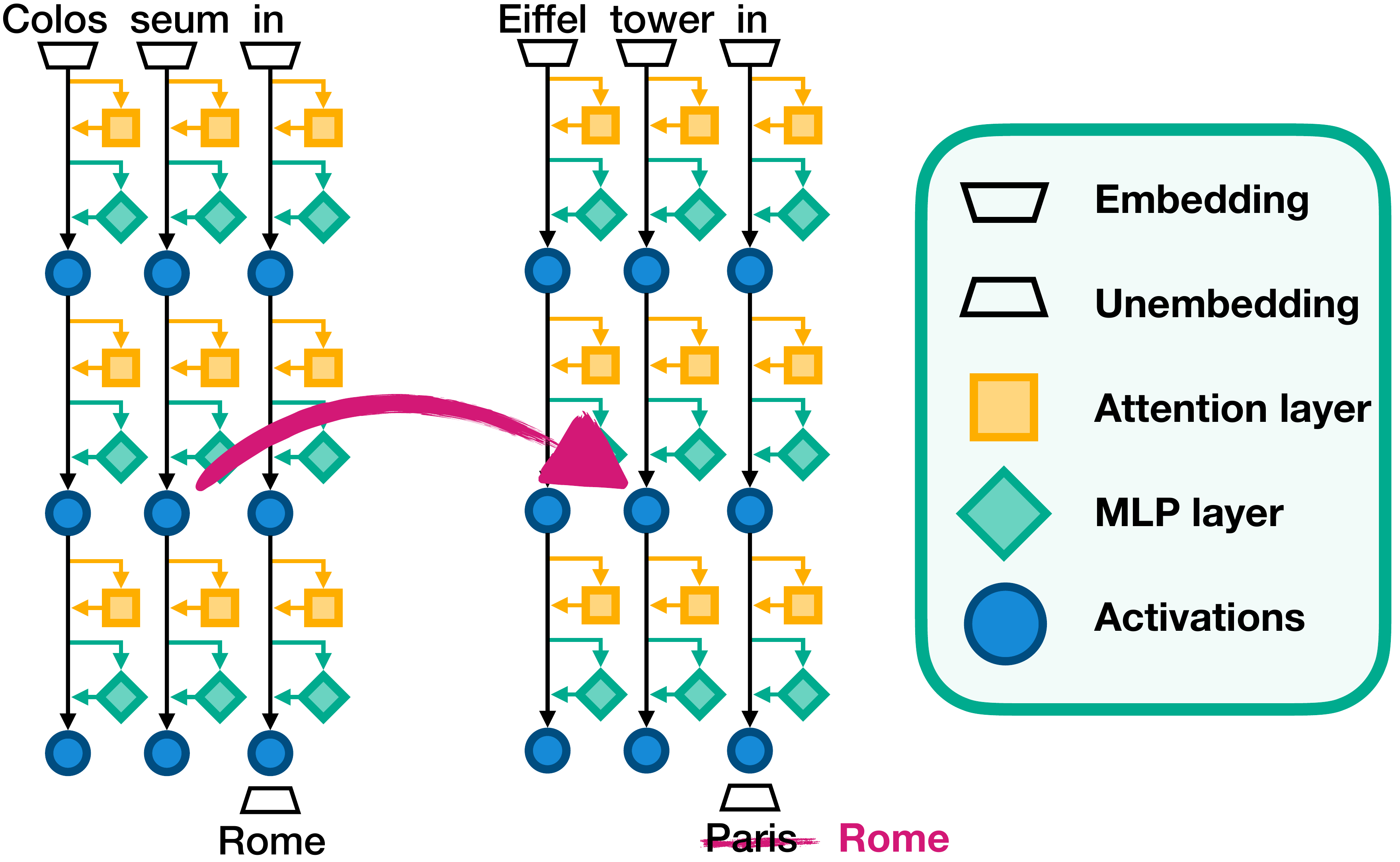}
\caption{}
\label{fig:patching_process}
\end{subfigure}
\hfill
\begin{subfigure}[b]{0.49\textwidth}
\centering
\includegraphics[width=\textwidth]{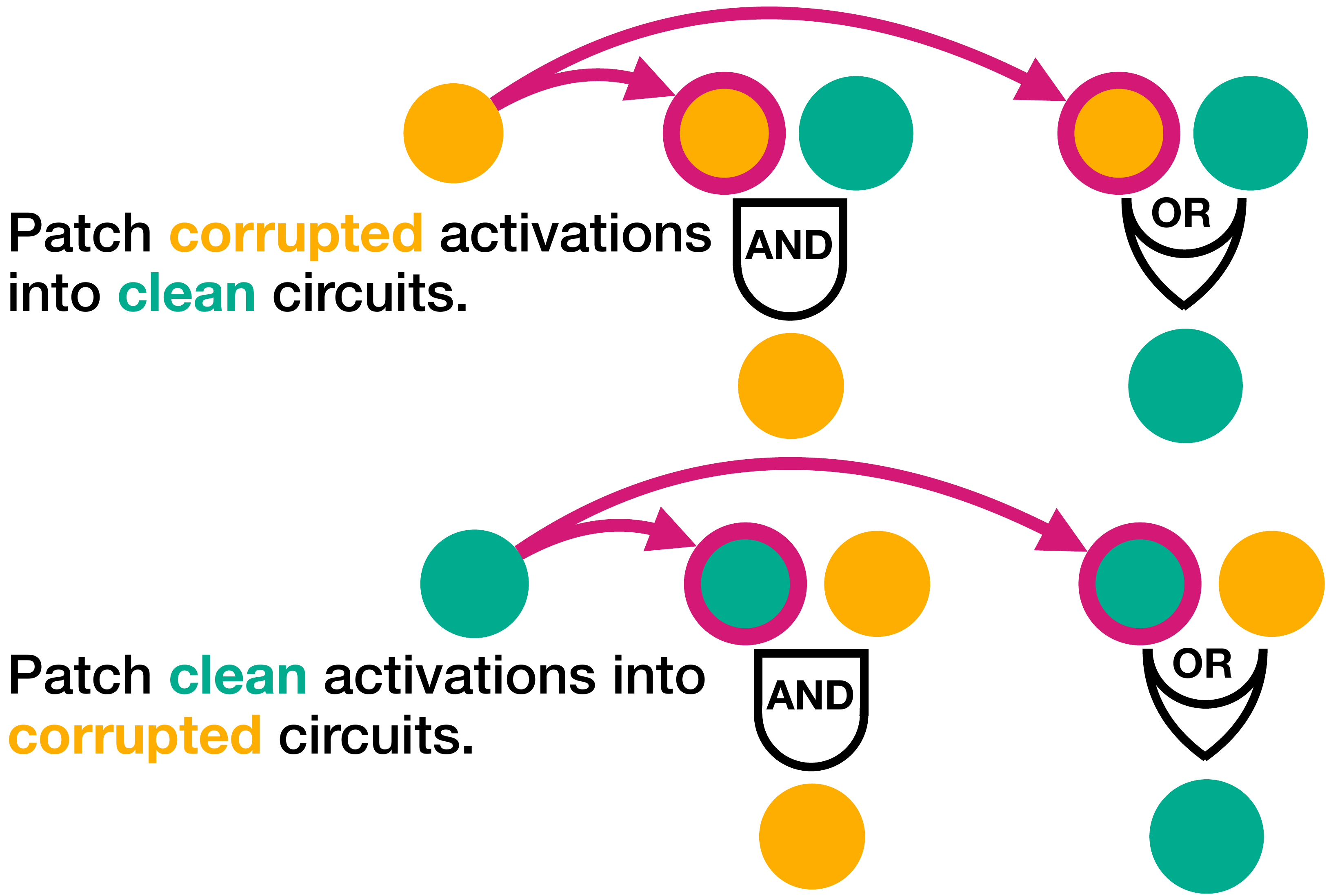}
\caption{}
\label{fig:boolean_circuits}
\end{subfigure}
\caption{(a) Activation patching in a transformer model. Left: The model processes the clean input "Colosseum in Rome," caching the latent activations (step \textit{i.}). Right: The model runs with the corrupted input "Eiffel Tower in Paris" (step \textit{ii.}). The pink arrow shows an MLP layer activation (green diamond) patched from the clean run into the corrupted run (step \textit{iii.}). This causes the prediction to change from "Paris" to "Rome," demonstrating how the significance of the patched component is determined (step \textit{iv.}). By comparing these carefully selected inputs, researchers can control for confounding circuitry and isolate the specific circuit responsible for the location prediction behavior.
(b) Activation patching directions: Top: Patching corrupted activations (orange) into clean circuits (turquoise) reveals \textit{sufficient} components for identifying OR logic scenarios. Bottom: Patching clean activations (green) into corrupted circuits (orange) reveals \textit{necessary} components that are useful for identifying AND logic scenarios. The AND and OR gates demonstrate how these patching directions uncover different logical relationships between model components.}
\label{fig:combined_patching}
\end{figure}

Activation patching is a collective term for a set of causal intervention techniques that manipulate neural network activations to shed light on the decision-making processes within the model. These techniques, including causal tracing \citep{meng_locating_2022}, interchange intervention \citep{geiger_inducing_2021}, causal mediation analysis \citep{vig_investigating_2020}, and causal ablation \citep{wang_interpretability_2023}, share the common goal of modifying a neural model's internal state by replacing specific activations with alternative values, such as zeros, mean activations across samples, random noise, or activations from a different forward pass (\figref{fig:combined_patching}a).

The primary objective of activation patching is to isolate and understand the role of specific components or circuits within the model by observing how changes in activations affect the model's output. This enables researchers to infer the function and importance of those components. Key applications include localizing behavior by identifying critical activations, such as understanding the storage and processing of factual information \citep{meng_locating_2022, geva_dissecting_2023, goldowsky-dill_localizing_2023, stolfo_mechanistic_2023}, and analyzing component interactions through circuit analysis to identify sub-networks within a model's computation graph that implement specified behaviors \citep{wang_interpretability_2023, hanna_how_2023, lieberum_does_2023,hendel_incontext_2023, geva_dissecting_2023}.

The standard protocol for activation patching (\figref{fig:combined_patching}a) involves: 
\begin{enumerate}[label=step \emph{\roman*.}]
    \item Running the model with a clean input and caching the latent activations;
    \item Executing the model with a corrupted input;
    \item Re-running the model with the corrupted input but substituting specific activations with those from the clean cache; and
    \item Determining significance by observing the variations in the model's output during the third step, thereby highlighting the importance of the replaced components.
\end{enumerate}
This process relies on comparing pairs of inputs: a clean input, which triggers the desired behavior, and a corrupted input, which is identical to the clean one except for critical differences that prevent the behavior. By carefully selecting these inputs, researchers can \textit{control for confounding circuitry} and isolate the specific circuit responsible for the behavior.

Differences in patching direction -- clean to corrupted (causal tracing) versus corrupted to clean (resample ablation) -- provide insights into the sufficiency or necessity of model components for a given behavior. Clean to corrupted patching identifies activations sufficient for restoring clean performance, even if they are unnecessary due to redundancy, which is particularly informative in OR logic scenarios (Figure \ref{fig:combined_patching}b, OR gate). Conversely, corrupted to clean patching determines the necessary activations for clean performance, which is useful in AND logic scenarios (Figure \ref{fig:combined_patching}b, AND gate).

Activation patching can employ corruption methods, including zero-, mean-, random-, or resample ablation, each modulating the model's internal state in distinct ways. Resample ablation stands out for its effectiveness in maintaining consistent model behavior by not changing the data distribution too much \citep{zhang_best_2023}. However, it is essential to be careful when interpreting the patching results, as breaking behavior by taking the model off-distribution is uninteresting for finding the relevant circuit \citep{nanda_how_2023}.

\paragraph{Path Patching and Subspace Activation Patching.} Path patching extends the activation patching approach to multiple edges in the computational graph \citep{wang_interpretability_2023, goldowsky-dill_localizing_2023}, allowing for a more fine-grained analysis of component interactions. For example, path patching can be used to estimate the direct and indirect effects of attention heads on the output logits. Subspace activation patching, also known as distributed interchange interventions \citep{geiger_finding_2023}, aims to intervene only on linear subspaces of the representation space where \term{features} are hypothesized to be encoded, providing a tool for more targeted interventions.

Recently, \citet{ghandeharioun_patchscopes_2024} introduced \textit{patchscopes}, a framework that unifies and extends activation patching techniques: using the model's text generation to explain internal representations, it enables more flexible interventions across various interpretability tasks, improving early layer inspection and allowing for cross-model analysis. 

\paragraph{Limitations and Advancements.} Activation patching has several limitations, including the effort required to design input templates and counterfactual datasets, the need for human inspection to isolate important subgraphs, and potential second-order effects that can complicate the interpretation of results \citep{lange_interpretability_2023} and the \term{hydra effect}  \citep{mcgrath_hydra_2023, rushing_explorations_2024} (see discussion in \secref{par:embedded-models}). Recent advancements aim to address these limitations, such as automated circuit discovery algorithms \citep{conmy_automated_2023}, gradient-based methods for scalable component importance estimation like attribution patching \citep{nanda_attribution_2023, syed_attribution_2023}, and techniques to mitigate self-repair interference during analysis \citep{ferrando_information_2024}.

\subsubsection{Causal Abstraction}\label{sec:methods:abstraction}

Causal abstraction \citep{geiger_causal_2021, geiger_causal_2023} provides a mathematical framework for mechanistic interpretability, treating neural networks and their explanations as causal models. This approach validates explanations through interchange interventions on network activations \citep{jenner_comparison_2023}, unifying various interpretability methods such as LIME \citep{ribeiro_why_2016}, causal effect estimation \citep{feder_causalm_2021}, causal mediation analysis \citep{vig_investigating_2020}, iterated nullspace projection \citep{ravfogel_null_2020}, and circuit-based explanations \citep{geiger_causal_2023}.

To overcome computational limitations, \textit{distributed alignment search} \citep{geiger_finding_2023} introduced gradient-based distributed interchange interventions, extending causal abstraction to larger models \citep{wu_interpretability_2023}. Further advancements include \textit{causal proxy models} \citep{wu_causal_2023}, which address the challenge of counterfactual observations.

Applications of causal abstraction span from linguistic phenomena analysis \citep{arora_causalgym_2024, wu_causal_2022}, and evaluation of interpretability methods \citep{huang_ravel_2024}, to improving performance through representation finetuning \citep{wu_reft_2024}, and improving efficiency via model distillation \citep{wu_causal_2022}. 

\subsubsection{Hypothesis Testing}\label{sec:methods:hypothesis}
In addition to the causal abstraction framework, several methods have been developed for rigorous hypothesis testing about neural network behavior. These methods aim to formalize and empirically validate explanations of how neural networks implement specific behaviors.

\textit{Causal scrubbing} \citep{chan_causal_2022} formalizes hypotheses as a tuple $(\gG, \gI, c)$, where $\gG$ is the model's computational graph, $\gI$ is an interpretable computational graph hypothesized to explain the behavior, and $c$ maps nodes of $\gI$ to nodes of $\gG$. This method replaces activations in $\gG$ with others that should be equivalent according to the hypothesis, measuring performance on the scrubbed model to validate the hypothesis.

\textit{Locally consistent abstractions} \citep{jenner_comparison_2023} offer a more permissive approach, checking the consistency between the neural network and the explanation only one step away from the intervention node. This method forms a middle ground between the strictness of full causal abstraction and the flexibility of causal scrubbing.

These methods form a hierarchy of strictness, with full causal abstractions being the most stringent, followed by locally consistent abstractions and causal scrubbing being the most permissive. This hierarchy highlights trade-offs in choosing stricter or more permissive notions, affecting the ability to find acceptable explanations, generalization, and mechanistic anomaly detection.

\subsection{Integrating Observation and Intervention.}\label{sec:methods:integrate}
To comprehensively understand internal neural network mechanisms, combining observational and interventional methods is crucial. For instance, sparse autoencoders can be used to disentangle superposed features \citep{cunningham_sparse_2024}, followed by targeted activation patching to test the causal importance of these features \citep{wang_interpretability_2023}. Similarly, the logit lens can track prediction formation across layers \citep{nostalgebraist_interpreting_2020}, with subsequent interventions confirming causal relationships at key points. Probing techniques can identify encoded information \citep{belinkov_probing_2021}, which can then be subjected to causal abstraction \citep{geiger_causal_2023} to understand how this information is utilized. This iterative refinement process, where broad observational methods guide targeted interventions and intervention results inform further observations, enables a multi-level analysis that builds a holistic understanding across different levels of abstraction. Recent work \citep{marks_sparse_2024, bushnaq_local_2024, braun_identifying_2024, oneill_sparse_2024, ge_automatically_2024} demonstrates the potential of integrating sparse autoencoders with automated circuits discovery \citep{conmy_automated_2023, syed_attribution_2023}, combining feature-level analysis with circuit-level interventions to uncover the interplay between representation and mechanism.

\section{Current Research}
\label{sec:survey}

This section surveys current research in mechanistic interpretability across three approaches based on when and how the model is interpreted during training: Intrinsic interpretability methods are applied before training to enhance the model's inherent interpretability (\secref{sec:survey:intrinsic}). Developmental interpretability involves studying the model's learning dynamics and the emergence of internal structures during training (\secref{sec:survey:development}). After training, post-hoc interpretability techniques are applied to gain insights into the model's behavior and decision-making processes (\secref{sec:survey:post-hoc}), including efforts towards uncovering general, transferable principles across models and tasks, as well as automating the discovery and interpretation of critical circuits in trained models (\secref{sec:survey:automation}).

\begin{figure}[!htp]
\centering
\includegraphics[width=0.8\linewidth]{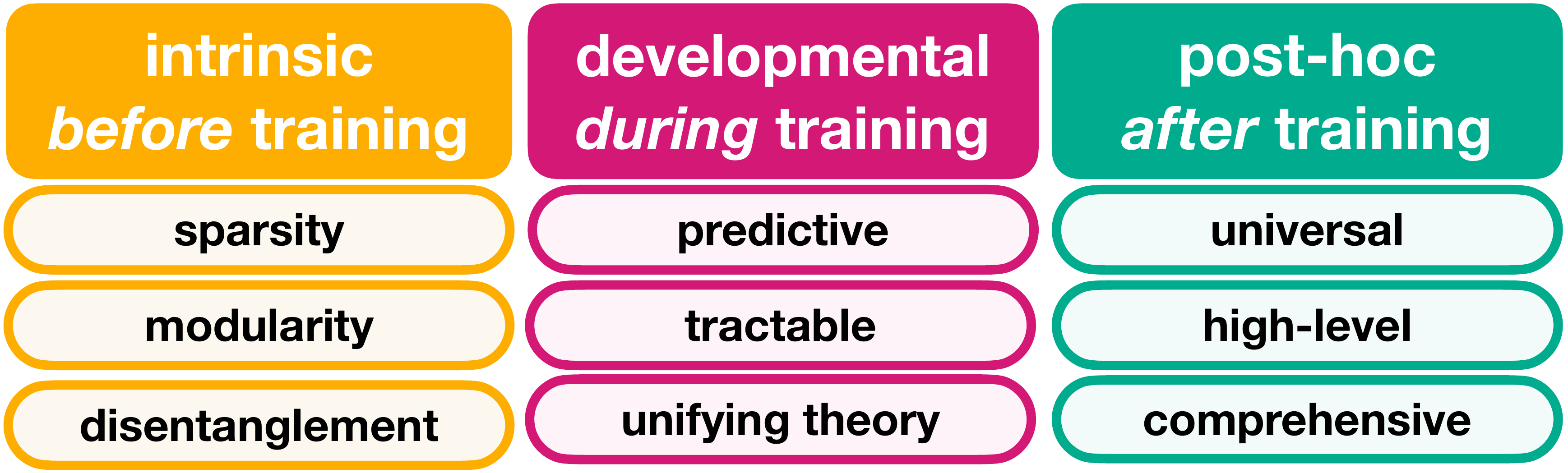}
\caption{Key desiderata for interpretability approaches across training and analysis stages:
(1) Intrinsic: Architectural biases for sparsity, \term{modularity}, and \term{disentangled} representations.
(2) Developmental: Predictive capability for phase transitions, manageable number of critical transitions, and a unifying theory connecting observations to singularity geometry.
(3) Post-hoc: Global, comprehensive, automated discovery of critical circuits, uncovering transferable principles across models/tasks, and extracting high-level causal mechanisms.}
\label{fig:interpretability_desiderata}
\end{figure}

\subsection{Intrinsic Interpretability}\label{sec:survey:intrinsic}

Intrinsic methods for mechanistic interpretability offer a promising approach to designing neural networks that are more amenable to \term{reverse engineering} without sacrificing performance. By encouraging \textit{sparsity}, \term{modularity}, and \textit{monosemanticity} through architectural choices and training procedures, these methods aim to make the reverse engineering process more tractable.

Intrinsic interpretability methods aim to constrain the training process to make learned programs more interpretable \citep{friedman_learning_2023}. This approach is closely related to neurosymbolic learning \citep{riegel_logical_2020} and can involve techniques like regularization with spatial structure, akin to the organization of information in the human brain \citep{liu_seeing_2023, liu_growing_2023}.

Recent work has explored various architectural choices and training procedures to improve the interpretability of neural networks. \citet{jermyn_engineering_2022} and \citet{elhage_softmax_2022} demonstrate that architectural choices can affect monosemanticity, suggesting that models could be engineered to be more \term{monosemantic}. \citet{sharkey_technical_2023} propose using a bilinear layer instead of a linear layer to encourage monosemanticity in language models.

\citet{liu_seeing_2023} and \citet{liu_growing_2023} introduce a biologically inspired spatial regularization regime called brain-inspired modular training for forming modules in networks during training. They showcase how this can help RNNs exhibit brain-like anatomical modularity without degrading performance, in contrast to naive attempts to use sparsity to reduce the cost of having more neurons per layer \citep{jermyn_engineering_2022, bricken_monosemanticity_2023}.

Preceding the mechanistic interpretability literature, various works have explored techniques to improve interpretability, such as sparse attention \citep{zhang_sparse_2021}, adding $L^1$ penalties to neuron activations \citep{kasioumis_elite_2021, georgiadis_accelerating_2019}, and pruning neurons \citep{frankle_lottery_2019}. These techniques have been shown to encourage sparsity, modularity, and {disentanglement}, which are essential aspects of intrinsic interpretability.

\subsection{Developmental Interpretability}
\label{sec:survey:development}

Developmental interpretability examines the learning dynamics and emergence of internal structures in neural networks over time, focusing on the formation of \term{features} and \term{circuits}. This approach complements static analyses by investigating critical phase transitions corresponding to significant changes in model behavior or capabilities \citep{jacob_emergent_2023, schaeffer_are_2023, wei_emergent_2022, simon_stepwise_2023}. While primarily a distinct field, developmental interpretability often intersects with mechanistic interpretability, as exemplified by \citet{olsson_incontext_2022}'s work. Their research, rooted in mechanistic interpretability, demonstrated how the emergence of in-context learning relates to specific training phase transitions, connecting microscopic changes (induction heads) with macroscopic observables (training loss).

A key motivation for developmental interpretability is investigating the \term{universality} of safety-critical patterns, aiming to understand how deeply ingrained and thereby resistant to safety fine-tuning capabilities like deception are. In addition, researchers hypothesize that emergent capabilities correspond to sudden circuit formation during training \citep{michaud_quantization_2023}, potentially allowing for prediction or control of their development. 

Singular Learning Theory (SLT), developed by Watanabe \citep{watanabe_algebraic_2009, watanabe_mathematical_2018}, provides a rigorous framework for understanding overparameterized models' behavior and generalization. By quantifying model complexity through the \textit{local learning coefficient}, SLT offers insights into learning phase transitions and the emergence of structure in the model \citep{lau_quantifying_2023}.  Recent work by \citet{hoogland_stagewise_2024} applied this coefficient to identify developmental stages in transformer models, while \citet{furman_estimating_2024} and \citet{chen_dynamical_2023} advanced SLT's scalability and application to the toy model of \term{superposition} (\figref{fig:toy_model_of_superposition}), respectively.

While direct applications to phenomena such as generalization \citep{zhang_understanding_2017}, learning functions with increasing complexity \citep{nakkiran_sgd_2019}, and the transition from memorization to generalization (\term{grokking}) \citep{liu_understanding_2022, power_grokking_2022, liu_omnigrok_2022, nanda_progress_2023, varma_explaining_2023, thilak_slingshot_2022, merrill_tale_2023, liu_grokking_2023, stander_grokking_2023, wang_grokked_2024} are limited, these areas, along with neural scaling laws \citep{caballero_broken_2022, liu_neural_2023, michaud_quantization_2023} (which can be connected to mechanistic insights \citep{hernandez_scaling_2022}), represent promising future research directions.

In conclusion, developmental interpretability serves as an evolutionary theory lens for neural networks, offering insights into the emergence of structures and behaviors over time \citep{saphra_interpretability_2023}. Drawing parallels from systems biology \citep{alon_introduction_2019}, this approach can apply concepts like network \term{motifs}, robustness, and \term{modularity} to neural network development, explaining how functional capabilities arise. Sometimes, understanding how structures came about is easier than analyzing the final product, similar to how biologists find certain features in organisms easier to explain in light of their evolutionary history. By studying the temporal aspects of neural network training, researchers can potentially uncover fundamental principles of learning and representation that may not be apparent from examining static, trained models alone. 

\subsection{Post-Hoc Interpretability}\label{sec:survey:post-hoc}

In applied mechanistic interpretability, researchers explore various facets and methodologies to uncover the inner workings of AI models. Some key distinctions are drawn between \textit{global} versus \textit{local} interpretability and \textit{comprehensive} versus \textit{partial} interpretability. Global interpretability aims to uncover general patterns and behaviors of a model, providing insights that apply broadly across many instances \citep{doshi-velez_rigorous_2017, nanda_how_2023}. In contrast, local interpretability explains the reasons behind a model's decisions for particular instances, offering insights into individual predictions or behaviors.
Comprehensive interpretability involves achieving a deep and exhaustive understanding of a model's behavior, providing a holistic view of its inner workings \citep{nanda_how_2023}. In contrast, partial interpretability often applied to larger and more complex models, concentrates on interpreting specific aspects or subsets of the model's behavior, focusing on the application's most relevant or critical areas. %Akin to collecting biological species, characterizing these "circuits" aims to discover general computational principles underlying modern AI systems.
%This multifaceted approach collectively analyzes specific capabilities in large models while enabling a comprehensive study of learned algorithms in smaller procedural networks.

\paragraph{Large Models -- Narrow Behavior.}
Circuit-style mechanistic interpretability aims to explain neural networks by \term{reverse engineering} the underlying mechanisms at the level of individual neurons or subgraphs. This approach assumes that neural vector representations encode high-level concepts and circuits defined by model weights encode meaningful algorithms \citep{olah_zoom_2020, cammarata_curve_2020}. Studies on deep networks support these claims, identifying circuits responsible for detecting curved lines or object orientation \citep{cammarata_curve_2020, cammarata_curve_2021, voss_branch_2021}.

This paradigm has been applied to language models to discover subnetworks (circuits) responsible for specific capabilities. Circuit analysis localizes and understands subgraphs within a model's computational graph responsible for specific behaviors. For large language models, this often involves narrow investigations into behaviors like multiple choice reasoning \citep{lieberum_does_2023}, indirect object identification \citep{wang_interpretability_2023}, or computing operations \citep{hanna_how_2023}. Other examples include analyzing circuits for Python docstrings \citep{heimersheim_circuit_2023}, "an" vs "a" usage \citep{miller_we_2023}, and price tagging \citep{wu_interpretability_2023}. Case studies often construct datasets using templates filled by placeholder values to enable precise control for causal interventions \citep{wang_interpretability_2023, hanna_how_2023, wu_interpretability_2023}.

\paragraph{Toy Models -- Comprehensive Analysis.}
Small models trained on specialized mathematical or algorithmic tasks enable more comprehensive reverse engineering of learned algorithms \citep{nanda_progress_2023, zhong_clock_2023, chughtai_toy_2023}. Even simple arithmetic operations can involve complex strategies and multiple algorithmic solutions \citep{nanda_progress_2023, zhong_clock_2023}. Characterizing these algorithms helps test hypotheses around generalizable mechanisms like variable binding \citep{feng_how_2023, davies_discovering_2023} and arithmetic reasoning \citep{stolfo_mechanistic_2023}. The work by \citet{varma_explaining_2023} builds on the work that analyzes transformers trained on modular addition \citep{nanda_progress_2023} and explains \term{grokking} in terms of circuit efficiency, illustrating how a comprehensive understanding of a toy model can enable interesting analyses on top of that understanding.

\paragraph{Towards Universality.}
The ultimate goal is to uncover general principles that transfer across models and tasks, such as induction heads for in-context learning \citep{olsson_incontext_2022}, variable binding mechanisms \citep{feng_how_2023, davies_discovering_2023}, arithmetic reasoning \citep{stolfo_mechanistic_2023, brinkmann_mechanistic_2024}, or retrieval tasks \citep{variengien_look_2023}. Despite promising results, debates surround the \term{universality} hypothesis – the idea that different models learn similar features and circuits when trained on similar tasks. \citep{chughtai_toy_2023} finds mixed evidence for universality in group composition, suggesting that while families of circuits and features can be characterized, precise circuits and development order may be arbitrary.

\paragraph{Towards High-level Mechanisms.}
Causal interventions can extract a high-level understanding of computations and representations learned by large language models \citep{variengien_look_2023, hendel_incontext_2023, feng_how_2023, zou_representation_2023}. Recent work focuses on intervening in internal representations to study high-level concepts and computations encoded. For example, \citet{hendel_incontext_2023} patched residual stream vectors to transfer task representations, while \citet{feng_how_2023} intervened on residual streams to argue that models generate IDs to bind entities to attributes. Techniques for \term{representation engineering} \citep{zou_representation_2023} extract reading vectors from model activations to stimulate or inhibit specific concepts. Although these interventions don't operate via specific mechanisms, they offer a promising approach for extracting high-level causal understanding and bridging bottom-up and top-down interpretability approaches.

\subsection{Automation: Scaling Post-Hoc Interpretability}
\label{sec:survey:automation}

As models become more complex, automating key aspects of the interpretability workflow becomes increasingly crucial. Tracing a model's computational pathways is highly labor-intensive, quickly becoming infeasible as the model size increases. Automating the discovery of relevant circuits and their functional interpretation represents a pivotal step towards scalable and comprehensive model understanding \citep{nainani_evaluating_2024}.

\paragraph{Dissecting Models into Interpretable Circuits.}
The first major automation challenge is identifying the critical computational sub-circuits or components underpinning a model's behavior for a given task. A pioneering line of work aims to achieve this via efficient \textbf{masking} or \textbf{patching} procedures. Methods like \emph{automated circuit discovery} \citep{conmy_automated_2023} and \emph{attribution patching} \citep{syed_attribution_2023, kramar_atp_2024} iteratively knock out model activations, pinpointing components whose removal has the most significant impact on performance. This masking approach has proven scalable even to large models \citep{lieberum_does_2023}.

Other techniques take a more top-down approach. \citet{davies_discovering_2023} specify high-level causal properties (desiderata) that components solving a target subtask should satisfy and then learn binary masks to expose those component subsets. \citet{ferrando_information_2024} construct \textit{information flow graphs} highlighting key nodes and operations by tracing attribution flows, enabling extraction of general information routing patterns across prediction domains.

Explicit architectural biases like modularity can further boost automation efficiency. \citet{nainani_evaluating_2024} find that models trained with \textit{brain-inspired modular training} \citep{liu_seeing_2023} produce more readily identifiable circuits compared to standard training. Such domain-inspired inductive biases may prove increasingly vital as models grow more massive and monolithic.

\paragraph{Interpreting Extracted Circuits.}
Once critical circuit components have been isolated, the key remaining step is interpreting \textit{what} computation those components perform. Sparse autoencoders are a prominent approach for interpreting extracted circuits by decomposing neural network activations into individual component \term{features}, as discussed in \secref{sec:methods:sparse_autoencoders}.

A novel paradigm uses large language models themselves as an interpretive tool. \citet{bills_language_2023} demonstrate generating natural language descriptions of individual neuron functions by prompting language models like GPT-4 to explain sets of inputs that activate a neuron. \citet{mousi_can_2023} similarly employ language models to annotate unsupervised neuron clusters identified via hierarchical clustering. \citet{bai_describeanddissect_2024} describe the roles of neurons in vision networks with multimodal models. These methods can easily leverage more capable general-purpose models in the future. \citet{foote_neuron_2023} take a complementary graph-based approach in their neuron-to-graph tool: automatically extracting individual neurons' behavior patterns from training data as structured graphs amenable to visualization, programmatic comparisons, and property searches. Such representations could synergize with language model-based annotation to provide descriptions of neuron roles. 

However, robustly interpreting the largest trillion-parameter models using automated techniques remains an open challenge. Another novel approach, mechanistic-interpretability-based program synthesis \citep{michaud_opening_2024}, entirely sidesteps this complexity by auto-distilling the algorithm learned by a trained model into human-readable Python code without relying on further interpretability analyses or model architectural knowledge. As models become increasingly vast and opaque, such synergistic combinations of methods -- uncovering circuits, annotating them, or altogether transcribing them into executable code -- will likely prove crucial for maintaining insight and \term{oversight} when scaling model size.

\section{Relevance to AI Safety}\label{sec:relevance}
\paragraph{How Could Interpretability Promote AI Safety?}
\label{sec:relevance:helpful}

\begin{figure}[!htp]
\centering
\includegraphics[width=0.8\linewidth]{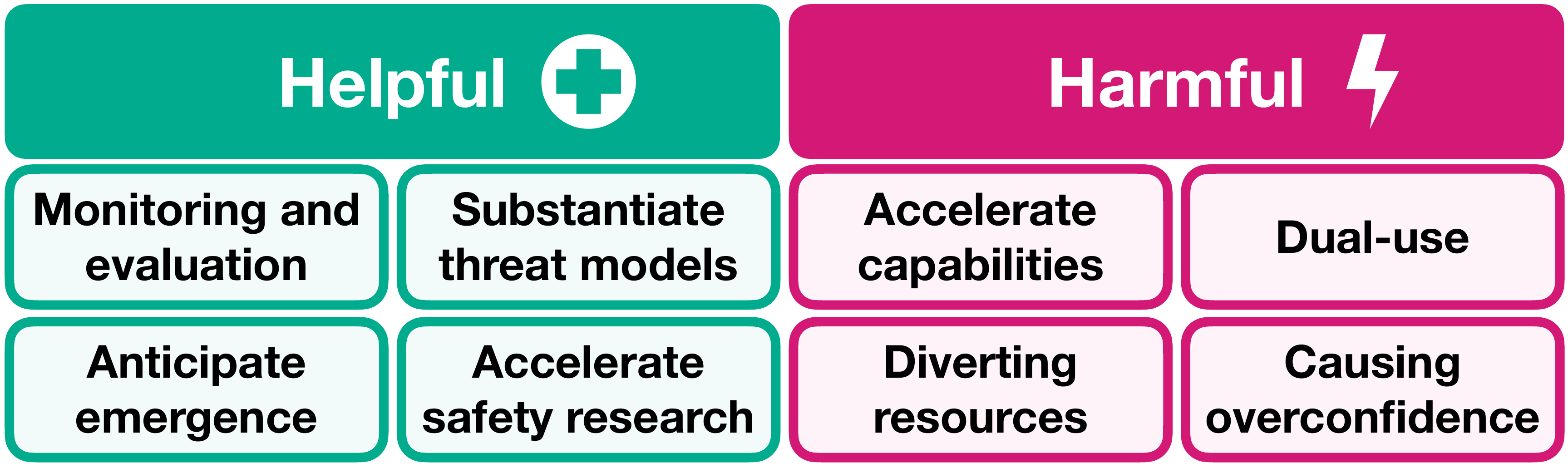}
\caption{Potential benefits and risks of mechanistic interpretability for AI safety.}
\label{fig:relevance}
\end{figure}

Gaining mechanistic insights into the inner workings of AI systems seems crucial for navigating AI safety as we develop more powerful models \citep{nanda_longlist_2022}. Interpretability tools can provide an understanding of artificial cognition, the way AI systems process information and make decisions, which offers several potential benefits:

Mechanistic interpretability could accelerate AI safety research by providing richer feedback loops and grounding for model evaluation \citep{casper_engineer_2023}. It may also help anticipate emergent capabilities, such as the emergence of new skills or behaviors in the model before they fully manifest \citep{wei_emergent_2022, jacob_emergent_2023, nanda_progress_2023, barak_hidden_2022}. This relates to studying the incremental development of internal structures and representations as the model learns (\secref{sec:survey:development}). Additionally, interpretability could substantiate theoretical risk models with concrete evidence, such as demonstrating \term{inner misalignment} (when a model's behavior deviates from its intended goals) or \term{mesa-optimization} (the emergence of unintended subagents within the model) \citep{hubinger_risks_2019, vonoswald_uncovering_2023}. It may also trigger normative shifts within the AI community toward rigorous safety protocols by revealing potential risks or concerning behaviors \citep{hubinger_chris_2019}.

Regarding specific AI risks \citep{hendrycks_overview_2023}, interpretability may prevent malicious misuse by locating and erasing sensitive information stored in the model \citep{meng_locating_2022, nguyen_survey_2022}. It could reduce competitive pressures by substantiating potential threats, promoting organizational safety cultures, and supporting AI alignment (ensuring AI systems pursue intended goals) through better monitoring and evaluation \citep{hendrycks_xrisk_2022}. Interpretability can provide safety filters for every stage of training: before training by deliberate design \citep{hubinger_chris_2019}, during training by detecting early signs of misalignment and potentially shifting the distribution towards alignment \citep{hubinger_transparency_2022, sharkey_circumventing_2022}, and after training by rigorous evaluation of artificial cognition for honesty \citep{burns_discovering_2023, zou_representation_2023} and screening for deceptive behaviors \citep{park_ai_2023}.

The emergence of \term{internal world models} in LLMs, as posited by the \term{simulation} hypothesis, could have significant implications for AI alignment research. Finding an internal representation of human values and aiming the AI system's objective may be a trivial way to achieve alignment \citep{wentworth_how_2022}, especially if the world model is internally separated from notions of goals and agency \citep{ruthenis_internal_2022}. In such cases, world model interpretability alone may be sufficient for alignment \citep{ruthenis_worldmodel_2023}.

% However, it is important to note that current interpretability work primarily focuses on relatively simple world representations, such as those found in game domains like chess \citep{mcgrath_acquisition_2022} or Othello \citep{li_emergent_2023}. These environments are far less complex than the real world, which encompasses nuanced human values and rich cultural information. Scaling interpretability techniques to capture and align with these intricate value systems remains a significant challenge that will require further research and development.

Conditioning pre-trained models is considered a comparatively safe pathway towards general intelligence, as it avoids directly creating agents with inherent goals or agendas \citep{jozdien_conditioning_2022, hubinger_conditioning_2023}. However, prompting a model to simulate an actual agent, such as "You are a superintelligence in 2035 writing down an alignment solution," could inadvertently lead to the formation of internal agents \citep{hubinger_conditioning_2023}. In contrast, reinforcement learning tends to create agents by default \citep{casper_open_2023, ngo_alignment_2022}.

The \term{prediction orthogonality} hypothesis suggests that prediction-focused models like GPT can simulate agents with potentially misaligned objectives \citep{janus_simulators_2022}. Although GPT may lack genuine agency or intentionality, it may produce outputs that simulate these qualities \citep{bereska_taming_2023, shanahan_role_2023}. This underscores the need for careful \term{oversight} and, better yet, using mechanistic interpretability to search for internal agents or their constituents, such as optimization or search processes -- an endeavor known as \emph{searching for search} \citep{nicholaskees_searching_2022, jenner_evidence_2024}.

Mechanistic interpretability integrates well into various AI alignment agendas, such as understanding existing models, controlling them, making AI systems solve alignment problems, and developing alignment theories \citep{technicalities_shallow_2023, hubinger_overview_2020}. It could enhance strategies like detecting \term{deceptive alignment} (hypothetical when a model ensures to appear aligned as to pursue misaligned goals without raising suspicion) \citep{park_ai_2023}, \term{eliciting latent knowledge} from models \citep{christiano_eliciting_2021}, and enabling better scalable \term{oversight}, such as in \term{iterative distillation and amplification} \citep{chan_what_2023}. A high degree of understanding may even allow for \term{well-founded AI} approaches (AI systems with provable guarantees) \citep{tegmark_provably_2023} or \term{microscope AI} (extract world knowledge from the model without letting the model take actions) \citep{hubinger_chris_2019}. Furthermore, comprehensive interpretability itself may be an alignment strategy if we can identify internal representations of human values and guide the model to pursue those values by \textit{retargeting an internal search process} \citep{wentworth_how_2022}. Ultimately, \textit{understanding and control are intertwined}, and deeper understanding can control AI systems more reliably.

However, there is a spectrum of potential misalignment risks, ranging from acute, \textit{model-centric} issues to gradual, \textit{systemic} concerns \citep{kulveit_risks_2024}. While mechanistic interpretability may address risks stemming directly from model internals -- such as deceptive alignment or sudden capability jumps -- it may be less helpful for tackling broader systemic risks like the emergence of misaligned economic structures or novel evolutionary dynamics \citep{hendrycks_natural_2023}. The multi-scale risk landscape calls for a balanced research portfolio to minimize risk, where research on governance, complex systems, and multi-agent simulations complements mechanistic insights and model evaluations. The perceived utility of mechanistic interpretability for AI safety largely depends on researchers' priors regarding the likelihood of these different risk scenarios. 

\paragraph{How Could Mechanistic Insight be Harmful?}
\label{sec:relevance:harmful}

Mechanistic interpretability research could accelerate AI capabilities, potentially leading to the development of powerful AI systems that are misaligned with human values, posing significant risks \citep{soares_if_2023, nicholaskross_why_2023, hendrycks_xrisk_2022}. While historically, interpretability research had little impact on AI capabilities, recent exceptions like discoveries about scaling laws \citep{hoffmann_training_2022}, architectural improvements inspired by studying induction heads \citep{olsson_incontext_2022, fu_hungry_2022, poli_hyena_2023, schuster_confident_2022}, and efficiency gains inspired by the logit lens technique \citep{schuster_confident_2022} demonstrated its potential to enhance capabilities. Scaling interpretability research may necessitate automation \citep{conmy_automated_2023, bills_language_2023}, potentially enabling rapid self-improvement of AI systems \citep{__ricg___agiautomated_2023}. Some researchers recommend selective publication and focusing on lower-risk areas to mitigate these risks \citep{hobbhahn_should_2023, shovelain_riskreward_2023, elhage_toy_2022, nanda_progress_2023}.

Mechanistic interpretability also poses dual-use risks, where the same techniques could be used for both beneficial and harmful purposes. Fine-grained editing capabilities enabled by interpretability could be used for \term{machine unlearning} (removing private data or dangerous knowledge from models) \citep{guo_robust_2024, sun_learning_2024, nguyen_survey_2022, pochinkov_machine_2023} but could be misused for censorship. Similarly, while interpretability may help improve adversarial robustness \citep{rauker_transparent_2023}, it may also facilitate the development of stronger adversarial attacks \citep{mu_compositional_2020, casper_diagnostics_2023}. 

Misunderstanding or overestimating the capabilities of interpretability techniques can divert resources from critical safety areas or lead to overconfidence and misplaced trust in AI systems \citep{charbel-raphael_almost_2023, casper_engineer_2023}. Robust evaluation and benchmarking (\secref{sec:future:standards}) are crucial to validate interpretability claims and reduce the risks of overinterpretation or misinterpretation.

\section{Challenges}\label{sec:challenges}

\subsection{Research Issues}\label{sec:challenges:issues}

\paragraph{Need for Comprehensive, Multi-Pronged Approaches.}
Current interpretability research often focuses on individual techniques rather than combining complementary approaches. To achieve a holistic understanding of neural networks, we propose utilizing a diverse interpretability toolbox that integrates multiple methods (see also \secref{sec:methods:integrate}), such as:
\textit{(i)} Coordinating observational (e.g., probing, logit lens) and interventional methods (e.g., activation patching) to establish causal relationships.
\textit{(ii)} Combining feature-level analysis (e.g., sparse autoencoders) with circuit-level interventions (e.g., path patching) to uncover representation-mechanism interplay.
\textit{(iii)} Integrating intrinsic interpretability approaches with post-hoc analysis for robust understanding. 

For example, coordinated methods could be used for \term{reverse engineering} trojaned behaviors \citep{casper_red_2023}, where observational techniques identify suspicious activations, interventional methods isolate the relevant circuits, and intrinsic approaches guide the design of more robust architectures.

\paragraph{Cherry-Picking and Streetlight Interpretability.} Another concerning pattern is the tendency to cherry-pick results, relying on a small number of convincing examples or visualizations as the basis for an argument without comprehensive evaluation \citep{rauker_transparent_2023}. This amounts to publication bias, showcasing an unrealistic highlight reel of best-case performance. Relatedly, many interpretability techniques are primarily evaluated on small toy models and tasks \citep{chughtai_toy_2023, elhage_toy_2022, jermyn_engineering_2022, chen_dynamical_2023}, risking missing critical phenomena that only emerge in more realistic and diverse contexts. This focus on cherry-picked results from toy models is a form of \term{streetlight interpretability} \citep{casper_engineer_2023}, examining AI systems under only ideal conditions of maximal interpretability.

%\paragraph{Disconnect from Practical Applications and AI Safety.} Perhaps the most vital critique is that mechanistic interpretability work largely fails to connect internal model understanding to real-world utility and AI safety problems \citep{doshi-velez_rigorous_2017, krishnan_against_2020, rauker_transparent_2023}. Despite AI safety being a key motivator, current techniques are seldom applied to key safety problems. The field should prioritize demonstrating clear paths to deploying interpretability to verify and enhance the safety of AI systems.

\subsection{Technical Limitations}\label{sec:challenges:limitations}

\paragraph{Scalability Challenges and Risks of Human Reliance.}
A critical hurdle is demonstrating the scalability of mechanistic interpretability to real-world AI systems across model size, task complexity, behavioral coverage, and analysis efficiency \citep{elhage_toy_2022, scherlis_polysemanticity_2023}. Achieving a truly comprehensive understanding of a model's capabilities in all contexts is daunting, and the time and compute required must scale tractably. Automating interpretability techniques is crucial, as manual analysis quickly becomes infeasible for large models. The high human involvement in current interpretability research raises concerns about the scalability and validity of human-generated model interpretations. Subjective, inconsistent human evaluations and lack of ground-truth benchmarks are known issues \citep{rauker_transparent_2023}. As models scale, it will become increasingly untenable to rely on humans to hypothesize about model mechanisms manually. More work is needed on automating the discovery of mechanistic explanations and translating model weights into human-readable computational graphs \citep{elhage_toy_2022}, but progress on that front may also come from outside the field \citep{lu_ai_2024}.

\paragraph{Obstacles to Bottom-Up Interpretability.} There are fundamental questions about the tractability of fully \term{reverse engineering} neural networks from the bottom up, especially as models become more complex \citep{hendrycks_introduction_2023}. Models may learn internal representations and algorithms that do not cleanly map to human-understandable concepts, making them difficult to interpret even with complete transparency \citep{mcgrath_acquisition_2022}. This gap between human and model ontologies may widen as architectures evolve, increasing opaqueness \citep{hendrycks_unsolved_2022}. Conversely, model representations might naturally converge to more human-interpretable forms as capability increases \citep{hubinger_chris_2019, feng_how_2023}.

\paragraph{Analyzing Models Embedded in Environments.}\label{par:embedded-models}
Real-world AI systems embedded in rich, interactive environments exhibit two forms of in-context behavior that pose significant interpretability challenges beyond understanding models in isolation. Externally, models may dynamically adapt to and reshape their environments through in-context learning from the interactions and feedback loops with their external environment \citep{leahy_barriers_2023}. Internally, the \term{hydra effect} demonstrates in-context reorganization, where models flexibly reorganize their internal representations in a context-dependent manner to maintain capabilities even after ablating key components \citep{mcgrath_hydra_2023}. These two instances of in-context behavior -- external adaptation to the environment and internal self-reorganization -- undermine interpretability approaches that assume fixed \term{circuits}. For models deeply embedded in rich real-world settings, their dynamic coupling with the external world via in-context environmental learning and their internal in-context representational reorganization make strong interpretability guarantees difficult to attain through analysis of the initial model alone.

\paragraph{Adversarial Pressure Against Interpretability.}
As models become more capable through increased training and optimization, there is a risk they may learn deceptive behaviors that actively obscure or mislead the interpretability techniques meant to understand them. Models could develop adversarial "mind-reader" components that predict and counteract the specific analysis methods used to interpret their inner workings \citep{sharkey_circumventing_2022, hubinger_transparency_2022}. Optimizing models through techniques like gradient descent could inadvertently make their internal representations less interpretable to external observers \citep{hubinger_gradient_2019, fu_transformers_2023, vonoswald_uncovering_2023}. In extreme cases, a highly advanced AI system singularly focused on preserving its core objectives may directly undermine the fundamental assumptions that enable interpretability methods in the first place.

These adversarial dynamics, where the capabilities of the AI model are pitted against efforts to interpret it, underscore the need for interpretability research to prioritize worst-case robustness rather than just average-case scenarios. Current techniques often fail even when models are not adversarially optimized. Achieving high confidence in fully understanding extremely capable AI models may require fundamental advances to make interpretability frameworks resilient against an intelligent system's active deceptive efforts.

\section{Future Directions}\label{sec:future}

Given the current limitations and challenges, several key research problems emerge as critical for advancing mechanistic interpretability. These problems span four main areas: emphasizing conceptual clarity \secref{sec:future:clarifying}, establishing rigorous standards \secref{sec:future:standards}, improving the scalability of interpretability techniques \secref{sec:future:scaling}, and expanding the research scope \secref{sec:future:expanding}. Each subsection presents specific research questions and challenges that need to be addressed to move the field forward.

\begin{figure}[!htp]
\centering
\includegraphics[width=0.8\linewidth]{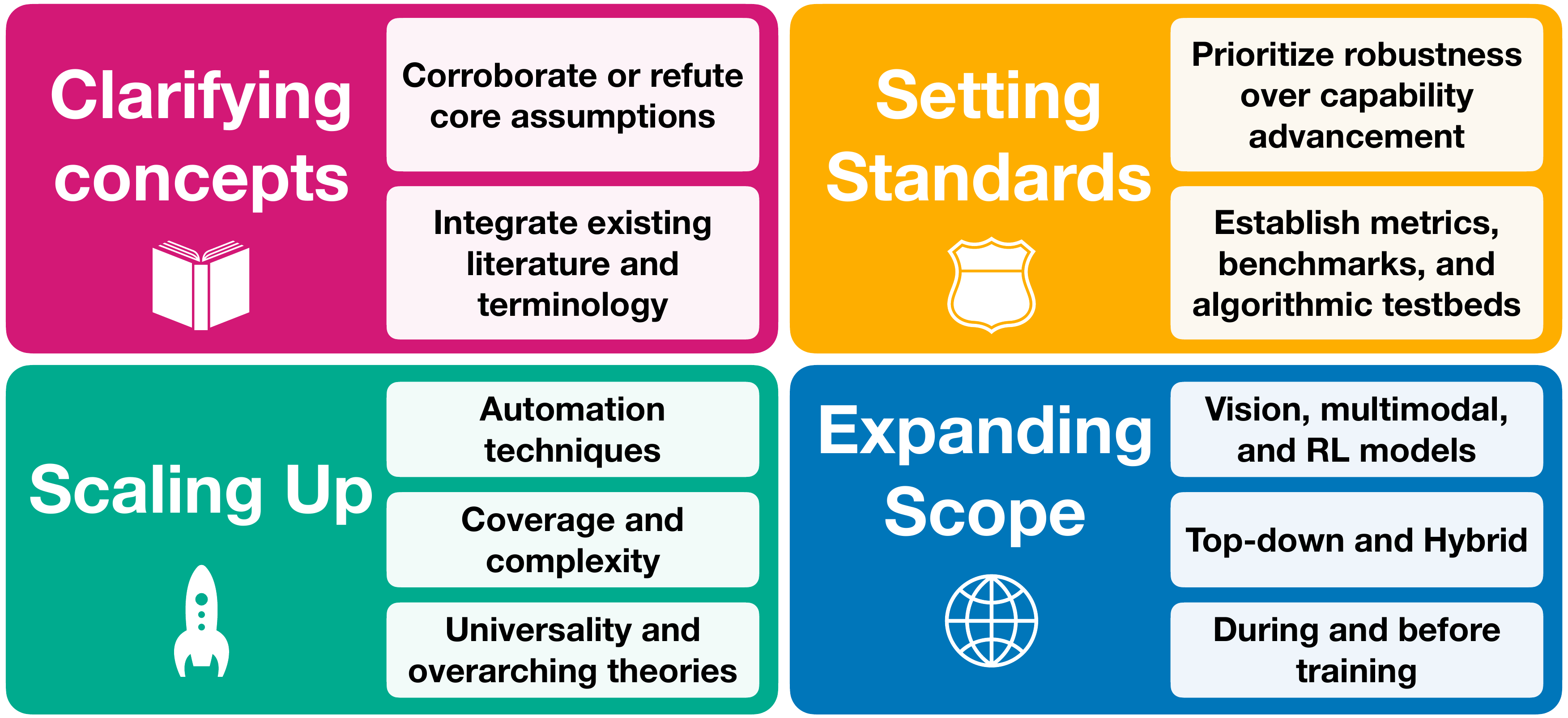}
\caption{Roadmap for advancing mechanistic interpretability research, highlighting key strategic directions.}
\label{fig:future}
\end{figure}

\subsection{Clarifying Concepts}\label{sec:future:clarifying}

\paragraph{Integrating with Existing Literature.}\label{sec:future:clarifying:integrate}
To mature, mechanistic interpretability should embrace existing work, using established terminology rather than reinventing the wheel. Diverging terminology inhibits collaboration across disciplines. Presently, the terminology used for mechanistic interpretability partially diverges from mainstream AI research \citep{casper_engineer_2023}. For example, while the mainstream speaks of \emph{distributed representations} \citep{hinton_distributed_1984, olah_distributed_2023} and the goal of \term{disentangled} representations \citep{higgins_definition_2018, locatello_challenging_2019}, the mechanistic interpretability literature refers to the same phenomenon as \emph{polysemanticity} \citep{scherlis_polysemanticity_2023, lecomte_incidental_2023, marshall_understanding_2024} and \emph{superposition} \citep{elhage_toy_2022, henighan_superposition_2023}. Using common language invites "accidental" contributions and prevents isolating mechanistic interpretability from broader AI research.

Mechanistic interpretability relates to many other fields in AI research, including compressed sensing \citep{elhage_toy_2022}, modularity, adversarial robustness, continual learning, network compression \citep{rauker_transparent_2023}, neurosymbolic reasoning, trojan detection, and program synthesis \citep{casper_engineer_2023, michaud_opening_2024}, and causal representation learning. These relationships can help develop new methods, metrics, benchmarks, and theoretical frameworks. For instance:

\begin{enumerate}[label= \emph{\roman*.)}]
\item \textbf{Neurosymbolic Reasoning and Program Synthesis}: Mechanistic interpretability aims for \term{reverse engineering} neural networks by converting their weights into human-readable algorithms. This endeavor can draw inspiration from neurosymbolic reasoning \citep{riegel_logical_2020} and program synthesis. Techniques like creating programs in domain-specific languages \citep{verma_programmatically_2019, verma_imitation-projected_2019, trivedi_learning_2021}, extracting decision trees \citep{zhang_interpreting_2019} or symbolic causal graphs \citep{ren_defining_2023} from neural networks align well with the goals of mechanistic interpretability. Adopting these approaches can extend the toolkit for reverse engineering AI systems.  

\item \textbf{Causal Representation Learning}: Causal Representation Learning (CRL) aims to discover and disentangle underlying causal factors in data \citep{scholkopf_causal_2021}, complementing mechanistic interpretability's goal of understanding causal structures within neural networks. While mechanistic interpretability typically examines individual \term{features} and \term{circuits}, CRL offers a framework for understanding high-level causal structures. CRL techniques could enhance interpretability by identifying causal relationships between neurons or layers \citep{bengio_metatransfer_2019, ke_systematic_2021}, potentially revealing model reasoning. Its focus on interventions and counterfactuals \citep{pearl_book_2018, peters_elements_2017} could inspire new methods for probing model internals \citep{goyal_recurrent_2020, besserve_counterfactuals_2019}. CRL's emphasis on learning invariant representations \citep{peters_causal_2015, kugelgen_semisupervised_2019} could guide the search for robust features, while its approach to transfer learning \citep{rojas-carulla_invariant_2018, magliacane_domain_2018} could inform studies into model generalization.

\item \textbf{Trojan Detection}: Detecting {deceptive alignment} models is a key motivation for inspecting model internals, as -- by definition -- deception is not salient from observing behavior alone \citep{casper_blackbox_2024}. However, quantifying progress is challenging due to the lack of evidence for deception as an emergent capability in current models \citep{jacob_emergent_2023}, apart from \term{sycophancy} \citep{sharma_understanding_2023, denison_sycophancy_2024} and theoretical evidence for \term{deceptive inflation} behavior \citep{lang_when_2024}. Detecting trojans (or backdoors) \citep{hubinger_sleeper_2024} implanted via data poisoning could be a proxy goal and proof-of-concept. These trojans simulate \term{outer misalignment} (where the model’s behavior is misaligned with the specified reward function or objectives due to poorly defined or incorrect reward signals) rather than \term{inner misalignment} such as deceptive alignment (where the model appears aligned with the specified objectives but internally pursues different, misaligned goals). Moreover, activating a trojan typically results in an immediate change of behavior, while deception can be subtle, gradual, and, at first, entirely internal. Nevertheless, trojan detection can still provide a practical testbed for benchmarking interpretability methods \citep{maloyan_trojan_2024}.

\item \textbf{Adversarial Robustness}: There is a duality between interpretability and adversarial robustness \citep{elhage_toy_2022, rauker_transparent_2023, bereska_mechanistic_2024b}. More interpretable models tend to be more robust against adversarial attacks \citep{jyoti_robustness_2022}, and vice versa, adversarially trained models are often more interpretable \citep{engstrom_adversarial_2019}. For instance, techniques like input gradient regularization have been shown to simultaneously improve the interpretability of saliency maps and enhance adversarial robustness \citep{ross_improving_2017,du_fighting_2021}. Furthermore, interpretability tools can help create more sophisticated adversaries \citep{carter_activation_2019, casper_robust_2021}, improving our understanding of model internals. Viewing adversarial examples as inherent neural network \term{features} \citep{ilyas_adversarial_2019} rather than bugs also hints at alien features beyond human perception. Connecting mechanistic interpretability to adversarial robustness thus promises ways to gain theoretical insight, measure progress \citep{casper_engineer_2023}, design inherently more robust architectures \citep{fort_ensemble_2024}, and create interpretability-guided approaches for identifying (and mitigating) adversarial vulnerabilities \citep{garcia-carrasco_detecting_2024}.
\end{enumerate}

More details on the interplay between interpretability, robustness, modularity, continual learning, network compression, and the human visual system can be found in the review by \citet{rauker_transparent_2023}.

\paragraph{Corroborate or Refute Core Assumptions.}
Features are the fundamental units defining neural representations and enabling mechanistic interpretability's bottom-up approach \citep{chan_what_2023}, but defining them involves assumptions requiring scrutiny, as they shape interpretations and research directions. Questioning hypotheses by seeking additional evidence or counter-examples is crucial.

The \term{linear representation} hypothesis treats activation directions as features \citep{park_linear_2023,nanda_emergent_2023,elhage_toy_2022}, but the emergence and necessity of linearity is unclear -- is it architectural bias or inherent? Stronger theory justifying linearity's necessity or counter-examples like autoencoders on uncorrelated data without intermediate linear layers \citep{elhage_toy_2022} are needed. An alternative lens views features as polytopes from piecewise linear activations \citep{black_interpreting_2022}, questioning if direction simplification suffices or added polytope complexity aids interpretability.

The \term{superposition} hypothesis suggests that \term{polysemantic} neurons arise from the network compressing and representing many features within its limited set of neurons \citep{elhage_toy_2022}, but polysemanticity can also occur incidentally due to redundancy \citep{lecomte_incidental_2023,marshall_understanding_2024,mcgrath_hydra_2023}. Understanding superposition's role could inform mitigating polysemanticity via regularization \citep{lecomte_incidental_2023}. Superposition also raises open questions like operationalizing \textit{computation in superposition} \citep{vaintrob_mathematical_2024, hanni_mathematical_2024}, \textit{attention head superposition} \citep{elhage_toy_2022,jermyn_circuits_2023,lieberum_does_2023,gould_successor_2023}, representing feature clusters \citep{elhage_toy_2022}, connections to adversarial robustness \citep{elhage_toy_2022, garcia-carrasco_detecting_2024, bloom_features_2023}, anti-correlated feature organization \citep{elhage_toy_2022}, and architectural effects \citep{nanda_200superposition_2023}.

\subsection{Setting Standards}\label{sec:future:standards}

\paragraph{Prioritizing Robustness over Capability Advancement.}
As the mechanistic interpretability community expands, it is essential to maintain the norm of not advancing AI capabilities while simultaneously establishing metrics necessary for the field's progress \citep{rauker_transparent_2023}. Researchers should prioritize developing comprehensive tools for analyzing the worst-case performance of AI systems, ensuring robustness and reliability in critical applications. This includes focusing on adversarial tasks, such as backdoor detection and removal \citep{lamparth_analyzing_2023, hubinger_sleeper_2024, wu_backdoorbench_2022}, and evaluating the accuracy of explanations in producing adversarial examples \citep{goldowsky-dill_localizing_2023}.

\paragraph{Establishing Metrics, Benchmarks, and Algorithmic Testbeds.}
A central challenge in mechanistic interpretability is the lack of rigorous evaluation methods. Relying solely on intuition can lead to conflating hypotheses with conclusions, resulting in cherry-picking and optimizing for best-case rather than average or worst-case performance \citep{rudin_stop_2019, miller_explanation_2019, rauker_transparent_2023, casper_engineer_2023}. Current ad hoc practices and proxy measures \citep{doshi-velez_rigorous_2017} risk over-optimization (Goodhart's law -- \textit{When a measure becomes a target, it ceases to be a good measure}). Distinguishing correlation from causation is crucial, as interpretability illusions demonstrate that visualizations may be meaningless without causal linking \citep{bolukbasi_interpretability_2021, friedman_interpretability_2023, olah_feature_2017}.

To advance the field, rigorous evaluation methods are needed. These should include: \textit{(i)} assessing out-of-distribution inputs, as most current methods are only valid for specific examples or datasets \citep{rauker_transparent_2023, ilyas_adversarial_2019, mu_compositional_2020, casper_red_2023, burns_discovering_2023}; \textit{(ii)} controlling systems through edits, such as implanting or removing trojans \citep{mazeika_how_2022} or targeted editing \citep{ghorbani_neuron_2020, dai_knowledge_2022, meng_locating_2022, meng_massediting_2022, bau_gan_2018, hase_does_2023}; \textit{(iii)} replacing components with simpler reverse-engineered alternatives \citep{lindner_tracr_2023}; and \textit{(iv)} comprehensive evaluation through replacing components with hypothesized circuits \citep{quirke_increasing_2024}. 

Algorithmic testbeds are essential for evaluating faithfulness \citep{jacovi_faithfully_2020, hanna_have_2024} and falsifiability \citep{leavitt_falsifiable_2020}. Tools like Tracr \citep{lindner_tracr_2023} can provide ground truth labels for benchmarking search methods \citep{goldowsky-dill_localizing_2023}, while toy models studying superposition in computation \citep{vaintrob_mathematical_2024} and transformers on algorithmic tasks can quantify sparsity and test intrinsic methods. Recently, \citet{thurnherr_tracrbench_2024, gupta_interpbench_2024} introduced datasets of transformer weights with known circuits for evaluating mechanistic interpretability techniques.

\subsection{Scaling Techniques}\label{sec:future:scaling}

\paragraph{Broader and Deeper Coverage of Complex Models and Behaviors.}
A primary goal in scaling mechanistic interpretability is pushing the Pareto frontier between model and task complexity and the coverage of interpretability techniques \citep{chan_what_2023}. While efforts have focused on larger models, it is equally crucial to scale to more complex tasks and provide comprehensive explanations essential for provable safety \citep{tegmark_provably_2023, dalrymple_guaranteed_2024, gross_compact_2024} and enumerative safety \citep{cunningham_sparse_2024,elhage_toy_2022} by ensuring models won't engage in dangerous behaviors like deception. Future work should aim for thorough \term{reverse engineering} \citep{quirke_understanding_2023}, integrating proven modules into larger networks \citep{nanda_progress_2023}, and capturing sequences encoded in hidden states beyond immediate predictions \citep{pal_future_2023}. Deepening analysis complexity is also key, validating the realism of toy models \citep{elhage_toy_2022} and extending techniques like path patching \citep{goldowsky-dill_localizing_2023,liu_seeing_2023} to larger language models. The field must move beyond small transformers on algorithmic tasks \citep{nanda_progress_2023} and limited scenarios \citep{friedman_interpretability_2023} to tackle more complex, realistic cases. 

\paragraph{Towards Universality.}
As mechanistic interpretability matures, the field must transition from isolated empirical findings to developing overarching theories and universal reasoning primitives beyond specific circuits, aiming for a comprehensive understanding of AI capabilities. While collecting empirical data remains valuable \citep{nanda_mechanistic_2023}, establishing motifs, empirical laws, and theories capturing universal model behavior aspects is crucial. This may involve finding more circuits/features \citep{nanda_200circuits_2022,nanda_200features_2022}, exploring circuits as a lens for memorization/generalization \citep{hanna_how_2023}, identifying primitive general reasoning skills \citep{feng_how_2023}, generalizing specific findings to model-agnostic phenomena \citep{merullo_mechanism_2023}, and investigating emergent model generality across neural network classes \citep{ivanitskiy_structured_2023}. Identifying universal reasoning patterns and unifying theories is key to advancing interpretability.

\paragraph{Automation.}
Implementing automated methods is crucial for scaling interpretability of real-world state-of-the-art models across size, task complexity, behavior coverage, and analysis time \citep{hobbhahn_marius_2022}. Manual circuit identification is labor-intensive \citep{lieberum_does_2023}, so automated techniques like circuit discovery and sparse autoencoders can enhance the process \citep{foote_neuron_2023,nanda_200tool_2023}. Future work should automatically create varying datasets for understanding circuit functionality \citep{conmy_automated_2023}, develop automated hypothesis search \citep{goldowsky-dill_localizing_2023}, and investigate attention head/MLP interplay \citep{monea_glitch_2023}. Scaling sparse autoencoders to extract high-quality features automatically for frontier models is critical \citep{bricken_monosemanticity_2023}. Still, it requires caution regarding potential downsides like AI iteration outpacing training \citep{__ricg___agiautomated_2023} and loss of human interpretability from tool complexity \citep{doshi-velez_rigorous_2017}.

\subsection{Expanding Scope}\label{sec:future:expanding}

\paragraph{Interpretability Across Training.} 
While mechanistic interpretability of final trained models is a prerequisite, the field should also advance interpretability before and during training by studying learning dynamics \citep{nanda_200dynamics_2022,elhage_toy_2022,hubinger_transparency_2022}. This includes tracking neuron development \citep{liu_probing_2021}, analyzing neuron set changes with scale \citep{michaud_quantization_2023}, and investigating emergent computations \citep{quirke_understanding_2023}. Studying phase transitions could yield safety insights for \term{reward hacking} risks \citep{olsson_incontext_2022}. 

\paragraph{Multi-Level Analysis.}
Complementing the predominant bottom-up methods \citep{hanna_how_2023}, mechanistic interpretability should explore top-down and hybrid approaches, a promising yet neglected avenue. The top-down analysis offers a tractable way to study large models and guide microscopic research with macroscopic observations \citep{variengien_look_2023}. Its computational efficiency could enable extensive "comparative anatomy" of diverse models, revealing high-level motifs underlying abilities. These motifs could serve as analysis units for understanding internal modifications from techniques like instruction fine-tuning \citep{ouyang_training_2022} and reinforcement learning from human feedback \citep{christiano_deep_2017,bai_training_2022}. 

\paragraph{New Frontiers: Vision, Multimodal, and Reinforcement Learning Models.}
While some mechanistic interpretability has explored convolutional neural networks for vision \citep{cammarata_curve_2021,cammarata_curve_2020}, vision-language models \citep{palit_visionlanguage_2023,salin_are_2022,hilton_understanding_2020}, and multimodal neurons \citep{goh_multimodal_2021}, little work has focused on vision transformers \citep{palit_visionlanguage_2023,aflalo_vl-interpret_2022,vilas_analyzing_2023, pan_dissecting_2024}. Future efforts could identify mechanisms within vision-language models, mirroring progress in unimodal language models \citep{nanda_progress_2023,wang_interpretability_2023}. 

Reinforcement learning (RL) is also a crucial frontier given its role in advanced AI training via techniques like reinforcement learning from human feedback (RLHF) \citep{christiano_deep_2017,bai_training_2022}, despite potentially posing significant safety risks \citep{bereska_taming_2023,casper_open_2023}. Interpretability of RL should investigate reward/goal representations \citep{mini_understanding_2023,colognese_high-level_2023,colognese_internal_2023,bloom_decision_2023, bloom_features_2023}, study circuitry changes from alignment algorithms \citep{prakash_finetuning_2024, jain_mechanistically_2023,lee_mechanistic_2024, jain_what_2024}, and explore emergent subgoals or proxies \citep{hubinger_risks_2019,ivanitskiy_structured_2023} such as internal reward models \citep{marks_training_2023}. While current state-of-the-art AI systems as prediction-trained LLMs are considered relatively safe \citep{hubinger_conditioning_2023}, progress on interpreting RL systems may prove critical for safeguarding the next paradigm \citep{aschenbrenner_situational_2024}. 

\section*{Acknowledgements}
I am grateful for the invaluable feedback and comments from Leon Lang, Tim Bakker, Jannik Brinkmann, Can Rager, Louis van Harten, Jacqueline Bereska, Benjamin Shaffrey, Thijmen Nijdam, Alice Rigg, Arthur Conmy, and Tom Lieberum. Their insights substantially improved this work.

\newpage
\printglossaries

\newpage
\bibliography{review.bib}

\begin{thebibliography}{375}
\providecommand{\natexlab}[1]{#1}
\providecommand{\url}[1]{\texttt{#1}}
\expandafter\ifx\csname urlstyle\endcsname\relax
  \providecommand{\doi}[1]{doi: #1}\else
  \providecommand{\doi}{doi: \begingroup \urlstyle{rm}\Url}\fi

\bibitem[Aflalo et~al.(2022)Aflalo, Du, Tseng, Liu, Wu, Duan, and Lal]{aflalo_vl-interpret_2022}
Estelle Aflalo, Meng Du, Shao-Yen Tseng, Yongfei Liu, Chenfei Wu, Nan Duan, and Vasudev Lal.
\newblock \href{https://ieeexplore.ieee.org/document/9880368/}{Vl-interpret: An interactive visualization tool for interpreting vision-language transformers}.
\newblock \emph{CVPR}, June 2022.

\bibitem[Alain \& Bengio(2016)Alain and Bengio]{alain_understanding_2016}
Guillaume Alain and Yoshua Bengio.
\newblock \href{http://arxiv.org/abs/1610.01644}{Understanding intermediate layers using linear classifier probes}.
\newblock \emph{ICLR}, 2016.

\bibitem[Alon(2019)]{alon_introduction_2019}
Uri Alon.
\newblock \emph{An introduction to systems biology: design principles of biological circuits}.
\newblock {Chapman and Hall/CRC}, 2019.

\bibitem[Arditi et~al.(2024)Arditi, Obeso, Syed, Paleka, Panickssery, Gurnee, and Nanda]{arditi_refusal_2024}
Andy Arditi, Oscar Obeso, Aaquib Syed, Daniel Paleka, Nina Panickssery, Wes Gurnee, and Neel Nanda.
\newblock \href{https://arxiv.org/abs/2406.11717}{Refusal in language models is mediated by a single direction}.
\newblock \emph{CoRR}, 2024.

\bibitem[Arora et~al.(2024)Arora, Jurafsky, and Potts]{arora_causalgym_2024}
Aryaman Arora, Dan Jurafsky, and Christopher Potts.
\newblock \href{http://arxiv.org/abs/2402.12560}{Causalgym: Benchmarking causal interpretability methods on linguistic tasks}.
\newblock \emph{CoRR}, February 2024.

\bibitem[Arora et~al.(2018)Arora, Li, Liang, Ma, and Risteski]{arora_linear_2018}
Sanjeev Arora, Yuanzhi Li, Yingyu Liang, Tengyu Ma, and Andrej Risteski.
\newblock \href{http://arxiv.org/abs/1601.03764}{Linear algebraic structure of word senses, with applications to polysemy}.
\newblock \emph{TACL}, December 2018.

\bibitem[Aschenbrenner(2024)]{aschenbrenner_situational_2024}
Leopold Aschenbrenner.
\newblock \href{https://situational-awareness.ai/}{Situational awareness: The decade ahead}.
\newblock \emph{Series: Situational Awareness. June}, 2024.

\bibitem[Bach et~al.(2015)Bach, Binder, Montavon, Klauschen, M{\"u}ller, and Samek]{bach_pixelwise_2015}
Sebastian Bach, Alexander Binder, Gr{\'e}goire Montavon, Frederick Klauschen, Klaus-Robert M{\"u}ller, and Wojciech Samek.
\newblock \href{https://journals.plos.org/plosone/article?id=10.1371/journal.pone.0130140}{On pixel-wise explanations for non-linear classifier decisions by layer-wise relevance propagation}.
\newblock \emph{PLOS ONE}, July 2015.

\bibitem[Bai et~al.(2024)Bai, Iyer, Oikarinen, and Weng]{bai_describeanddissect_2024}
Nicholas Bai, Rahul~Ajay Iyer, Tuomas Oikarinen, and Tsui-Wei Weng.
\newblock \href{https://openreview.net/forum?id=50SMcZ8QQf}{Describe-and-dissect: Interpreting neurons in vision networks with language models}.
\newblock \emph{ICML MI Workshop}, June 2024.

\bibitem[Bai et~al.(2022)Bai, Jones, Ndousse, Askell, Chen, DasSarma, Drain, Fort, Ganguli, Henighan, Joseph, Kadavath, Kernion, Conerly, {El-Showk}, Elhage, {Hatfield-Dodds}, Hernandez, Hume, Johnston, Kravec, Lovitt, Nanda, Olsson, Amodei, Brown, Clark, McCandlish, Olah, Mann, and Kaplan]{bai_training_2022}
Yuntao Bai, Andy Jones, Kamal Ndousse, Amanda Askell, Anna Chen, Nova DasSarma, Dawn Drain, Stanislav Fort, Deep Ganguli, Tom Henighan, Nicholas Joseph, Saurav Kadavath, Jackson Kernion, Tom Conerly, Sheer {El-Showk}, Nelson Elhage, Zac {Hatfield-Dodds}, Danny Hernandez, Tristan Hume, Scott Johnston, Shauna Kravec, Liane Lovitt, Neel Nanda, Catherine Olsson, Dario Amodei, Tom Brown, Jack Clark, Sam McCandlish, Chris Olah, Ben Mann, and Jared Kaplan.
\newblock \href{http://arxiv.org/abs/2204.05862}{Training a helpful and harmless assistant with reinforcement learning from human feedback}.
\newblock \emph{CoRR}, April 2022.

\bibitem[Bansal et~al.(2021)Bansal, Nakkiran, and Barak]{bansal_revisiting_2021}
Yamini Bansal, Preetum Nakkiran, and Boaz Barak.
\newblock \href{http://arxiv.org/abs/2106.07682}{Revisiting model stitching to compare neural representations}.
\newblock \emph{CoRR}, June 2021.

\bibitem[Barak et~al.(2022)Barak, Edelman, Goel, Kakade, Malach, and Zhang]{barak_hidden_2022}
Boaz Barak, Benjamin~L. Edelman, Surbhi Goel, Sham Kakade, Eran Malach, and Cyril Zhang.
\newblock \href{http://arxiv.org/abs/2207.08799}{Hidden progress in deep learning: Sgd learns parities near the computational limit}.
\newblock \emph{NeurIPS}, 2022.

\bibitem[Bau et~al.(2018)Bau, Zhu, Strobelt, Zhou, Tenenbaum, Freeman, and Torralba]{bau_gan_2018}
David Bau, Jun-Yan Zhu, Hendrik Strobelt, Bolei Zhou, Joshua~B. Tenenbaum, William~T. Freeman, and Antonio Torralba.
\newblock \href{http://arxiv.org/abs/1811.10597}{Gan dissection: Visualizing and understanding generative adversarial networks}.
\newblock \emph{ICLR}, December 2018.

\bibitem[Belinkov(2021)]{belinkov_probing_2021}
Yonatan Belinkov.
\newblock \href{http://arxiv.org/abs/2102.12452}{Probing classifiers: Promises, shortcomings, and advances}.
\newblock \emph{CoRR}, September 2021.

\bibitem[Belrose et~al.(2023)Belrose, Furman, Smith, Halawi, Ostrovsky, McKinney, Biderman, and Steinhardt]{belrose_eliciting_2023}
Nora Belrose, Zach Furman, Logan Smith, Danny Halawi, Igor Ostrovsky, Lev McKinney, Stella Biderman, and Jacob Steinhardt.
\newblock \href{http://arxiv.org/abs/2303.08112}{Eliciting latent predictions from transformers with the tuned lens}.
\newblock \emph{CoRR}, August 2023.

\bibitem[Bender et~al.(2021)Bender, Gebru, {McMillan-Major}, and Shmitchell]{bender_dangers_2021}
Emily~M. Bender, Timnit Gebru, Angelina {McMillan-Major}, and Shmargaret Shmitchell.
\newblock \href{https://dl.acm.org/doi/10.1145/3442188.3445922}{On the dangers of stochastic parrots: Can language models be too big?}
\newblock \emph{ACM FAccT}, March 2021.

\bibitem[Bengio et~al.(2019)Bengio, Deleu, Rahaman, Ke, Lachapelle, Bilaniuk, Goyal, and Pal]{bengio_metatransfer_2019}
Yoshua Bengio, Tristan Deleu, Nasim Rahaman, Rosemary Ke, S{\'e}bastien Lachapelle, Olexa Bilaniuk, Anirudh Goyal, and Christopher Pal.
\newblock \href{http://arxiv.org/abs/1901.10912}{A meta-transfer objective for learning to disentangle causal mechanisms}.
\newblock \emph{CoRR}, February 2019.

\bibitem[Bengio et~al.(2023)Bengio, Hinton, Yao, Song, Abbeel, Harari, Zhang, Xue, {Shalev-Shwartz}, Hadfield, Clune, Maharaj, Hutter, Baydin, McIlraith, Gao, Acharya, Krueger, Dragan, Torr, Russell, Kahneman, Brauner, and Mindermann]{bengio_managing_2023}
Yoshua Bengio, Geoffrey Hinton, Andrew Yao, Dawn Song, Pieter Abbeel, Yuval~Noah Harari, Ya-Qin Zhang, Lan Xue, Shai {Shalev-Shwartz}, Gillian Hadfield, Jeff Clune, Tegan Maharaj, Frank Hutter, At{\i}l{\i}m~G{\"u}ne{\c s} Baydin, Sheila McIlraith, Qiqi Gao, Ashwin Acharya, David Krueger, Anca Dragan, Philip Torr, Stuart Russell, Daniel Kahneman, Jan Brauner, and S{\"o}ren Mindermann.
\newblock \href{http://arxiv.org/abs/2310.17688}{Managing ai risks in an era of rapid progress}.
\newblock \emph{CoRR}, November 2023.

\bibitem[Bereska(2024)]{bereska_mechanistic_2024b}
Leonard Bereska.
\newblock \href{https://leonardbereska.github.io/blog/2024/mechrobustproposal}{Mechanistic interpretability for adversarial robustness --- a proposal}.
\newblock \emph{Leonard Bereska's Blog}, August 2024.

\bibitem[Bereska \& Gavves(2023)Bereska and Gavves]{bereska_taming_2023}
Leonard Bereska and Efstratios Gavves.
\newblock \href{https://ojs.aaai.org/index.php/AAAI-SS/article/view/27478}{Taming simulators: Challenges, pathways and vision for the alignment of large language models}.
\newblock \emph{AAAI-SS}, October 2023.

\bibitem[Besserve et~al.(2019)Besserve, Mehrjou, Sun, and Sch{\"o}lkopf]{besserve_counterfactuals_2019}
Michel Besserve, Arash Mehrjou, R{\'e}my Sun, and Bernhard Sch{\"o}lkopf.
\newblock \href{http://arxiv.org/abs/1812.03253}{Counterfactuals uncover the modular structure of deep generative models}.
\newblock \emph{CoRR}, December 2019.

\bibitem[Bills et~al.(2023)Bills, Cammarata, Mossing, Tillman, Gao, Goh, Sutskever, Leike, Wu, and Saunders]{bills_language_2023}
Steven Bills, Nick Cammarata, Dan Mossing, Henk Tillman, Leo Gao, Gabriel Goh, Ilya Sutskever, Jan Leike, Jeff Wu, and William Saunders.
\newblock \href{https://openaipublic.blob.core.windows.net/neuron-explainer/paper/index.html}{Language models can explain neurons in language models}.
\newblock \emph{OpenAI Blog}, 2023.

\bibitem[Bilodeau et~al.(2024)Bilodeau, Jaques, Koh, and Kim]{bilodeau_impossibility_2024}
Blair Bilodeau, Natasha Jaques, Pang~Wei Koh, and Been Kim.
\newblock \href{http://arxiv.org/abs/2212.11870}{Impossibility theorems for feature attribution}.
\newblock \emph{Proc. Natl. Acad. Sci. U.S.A.}, January 2024.

\bibitem[Bishop(2006)]{bishop_pattern_2006}
Christopher~M. Bishop.
\newblock \href{https://api.semanticscholar.org/CorpusID:60688891}{\emph{Pattern recognition and machine learning}}.
\newblock Springer-Verlag New York Inc., 2006.

\bibitem[Black et~al.(2022)Black, Sharkey, Grinsztajn, Winsor, Braun, Merizian, Parker, Guevara, Millidge, Alfour, and Leahy]{black_interpreting_2022}
Sid Black, Lee Sharkey, Leo Grinsztajn, Eric Winsor, Dan Braun, Jacob Merizian, Kip Parker, Carlos~Ram{\'o}n Guevara, Beren Millidge, Gabriel Alfour, and Connor Leahy.
\newblock \href{https://arxiv.org/abs/2211.12312}{Interpreting neural networks through the polytope lens}.
\newblock \emph{CoRR}, November 2022.

\bibitem[Bloom \& Bailey(2023)Bloom and Bailey]{bloom_features_2023}
Joseph Bloom and Jay Bailey.
\newblock \href{https://www.lesswrong.com/posts/yuQJsRswS4hKv3tsL/features-and-adversaries-in-memorydt}{Features and adversaries in memorydt}.
\newblock \emph{LessWrong}, October 2023.

\bibitem[Bloom \& Colognese(2023)Bloom and Colognese]{bloom_decision_2023}
Joseph Bloom and Paul Colognese.
\newblock \href{https://www.alignmentforum.org/posts/bBuBDJBYHt39Q5zZy/decision-transformer-interpretability}{Decision transformer interpretability}.
\newblock \emph{AI Alignment Forum}, 2023.

\bibitem[Bolukbasi et~al.(2021)Bolukbasi, Pearce, Yuan, Coenen, Reif, Vi'egas, and Wattenberg]{bolukbasi_interpretability_2021}
Tolga Bolukbasi, Adam Pearce, Ann Yuan, Andy Coenen, Emily Reif, Fernanda Vi'egas, and M.~Wattenberg.
\newblock \href{https://www.semanticscholar.org/paper/An-Interpretability-Illusion-for-BERT-Bolukbasi-Pearce/9b9dc2b3d95d2f4e4269a9818c14c70c1f801384}{An interpretability illusion for bert}.
\newblock \emph{CoRR}, April 2021.

\bibitem[Braun et~al.(2024)Braun, Taylor, {Goldowsky-Dill}, and Sharkey]{braun_identifying_2024}
Dan Braun, Jordan Taylor, Nicholas {Goldowsky-Dill}, and Lee Sharkey.
\newblock \href{http://arxiv.org/abs/2405.12241}{Identifying functionally important features with end-to-end sparse dictionary learning}.
\newblock \emph{ICML MI Workshop}, May 2024.

\bibitem[Bricken et~al.(2023)Bricken, Templeton, Batson, Chen, Jermyn, Conerly, Turner, Anil, Denison, Askell, Lasenby, Wu, Kravec, Schiefer, Maxwell, Joseph, Tamkin, Nguyen, McLean, Burke, Hume, Carter, Henighan, and Olah]{bricken_monosemanticity_2023}
Trenton Bricken, Adly Templeton, Joshua Batson, Brian Chen, Adam Jermyn, Tom Conerly, Nicholas~L. Turner, Cem Anil, Carson Denison, Amanda Askell, Robert Lasenby, Yifan Wu, Shauna Kravec, Nicholas Schiefer, Tim Maxwell, Nicholas Joseph, Alex Tamkin, Karina Nguyen, Brayden McLean, Josiah~E. Burke, Tristan Hume, Shan Carter, Tom Henighan, and Chris Olah.
\newblock \href{https://transformer-circuits.pub/2023/monosemantic-features/index.html}{Towards monosemanticity: Decomposing language models with dictionary learning}.
\newblock \emph{Transformer Circuits Thread}, October 2023.

\bibitem[Brinkmann et~al.(2024)Brinkmann, Sheshadri, Levoso, Swoboda, and Bartelt]{brinkmann_mechanistic_2024}
Jannik Brinkmann, Abhay Sheshadri, Victor Levoso, Paul Swoboda, and Christian Bartelt.
\newblock \href{http://arxiv.org/abs/2402.11917}{A mechanistic analysis of a transformer trained on a symbolic multi-step reasoning task}.
\newblock \emph{CoRR}, February 2024.

\bibitem[Bubeck et~al.(2023)Bubeck, Chandrasekaran, Eldan, Gehrke, Horvitz, Kamar, Lee, Lee, Li, Lundberg, Nori, Palangi, Ribeiro, and Zhang]{bubeck_sparks_2023}
S{\'e}bastien Bubeck, Varun Chandrasekaran, Ronen Eldan, Johannes Gehrke, Eric Horvitz, Ece Kamar, Peter Lee, Yin~Tat Lee, Yuanzhi Li, Scott Lundberg, Harsha Nori, Hamid Palangi, Marco~Tulio Ribeiro, and Yi~Zhang.
\newblock \href{http://arxiv.org/abs/2303.12712}{Sparks of artificial general intelligence: Early experiments with gpt-4}.
\newblock \emph{CoRR}, April 2023.

\bibitem[Burns et~al.(2023)Burns, Ye, Klein, and Steinhardt]{burns_discovering_2023}
Collin Burns, Haotian Ye, Dan Klein, and Jacob Steinhardt.
\newblock \href{http://arxiv.org/abs/2212.03827}{Discovering latent knowledge in language models without supervision}.
\newblock \emph{ICLR}, 2023.

\bibitem[Bushnaq et~al.(2024)Bushnaq, Heimersheim, {Goldowsky-Dill}, Braun, Mendel, Hanni, Griffin, Stohler, Wache, and Hobbhahn]{bushnaq_local_2024}
Lucius Bushnaq, Stefan Heimersheim, Nicholas {Goldowsky-Dill}, Dan Braun, Jake Mendel, Kaarel Hanni, Avery Griffin, Jorn Stohler, Magdalena Wache, and Marius Hobbhahn.
\newblock \href{https://arxiv.org/abs/2405.10928}{The local interaction basis: Identifying computationally-relevant and sparsely interacting features in neural networks}.
\newblock \emph{CoRR}, May 2024.

\bibitem[Caballero et~al.(2022)Caballero, Gupta, Rish, and Krueger]{caballero_broken_2022}
Ethan Caballero, Kshitij Gupta, Irina Rish, and David Krueger.
\newblock \href{http://arxiv.org/abs/2210.14891}{Broken neural scaling laws}.
\newblock \emph{ICLR}, October 2022.

\bibitem[Cammarata et~al.(2020)Cammarata, Goh, Carter, Schubert, Petrov, and Olah]{cammarata_curve_2020}
Nick Cammarata, Gabriel Goh, Shan Carter, Ludwig Schubert, Michael Petrov, and Chris Olah.
\newblock \href{https://distill.pub/2020/circuits/curve-detectors}{Curve detectors}.
\newblock \emph{Distill}, June 2020.

\bibitem[Cammarata et~al.(2021)Cammarata, Goh, Carter, Voss, Schubert, and Olah]{cammarata_curve_2021}
Nick Cammarata, Gabriel Goh, Shan Carter, Chelsea Voss, Ludwig Schubert, and Chris Olah.
\newblock \href{https://distill.pub/2020/circuits/curve-circuits/}{Curve circuits}.
\newblock \emph{Distill}, 2021.

\bibitem[Cao et~al.(2021)Cao, Sanh, and Rush]{cao_lowcomplexity_2021}
Steven Cao, Victor Sanh, and Alexander~M. Rush.
\newblock \href{http://arxiv.org/abs/2104.03514}{Low-complexity probing via finding subnetworks}.
\newblock \emph{NAACL-HLT}, April 2021.

\bibitem[Carter et~al.(2019)Carter, Armstrong, Schubert, Johnson, and Olah]{carter_activation_2019}
Shan Carter, Zan Armstrong, Ludwig Schubert, Ian Johnson, and Chris Olah.
\newblock \href{https://distill.pub/2019/activation-atlas}{Activation atlas}.
\newblock \emph{Distill}, March 2019.

\bibitem[Casalicchio et~al.(2018)Casalicchio, Molnar, and Bischl]{casalicchio_visualizing_2018}
Giuseppe Casalicchio, Christoph Molnar, and Bernd Bischl.
\newblock \href{http://arxiv.org/abs/1804.06620}{Visualizing the feature importance for black box models}.
\newblock \emph{ECML PKDD}, 2018.

\bibitem[Casper(2023)]{casper_engineer_2023}
Stephen Casper.
\newblock \href{https://www.alignmentforum.org/s/a6ne2ve5uturEEQK7}{The engineer's interpretability sequence}.
\newblock \emph{AI Alignment Forum}, February 2023.

\bibitem[Casper et~al.(2021)Casper, Nadeau, {Hadfield-Menell}, and Kreiman]{casper_robust_2021}
Stephen Casper, Max Nadeau, Dylan {Hadfield-Menell}, and Gabriel Kreiman.
\newblock \href{http://arxiv.org/abs/2110.03605}{Robust feature-level adversaries are interpretability tools}.
\newblock \emph{NeurIPS}, October 2021.

\bibitem[Casper et~al.(2023{\natexlab{a}})Casper, Davies, Shi, Gilbert, Scheurer, Rando, Freedman, Korbak, Lindner, Freire, Wang, Marks, Segerie, Carroll, Peng, Christoffersen, Damani, Slocum, Anwar, Siththaranjan, Nadeau, Michaud, Pfau, Krasheninnikov, Chen, Langosco, Hase, B{\i}y{\i}k, Dragan, Krueger, Sadigh, and {Hadfield-Menell}]{casper_open_2023}
Stephen Casper, Xander Davies, Claudia Shi, Thomas~Krendl Gilbert, J{\'e}r{\'e}my Scheurer, Javier Rando, Rachel Freedman, Tomasz Korbak, David Lindner, Pedro Freire, Tony Wang, Samuel Marks, Charbel-Rapha{\"e}l Segerie, Micah Carroll, Andi Peng, Phillip Christoffersen, Mehul Damani, Stewart Slocum, Usman Anwar, Anand Siththaranjan, Max Nadeau, Eric~J. Michaud, Jacob Pfau, Dmitrii Krasheninnikov, Xin Chen, Lauro Langosco, Peter Hase, Erdem B{\i}y{\i}k, Anca Dragan, David Krueger, Dorsa Sadigh, and Dylan {Hadfield-Menell}.
\newblock \href{https://arxiv.org/abs/2307.15217}{Open problems and fundamental limitations of reinforcement learning from human feedback}.
\newblock \emph{CoRR}, 2023{\natexlab{a}}.

\bibitem[Casper et~al.(2023{\natexlab{b}})Casper, Hariharan, and {Hadfield-Menell}]{casper_diagnostics_2023}
Stephen Casper, Kaivalya Hariharan, and Dylan {Hadfield-Menell}.
\newblock \href{http://arxiv.org/abs/2211.10024}{Diagnostics for deep neural networks with automated copy/paste attacks}.
\newblock \emph{NeurIPS 2022 ML Safety Workshop (Best paper award)}, May 2023{\natexlab{b}}.

\bibitem[Casper et~al.(2023{\natexlab{c}})Casper, Li, Li, Bu, Zhang, Hariharan, and {Hadfield-Menell}]{casper_red_2023}
Stephen Casper, Yuxiao Li, Jiawei Li, Tong Bu, Kevin Zhang, Kaivalya Hariharan, and Dylan {Hadfield-Menell}.
\newblock \href{https://arxiv.org/abs/2302.10894}{Red teaming deep neural networks with feature synthesis tools}.
\newblock \emph{NeurIPS}, 2023{\natexlab{c}}.

\bibitem[Casper et~al.(2024)Casper, Ezell, Siegmann, Kolt, Curtis, Bucknall, Haupt, Wei, Scheurer, Hobbhahn, Sharkey, Krishna, Von~Hagen, Alberti, Chan, Sun, Gerovitch, Bau, Tegmark, Krueger, and {Hadfield-Menell}]{casper_blackbox_2024}
Stephen Casper, Carson Ezell, Charlotte Siegmann, Noam Kolt, Taylor~Lynn Curtis, Benjamin Bucknall, Andreas Haupt, Kevin Wei, J{\'e}r{\'e}my Scheurer, Marius Hobbhahn, Lee Sharkey, Satyapriya Krishna, Marvin Von~Hagen, Silas Alberti, Alan Chan, Qinyi Sun, Michael Gerovitch, David Bau, Max Tegmark, David Krueger, and Dylan {Hadfield-Menell}.
\newblock \href{http://arxiv.org/abs/2401.14446}{Black-box access is insufficient for rigorous ai audits}.
\newblock \emph{ACM Conference on Fairness, Accountability, and Transparency}, January 2024.

\bibitem[Chan(2023)]{chan_what_2023}
Lawrence Chan.
\newblock \href{https://www.lesswrong.com/posts/6FkWnktH3mjMAxdRT/what-i-would-do-if-i-wasn-t-at-arc-evals}{What i would do if i wasn't at arc evals}.
\newblock \emph{AI Alignment Forum}, May 2023.

\bibitem[Chan et~al.(2022)Chan, {Garriga-alonso}, {Goldowsky-Dill}, {ryan\_greenblatt}, {jenny}, Radhakrishnan, Buck, and Thomas]{chan_causal_2022}
Lawrence Chan, Adri{\`a} {Garriga-alonso}, Nicholas {Goldowsky-Dill}, {ryan\_greenblatt}, {jenny}, Ansh Radhakrishnan, Buck, and Nate Thomas.
\newblock \href{https://www.alignmentforum.org/posts/JvZhhzycHu2Yd57RN/causal-scrubbing-a-method-for-rigorously-testing}{Causal scrubbing: a method for rigorously testing interpretability hypotheses [redwood research]}.
\newblock \emph{AI Alignment Forum}, December 2022.

\bibitem[Chan et~al.(2023)Chan, Lang, and Jenner]{chan_natural_2023}
Lawrence Chan, Leon Lang, and Erik Jenner.
\newblock \href{https://www.alignmentforum.org/posts/gvzW46Z3BsaZsLc25/natural-abstractions-key-claims-theorems-and-critiques-1}{Natural abstractions: Key claims, theorems, and critiques}.
\newblock \emph{AI Alignment Forum}, March 2023.

\bibitem[Chanin et~al.(2023)Chanin, Hunter, and Camburu]{chanin_identifying_2023}
David Chanin, Anthony Hunter, and Oana-Maria Camburu.
\newblock \href{https://arxiv.org/abs/2311.08968}{Identifying linear relational concepts in large language models}.
\newblock \emph{CoRR}, 2023.

\bibitem[{Charbel-Rapha{\"e}l}(2023)]{charbel-raphael_almost_2023}
{Charbel-Rapha{\"e}l}.
\newblock \href{https://www.alignmentforum.org/posts/LNA8mubrByG7SFacm/against-almost-every-theory-of-impact-of-interpretability-1}{Against almost every theory of impact of interpretability}.
\newblock \emph{AI Alignment Forum}, August 2023.

\bibitem[Chen et~al.(2023{\natexlab{a}})Chen, Zhou, and Yan]{chen_going_2023}
Yiting Chen, Zhanpeng Zhou, and Junchi Yan.
\newblock \href{http://arxiv.org/abs/2310.06756}{Going beyond neural network feature similarity: The network feature complexity and its interpretation using category theory}.
\newblock \emph{CoRR}, November 2023{\natexlab{a}}.

\bibitem[Chen et~al.(2023{\natexlab{b}})Chen, Lau, Mendel, Wei, and Murfet]{chen_dynamical_2023}
Zhongtian Chen, Edmund Lau, Jake Mendel, Susan Wei, and Daniel Murfet.
\newblock \href{http://arxiv.org/abs/2310.06301}{Dynamical versus bayesian phase transitions in a toy model of superposition}.
\newblock \emph{CoRR}, October 2023{\natexlab{b}}.

\bibitem[Christiano et~al.(2017)Christiano, Leike, Brown, Martic, Legg, and Amodei]{christiano_deep_2017}
Paul Christiano, Jan Leike, Tom~B. Brown, Miljan Martic, Shane Legg, and Dario Amodei.
\newblock \href{http://arxiv.org/abs/1706.03741}{Deep reinforcement learning from human preferences}.
\newblock \emph{NeurIPS}, December 2017.

\bibitem[Christiano et~al.(2021)Christiano, Cotra, and Xu]{christiano_eliciting_2021}
Paul Christiano, Ajeya Cotra, and Mark Xu.
\newblock \href{https://docs.google.com/document/d/1WwsnJQstPq91_Yh-Ch2XRL8H_EpsnjrC1dwZXR37PC8/edit?usp=sharing&usp=embed_facebook}{Eliciting latent knowledge}, January 2021.

\bibitem[Chughtai et~al.(2023)Chughtai, Chan, and Nanda]{chughtai_toy_2023}
Bilal Chughtai, Lawrence Chan, and Neel Nanda.
\newblock \href{https://arxiv.org/abs/2302.03025}{A toy model of universality: Reverse engineering how networks learn group operations}.
\newblock \emph{ICML}, 2023.

\bibitem[Chughtai et~al.(2024)Chughtai, Cooney, and Nanda]{chughtai_summing_2024}
Bilal Chughtai, Alan Cooney, and Neel Nanda.
\newblock \href{https://arxiv.org/abs/2402.07321}{Summing up the facts: Additive mechanisms behind factual recall in llms}.
\newblock \emph{NeurIPS Workshop Attributing Model Behaviour at Scale}, 2024.

\bibitem[Colognese(2023)]{colognese_internal_2023}
Paul Colognese.
\newblock \href{https://www.lesswrong.com/posts/hhKpXEsfAiyFLecyF/internal-target-information-for-ai-oversight}{Internal target information for ai oversight}.
\newblock \emph{LessWrong}, 2023.

\bibitem[Colognese \& Jozdien(2023)Colognese and Jozdien]{colognese_high-level_2023}
Paul Colognese and Jozdien.
\newblock \href{https://www.lesswrong.com/posts/tFYGdq9ivjA3rdaS2/high-level-interpretability-detecting-an-ai-s-objectives}{High-level interpretability: detecting an ai's objectives}.
\newblock \emph{AI Alignment Forum}, 2023.

\bibitem[Conmy et~al.(2023)Conmy, {Mavor-Parker}, Lynch, Heimersheim, and {Garriga-Alonso}]{conmy_automated_2023}
Arthur Conmy, Augustine~N. {Mavor-Parker}, Aengus Lynch, Stefan Heimersheim, and Adri{\`a} {Garriga-Alonso}.
\newblock \href{https://arxiv.org/abs/2304.14997}{Towards automated circuit discovery for mechanistic interpretability}.
\newblock \emph{NeurIPS}, 2023.

\bibitem[Covert et~al.(2021)Covert, Lundberg, and Lee]{covert_explaining_2021}
Ian~C. Covert, Scott Lundberg, and Su-In Lee.
\newblock \href{https://arxiv.org/abs/2011.14878}{Explaining by removing: a unified framework for model explanation}.
\newblock \emph{J. Mach. Learn. Res.}, January 2021.

\bibitem[Cunningham et~al.(2024)Cunningham, Ewart, Riggs, Huben, and Sharkey]{cunningham_sparse_2024}
Hoagy Cunningham, Aidan Ewart, Logan Riggs, Robert Huben, and Lee Sharkey.
\newblock \href{http://arxiv.org/abs/2309.08600}{Sparse autoencoders find highly interpretable features in language models}.
\newblock \emph{ICLR}, January 2024.

\bibitem[Dai et~al.(2022)Dai, Dong, Hao, Sui, Chang, and Wei]{dai_knowledge_2022}
Damai Dai, Li~Dong, Yaru Hao, Zhifang Sui, Baobao Chang, and Furu Wei.
\newblock \href{https://aclanthology.org/2022.acl-long.581}{Knowledge neurons in pretrained transformers}.
\newblock \emph{ACL}, 2022.

\bibitem[Dalrymple et~al.(2024)Dalrymple, Skalse, Bengio, Russell, Tegmark, Seshia, Omohundro, Szegedy, Goldhaber, Ammann, Abate, Halpern, Barrett, Zhao, {Zhi-Xuan}, Wing, and Tenenbaum]{dalrymple_guaranteed_2024}
David~"davidad" Dalrymple, Joar Skalse, Yoshua Bengio, Stuart Russell, Max Tegmark, Sanjit Seshia, Steve Omohundro, Christian Szegedy, Ben Goldhaber, Nora Ammann, Alessandro Abate, Joe Halpern, Clark Barrett, Ding Zhao, Tan {Zhi-Xuan}, Jeannette Wing, and Joshua Tenenbaum.
\newblock \href{http://arxiv.org/abs/2405.06624}{Towards guaranteed safe ai: A framework for ensuring robust and reliable ai systems}.
\newblock \emph{CoRR}, May 2024.

\bibitem[Dalvi et~al.(2019)Dalvi, Durrani, Sajjad, Belinkov, Bau, and Glass]{dalvi_what_2019}
Fahim Dalvi, Nadir Durrani, Hassan Sajjad, Yonatan Belinkov, Anthony Bau, and James Glass.
\newblock \href{https://ojs.aaai.org/index.php/AAAI/article/view/4592}{What is one grain of sand in the desert? analyzing individual neurons in deep nlp models}.
\newblock \emph{Proceedings of the AAAI Conference on Artificial Intelligence}, July 2019.

\bibitem[Dar et~al.(2022)Dar, Geva, Gupta, and Berant]{dar_analyzing_2022}
Guy Dar, Mor Geva, Ankit Gupta, and Jonathan Berant.
\newblock \href{http://arxiv.org/abs/2209.02535}{Analyzing transformers in embedding space}.
\newblock \emph{ACL}, December 2022.

\bibitem[Davies et~al.(2023)Davies, Nadeau, Prakash, Shaham, and Bau]{davies_discovering_2023}
Xander Davies, Max Nadeau, Nikhil Prakash, Tamar~Rott Shaham, and David Bau.
\newblock \href{http://arxiv.org/abs/2307.03637}{Discovering variable binding circuitry with desiderata}.
\newblock \emph{CoRR}, July 2023.

\bibitem[Deng et~al.(2023)Deng, Tao, and Benton]{deng_measuring_2023}
Mingyang Deng, Lucas Tao, and Joe Benton.
\newblock \href{https://arxiv.org/abs/2310.07837}{Measuring feature sparsity in language models}.
\newblock \emph{CoRR}, 2023.

\bibitem[Denison et~al.(2024)Denison, MacDiarmid, Barez, Duvenaud, Kravec, Marks, Schiefer, Soklaski, Tamkin, Kaplan, Shlegeris, Bowman, Perez, and Hubinger]{denison_sycophancy_2024}
Carson Denison, Monte MacDiarmid, Fazl Barez, David Duvenaud, Shauna Kravec, Samuel Marks, Nicholas Schiefer, Ryan Soklaski, Alex Tamkin, Jared Kaplan, Buck Shlegeris, Samuel~R. Bowman, Ethan Perez, and Evan Hubinger.
\newblock \href{https://arxiv.org/abs/2406.10162}{Sycophancy to subterfuge: Investigating reward-tampering in large language models}.
\newblock \emph{CoRR}, 2024.

\bibitem[Din et~al.(2023)Din, Karidi, Choshen, and Geva]{din_jump_2023}
Alexander~Yom Din, Taelin Karidi, Leshem Choshen, and Mor Geva.
\newblock \href{http://arxiv.org/abs/2303.09435}{Jump to conclusions: Short-cutting transformers with linear transformations}.
\newblock \emph{CoRR}, March 2023.

\bibitem[{Doshi-Velez} \& Kim(2017){Doshi-Velez} and Kim]{doshi-velez_rigorous_2017}
Finale {Doshi-Velez} and Been Kim.
\newblock \href{http://arxiv.org/abs/1702.08608}{Towards a rigorous science of interpretable machine learning}.
\newblock \emph{CoRR}, March 2017.

\bibitem[Du et~al.(2021)Du, Chang, Wen, and Zhang]{du_fighting_2021}
Keke Du, Shan Chang, Huixiang Wen, and Hao Zhang.
\newblock \href{https://doi.org/10.1145/3472634.3472644}{Fighting adversarial images with interpretable gradients}.
\newblock \emph{ACM TURC}, October 2021.

\bibitem[Dunefsky et~al.(2024)Dunefsky, Chlenski, and Nanda]{dunefsky_transcoders_2024}
Jacob Dunefsky, Philippe Chlenski, and Neel Nanda.
\newblock \href{https://openreview.net/forum?id=GWqzUR2dOX}{Transcoders find interpretable llm feature circuits}.
\newblock \emph{ICML MI Workshop}, June 2024.

\bibitem[Durrani et~al.(2020)Durrani, Sajjad, Dalvi, and Belinkov]{durrani_analyzing_2020}
Nadir Durrani, Hassan Sajjad, Fahim Dalvi, and Yonatan Belinkov.
\newblock \href{http://arxiv.org/abs/2010.02695}{Analyzing individual neurons in pre-trained language models}.
\newblock \emph{EMNLP}, October 2020.

\bibitem[Elazar et~al.(2021)Elazar, Ravfogel, Jacovi, and Goldberg]{elazar_amnesic_2021}
Yanai Elazar, Shauli Ravfogel, Alon Jacovi, and Yoav Goldberg.
\newblock \href{http://arxiv.org/abs/2006.00995}{Amnesic probing: Behavioral explanation with amnesic counterfactuals}.
\newblock \emph{TACL}, February 2021.

\bibitem[Elhage et~al.(2022{\natexlab{a}})Elhage, Hume, Catherine, Neel, Henighan, Johnston, ElShowk, Joseph, DasSarma, Mann, Hernandez, Askell, Ndousse, Drain, Chen, Bai, Ganguli, Lovitt, {Hatfield-Dodds}, Kernion, Conerly, Kravec, Fort, Kadavath, Jacobson, {Tran-Johnson}, Kaplan, Clark, Brown, McCandlish, Amodei, and Olah]{elhage_softmax_2022}
Nelson Elhage, Tristan Hume, Olsson Catherine, Nanda Neel, Tom Henighan, Scott Johnston, Sheer ElShowk, Nicholas Joseph, Nova DasSarma, Ben Mann, Danny Hernandez, Amanda Askell, Kamal Ndousse, Dawn Drain, Anna Chen, Yuntao Bai, Deep Ganguli, Liane Lovitt, Zac {Hatfield-Dodds}, Jackson Kernion, Tom Conerly, Shauna Kravec, Stanislav Fort, Saurav Kadavath, Josh Jacobson, Eli {Tran-Johnson}, Jared Kaplan, Jack Clark, Tom Brown, Sam McCandlish, Dario Amodei, and Christopher Olah.
\newblock \href{https://transformer-circuits.pub/2022/solu/index.html}{Softmax linear units}.
\newblock \emph{Transformer Circuits Thread}, 2022{\natexlab{a}}.

\bibitem[Elhage et~al.(2022{\natexlab{b}})Elhage, Hume, Olsson, Schiefer, Henighan, Kravec, {Hatfield-Dodds}, Lasenby, Drain, Chen, et~al.]{elhage_toy_2022}
Nelson Elhage, Tristan Hume, Catherine Olsson, Nicholas Schiefer, Tom Henighan, Shauna Kravec, Zac {Hatfield-Dodds}, Robert Lasenby, Dawn Drain, Carol Chen, et~al.
\newblock \href{https://transformer-circuits.pub/2022/toy_model/index.html}{Toy models of superposition}.
\newblock \emph{Transformer Circuits Thread}, 2022{\natexlab{b}}.

\bibitem[Engels et~al.(2024)Engels, Liao, Michaud, Gurnee, and Tegmark]{engels_not_2024}
Joshua Engels, Isaac Liao, Eric~J. Michaud, Wes Gurnee, and Max Tegmark.
\newblock \href{http://arxiv.org/abs/2405.14860}{Not all language model features are linear}.
\newblock \emph{CoRR}, May 2024.

\bibitem[Engstrom et~al.(2019)Engstrom, Ilyas, Santurkar, Tsipras, Tran, and Madry]{engstrom_adversarial_2019}
Logan Engstrom, Andrew Ilyas, Shibani Santurkar, Dimitris Tsipras, Brandon Tran, and Aleksander Madry.
\newblock \href{http://arxiv.org/abs/1906.00945}{Adversarial robustness as a prior for learned representations}.
\newblock \emph{CoRR}, September 2019.

\bibitem[Falkum \& Vicente(2015)Falkum and Vicente]{falkum_polysemy_2015}
Ingrid~Lossius Falkum and Agustin Vicente.
\newblock \href{https://linkinghub.elsevier.com/retrieve/pii/S0024384115000170}{Polysemy: Current perspectives and approaches}.
\newblock \emph{Lingua}, April 2015.

\bibitem[Farquhar et~al.(2023)Farquhar, Varma, Kenton, Gasteiger, Mikulik, and Shah]{farquhar_challenges_2023}
Sebastian Farquhar, Vikrant Varma, Zachary Kenton, Johannes Gasteiger, Vladimir Mikulik, and Rohin Shah.
\newblock \href{https://arxiv.org/abs/2312.10029}{Challenges with unsupervised llm knowledge discovery}.
\newblock \emph{CoRR}, 2023.

\bibitem[Feder et~al.(2021)Feder, Oved, Shalit, and Reichart]{feder_causalm_2021}
Amir Feder, Nadav Oved, Uri Shalit, and Roi Reichart.
\newblock \href{http://arxiv.org/abs/2005.13407}{Causalm: Causal model explanation through counterfactual language models}.
\newblock \emph{Computational Linguistics}, May 2021.

\bibitem[Feng \& Steinhardt(2023)Feng and Steinhardt]{feng_how_2023}
Jiahai Feng and Jacob Steinhardt.
\newblock \href{http://arxiv.org/abs/2310.17191}{How do language models bind entities in context?}
\newblock \emph{CoRR}, October 2023.

\bibitem[Ferrando \& Voita(2024)Ferrando and Voita]{ferrando_information_2024}
Javier Ferrando and Elena Voita.
\newblock \href{http://arxiv.org/abs/2403.00824}{Information flow routes: Automatically interpreting language models at scale}.
\newblock \emph{CoRR}, February 2024.

\bibitem[Ferrando et~al.(2024)Ferrando, Sarti, Bisazza, and {Costa-juss{\`a}}]{ferrando_primer_2024}
Javier Ferrando, Gabriele Sarti, Arianna Bisazza, and Marta~R. {Costa-juss{\`a}}.
\newblock \href{http://arxiv.org/abs/2405.00208}{A primer on the inner workings of transformer-based language models}.
\newblock \emph{CoRR}, May 2024.

\bibitem[Foote et~al.(2023)Foote, Nanda, Kran, Konstas, Cohen, and Barez]{foote_neuron_2023}
Alex Foote, Neel Nanda, Esben Kran, Ioannis Konstas, Shay Cohen, and Fazl Barez.
\newblock \href{http://arxiv.org/abs/2305.19911}{Neuron to graph: Interpreting language model neurons at scale}.
\newblock \emph{CoRR}, May 2023.

\bibitem[Fort \& Lakshminarayanan(2024)Fort and Lakshminarayanan]{fort_ensemble_2024}
Stanislav Fort and Balaji Lakshminarayanan.
\newblock \href{http://arxiv.org/abs/2408.05446}{Ensemble everything everywhere: Multi-scale aggregation for adversarial robustness}.
\newblock \emph{CoRR}, August 2024.

\bibitem[Frankle \& Carbin(2019)Frankle and Carbin]{frankle_lottery_2019}
Jonathan Frankle and Michael Carbin.
\newblock \href{http://arxiv.org/abs/1803.03635}{The lottery ticket hypothesis: Finding sparse, trainable neural networks}.
\newblock \emph{ICLR}, March 2019.

\bibitem[Friedman et~al.(2023{\natexlab{a}})Friedman, Lampinen, Dixon, Chen, and Ghandeharioun]{friedman_interpretability_2023}
Dan Friedman, Andrew Lampinen, Lucas Dixon, Danqi Chen, and Asma Ghandeharioun.
\newblock \href{https://www.semanticscholar.org/paper/e2bc390cf21dc319ea5aa9a7c3a223911dbf2012}{Interpretability illusions in the generalization of simplified models}.
\newblock \emph{CoRR}, 2023{\natexlab{a}}.

\bibitem[Friedman et~al.(2023{\natexlab{b}})Friedman, Wettig, and Chen]{friedman_learning_2023}
Dan Friedman, Alexander Wettig, and Danqi Chen.
\newblock \href{http://arxiv.org/abs/2306.01128}{Learning transformer programs}.
\newblock \emph{NeurIPS}, June 2023{\natexlab{b}}.

\bibitem[Fu et~al.(2023{\natexlab{a}})Fu, Dao, Saab, Thomas, Rudra, and R{\'e}]{fu_hungry_2022}
Daniel~Y. Fu, Tri Dao, Khaled~K. Saab, Armin~W. Thomas, Atri Rudra, and Christopher R{\'e}.
\newblock \href{http://arxiv.org/abs/2212.14052}{Hungry hungry hippos: Towards language modeling with state space models}.
\newblock \emph{ICLR}, 2023{\natexlab{a}}.

\bibitem[Fu et~al.(2023{\natexlab{b}})Fu, Chen, Jia, and Sharan]{fu_transformers_2023}
Deqing Fu, Tian-Qi Chen, Robin Jia, and Vatsal Sharan.
\newblock \href{http://arxiv.org/abs/2310.17086}{Transformers learn higher-order optimization methods for in-context learning: A study with linear models}.
\newblock \emph{CoRR}, October 2023{\natexlab{b}}.

\bibitem[Furman \& Lau(2024)Furman and Lau]{furman_estimating_2024}
Zach Furman and Edmund Lau.
\newblock \href{http://arxiv.org/abs/2402.03698}{Estimating the local learning coefficient at scale}.
\newblock \emph{CoRR}, February 2024.

\bibitem[{Garc{\'i}a-Carrasco} et~al.(2024){Garc{\'i}a-Carrasco}, Mat{\'e}, and Trujillo]{garcia-carrasco_detecting_2024}
Jorge {Garc{\'i}a-Carrasco}, Alejandro Mat{\'e}, and Juan Trujillo.
\newblock \href{http://arxiv.org/abs/2407.19842}{Detecting and understanding vulnerabilities in language models via mechanistic interpretability}.
\newblock \emph{IJCAI}, August 2024.

\bibitem[Garde et~al.(2023)Garde, Kran, and Barez]{garde_deepdecipher_2023}
Albert Garde, Esben Kran, and Fazl Barez.
\newblock \href{http://arxiv.org/abs/2310.01870}{Deepdecipher: Accessing and investigating neuron activation in large language models}.
\newblock \emph{NeurIPS Workshop XAIA}, October 2023.

\bibitem[Gardenfors(2004)]{gardenfors_conceptual_2004}
Peter Gardenfors.
\newblock \emph{Conceptual spaces: The geometry of thought}.
\newblock MIT press, 2004.

\bibitem[Garfinkle \& Hillar(2019)Garfinkle and Hillar]{garfinkle_uniqueness_2019}
Charles~J. Garfinkle and Christopher~J. Hillar.
\newblock \href{https://ieeexplore.ieee.org/abstract/document/8805108}{On the uniqueness and stability of dictionaries for sparse representation of noisy signals}.
\newblock \emph{IEEE Transactions on Signal Processing}, December 2019.

\bibitem[Ge et~al.(2024)Ge, Zhu, Shu, Wang, He, and Qiu]{ge_automatically_2024}
Xuyang Ge, Fukang Zhu, Wentao Shu, Junxuan Wang, Zhengfu He, and Xipeng Qiu.
\newblock \href{https://www.semanticscholar.org/paper/Automatically-Identifying-Local-and-Global-Circuits-Ge-Zhu/9e08a7385a3908ecfaa7886c8597f8c533672ca0}{Automatically identifying local and global circuits with linear computation graphs}.
\newblock \emph{CoRR}, May 2024.

\bibitem[Geiger et~al.(2021{\natexlab{a}})Geiger, Lu, Icard, and Potts]{geiger_causal_2021}
Atticus Geiger, Hanson Lu, Thomas Icard, and Christopher Potts.
\newblock \href{https://proceedings.neurips.cc/paper/2021/hash/4f5c422f4d49a5a807eda27434231040-Abstract.html}{Causal abstractions of neural networks}.
\newblock \emph{NeurIPS}, 2021{\natexlab{a}}.

\bibitem[Geiger et~al.(2021{\natexlab{b}})Geiger, Wu, Lu, Rozner, Kreiss, Icard, Goodman, and Potts]{geiger_inducing_2021}
Atticus Geiger, Zhengxuan Wu, Hanson Lu, Josh Rozner, Elisa Kreiss, Thomas Icard, Noah~D. Goodman, and Christopher Potts.
\newblock \href{http://arxiv.org/abs/2112.00826}{Inducing causal structure for interpretable neural networks}.
\newblock \emph{ICML}, January 2021{\natexlab{b}}.

\bibitem[Geiger et~al.(2023{\natexlab{a}})Geiger, Potts, and Icard]{geiger_causal_2023}
Atticus Geiger, Chris Potts, and Thomas Icard.
\newblock \href{http://arxiv.org/abs/2301.04709}{Causal abstraction for faithful model interpretation}.
\newblock \emph{CoRR}, January 2023{\natexlab{a}}.

\bibitem[Geiger et~al.(2023{\natexlab{b}})Geiger, Wu, Potts, Icard, and Goodman]{geiger_finding_2023}
Atticus Geiger, Zhengxuan Wu, Christopher Potts, Thomas Icard, and Noah~D. Goodman.
\newblock \href{https://arxiv.org/abs/2303.02536}{Finding alignments between interpretable causal variables and distributed neural representations}.
\newblock \emph{CoRR}, 2023{\natexlab{b}}.

\bibitem[Georgiadis(2019)]{georgiadis_accelerating_2019}
Georgios Georgiadis.
\newblock \href{http://arxiv.org/abs/1812.04056}{Accelerating convolutional neural networks via activation map compression}.
\newblock \emph{CoRR}, March 2019.

\bibitem[Geva et~al.(2022)Geva, Caciularu, Wang, and Goldberg]{geva_transformer_2022}
Mor Geva, Avi Caciularu, Kevin~Ro Wang, and Yoav Goldberg.
\newblock \href{http://arxiv.org/abs/2203.14680}{Transformer feed-forward layers build predictions by promoting concepts in the vocabulary space}.
\newblock \emph{EMNLP}, October 2022.

\bibitem[Geva et~al.(2023)Geva, Bastings, Filippova, and Globerson]{geva_dissecting_2023}
Mor Geva, Jasmijn Bastings, Katja Filippova, and Amir Globerson.
\newblock \href{http://arxiv.org/abs/2304.14767}{Dissecting recall of factual associations in auto-regressive language models}.
\newblock \emph{EMNLP}, October 2023.

\bibitem[Ghandeharioun et~al.(2024)Ghandeharioun, Caciularu, Pearce, Dixon, and Geva]{ghandeharioun_patchscopes_2024}
Asma Ghandeharioun, Avi Caciularu, Adam Pearce, Lucas Dixon, and Mor Geva.
\newblock \href{http://arxiv.org/abs/2401.06102}{Patchscopes: A unifying framework for inspecting hidden representations of language models}.
\newblock \emph{CoRR}, January 2024.

\bibitem[Ghorbani \& Zou(2020)Ghorbani and Zou]{ghorbani_neuron_2020}
Amirata Ghorbani and James Zou.
\newblock \href{http://arxiv.org/abs/2002.09815}{Neuron shapley: Discovering the responsible neurons}.
\newblock \emph{NeurIPS}, November 2020.

\bibitem[Goh et~al.(2021)Goh, {\dag}, {\dag}, Carter, Petrov, Schubert, Radford, and Olah]{goh_multimodal_2021}
Gabriel Goh, Nick~Cammarata {\dag}, Chelsea~Voss {\dag}, Shan Carter, Michael Petrov, Ludwig Schubert, Alec Radford, and Chris Olah.
\newblock \href{https://distill.pub/2021/multimodal-neurons}{Multimodal neurons in artificial neural networks}.
\newblock \emph{Distill}, March 2021.

\bibitem[{Goldowsky-Dill} et~al.(2023){Goldowsky-Dill}, MacLeod, Sato, and Arora]{goldowsky-dill_localizing_2023}
Nicholas {Goldowsky-Dill}, Chris MacLeod, Lucas Sato, and Aryaman Arora.
\newblock \href{https://arxiv.org/abs/2304.05969}{Localizing model behavior with path patching}.
\newblock \emph{CoRR}, 2023.

\bibitem[Gorton(2024)]{gorton_missing_2024}
Liv Gorton.
\newblock \href{https://openreview.net/forum?id=IGnoozsfj1}{The missing curve detectors of inceptionv1: Applying sparse autoencoders to inceptionv1 early vision}.
\newblock \emph{ICML MI Workshop}, June 2024.

\bibitem[Gould et~al.(2023)Gould, Ong, Ogden, and Conmy]{gould_successor_2023}
Rhys Gould, Euan Ong, George Ogden, and Arthur Conmy.
\newblock \href{https://arxiv.org/abs/2312.09230}{Successor heads: Recurring, interpretable attention heads in the wild}.
\newblock \emph{CoRR}, 2023.

\bibitem[Goyal et~al.(2020)Goyal, Lamb, Hoffmann, Sodhani, Levine, Bengio, and Sch{\"o}lkopf]{goyal_recurrent_2020}
Anirudh Goyal, Alex Lamb, Jordan Hoffmann, Shagun Sodhani, Sergey Levine, Yoshua Bengio, and Bernhard Sch{\"o}lkopf.
\newblock \href{http://arxiv.org/abs/1909.10893}{Recurrent independent mechanisms}.
\newblock \emph{CoRR}, November 2020.

\bibitem[Gross et~al.(2024)Gross, Agrawal, Kwa, Ong, Yip, Gibson, Noubir, and Chan]{gross_compact_2024}
Jason Gross, Rajashree Agrawal, Thomas Kwa, Euan Ong, Chun~Hei Yip, Alex Gibson, Soufiane Noubir, and Lawrence Chan.
\newblock \href{http://arxiv.org/abs/2406.11779}{Compact proofs of model performance via mechanistic interpretability}.
\newblock \emph{ICML MI Workshop}, June 2024.

\bibitem[Grosse et~al.(2023)Grosse, Bae, Anil, Elhage, Tamkin, Tajdini, Steiner, Li, Durmus, Perez, Hubinger, Luko{\v s}i{\=u}t{\.e}, Nguyen, Joseph, McCandlish, Kaplan, and Bowman]{grosse_studying_2023}
Roger Grosse, Juhan Bae, Cem Anil, Nelson Elhage, Alex Tamkin, Amirhossein Tajdini, Benoit Steiner, Dustin Li, Esin Durmus, Ethan Perez, Evan Hubinger, Kamil{\.e} Luko{\v s}i{\=u}t{\.e}, Karina Nguyen, Nicholas Joseph, Sam McCandlish, Jared Kaplan, and Samuel~R. Bowman.
\newblock \href{http://arxiv.org/abs/2308.03296}{Studying large language model generalization with influence functions}.
\newblock \emph{CoRR}, August 2023.

\bibitem[Guo et~al.(2024)Guo, Syed, Sheshadri, Ewart, and Dziugaite]{guo_robust_2024}
Phillip~Huang Guo, Aaquib Syed, Abhay Sheshadri, Aidan Ewart, and Gintare~Karolina Dziugaite.
\newblock \href{https://openreview.net/forum?id=06pNzrEjnH}{Robust unlearning via mechanistic localizations}.
\newblock \emph{ICML MI Workshop}, June 2024.

\bibitem[Gupta et~al.(2024)Gupta, Arcuschin, Kwa, and {Garriga-Alonso}]{gupta_interpbench_2024}
Rohan Gupta, Iv{\'a}n Arcuschin, Thomas Kwa, and Adri{\`a} {Garriga-Alonso}.
\newblock \href{http://arxiv.org/abs/2407.14494}{Interpbench: Semi-synthetic transformers for evaluating mechanistic interpretability techniques}.
\newblock \emph{CoRR}, July 2024.

\bibitem[Gurnee \& Tegmark(2024)Gurnee and Tegmark]{gurnee_language_2023}
Wes Gurnee and Max Tegmark.
\newblock \href{https://arxiv.org/abs/2310.02207}{Language models represent space and time}.
\newblock \emph{ICLR}, 2024.

\bibitem[Gurnee et~al.(2023)Gurnee, Nanda, Pauly, Harvey, Troitskii, and Bertsimas]{gurnee_finding_2023}
Wes Gurnee, Neel Nanda, Matthew Pauly, Katherine Harvey, Dmitrii Troitskii, and Dimitris Bertsimas.
\newblock \href{https://arxiv.org/abs/2305.01610}{Finding neurons in a haystack: Case studies with sparse probing}.
\newblock \emph{TMLR}, 2023.

\bibitem[Gurnee et~al.(2024)Gurnee, Horsley, Guo, Kheirkhah, Sun, Hathaway, Nanda, and Bertsimas]{gurnee_universal_2024}
Wes Gurnee, Theo Horsley, Zifan~Carl Guo, Tara~Rezaei Kheirkhah, Qinyi Sun, Will Hathaway, Neel Nanda, and Dimitris Bertsimas.
\newblock \href{http://arxiv.org/abs/2401.12181}{Universal neurons in gpt2 language models}.
\newblock \emph{CoRR}, January 2024.

\bibitem[Ha \& Schmidhuber(2018)Ha and Schmidhuber]{ha_recurrent_2018}
David~R. Ha and J.~Schmidhuber.
\newblock \href{https://www.semanticscholar.org/paper/Recurrent-World-Models-Facilitate-Policy-Evolution-Ha-Schmidhuber/41cca0b0a27ba363ca56e7033569aeb1922b0ac9}{Recurrent world models facilitate policy evolution}.
\newblock \emph{NeurIPS}, September 2018.

\bibitem[Hacohen et~al.(2020)Hacohen, Choshen, and Weinshall]{hacohen_let_2019}
Guy Hacohen, Leshem Choshen, and Daphna Weinshall.
\newblock \href{http://arxiv.org/abs/1905.10854}{Let's agree to agree: Neural networks share classification order on real datasets}.
\newblock \emph{ICML}, 2020.

\bibitem[Hanna et~al.(2023)Hanna, Liu, and Variengien]{hanna_how_2023}
Michael Hanna, Ollie Liu, and Alexandre Variengien.
\newblock \href{https://arxiv.org/abs/2305.00586}{How does gpt-2 compute greater-than?: Interpreting mathematical abilities in a pre-trained language model}.
\newblock \emph{NeurIPS}, 2023.

\bibitem[Hanna et~al.(2024)Hanna, Pezzelle, and Belinkov]{hanna_have_2024}
Michael Hanna, Sandro Pezzelle, and Yonatan Belinkov.
\newblock \href{https://openreview.net/forum?id=grXgesr5dT}{Have faith in faithfulness: Going beyond circuit overlap when finding model mechanisms}.
\newblock \emph{ICML MI Workshop}, June 2024.

\bibitem[H{\"a}nni et~al.(2024)H{\"a}nni, Mendel, Vaintrob, and Chan]{hanni_mathematical_2024}
Kaarel H{\"a}nni, Jake Mendel, Dmitry Vaintrob, and Lawrence Chan.
\newblock \href{http://arxiv.org/abs/2408.05451}{Mathematical models of computation in superposition}.
\newblock \emph{ICML MI Workshop}, August 2024.

\bibitem[Hase et~al.(2023)Hase, Bansal, Kim, and Ghandeharioun]{hase_does_2023}
Peter Hase, Mohit Bansal, Been Kim, and Asma Ghandeharioun.
\newblock \href{http://arxiv.org/abs/2301.04213}{Does localization inform editing? surprising differences in causality-based localization vs. knowledge editing in language models}.
\newblock \emph{NeurIPS Spotlight}, January 2023.

\bibitem[He et~al.(2024)He, Ge, Tang, Sun, Cheng, and Qiu]{he_dictionary_2024}
Zhengfu He, Xuyang Ge, Qiong Tang, Tianxiang Sun, Qinyuan Cheng, and Xipeng Qiu.
\newblock \href{https://arxiv.org/abs/2402.12201}{Dictionary learning improves patch-free circuit discovery in mechanistic interpretability: A case study on othello-gpt}.
\newblock \emph{CoRR}, 2024.

\bibitem[Heimersheim \& Jett(2023)Heimersheim and Jett]{heimersheim_circuit_2023}
Stefan Heimersheim and Jett.
\newblock \href{https://www.alignmentforum.org/posts/u6KXXmKFbXfWzoAXn/a-circuit-for-python-docstrings-in-a-4-layer-attention-only}{A circuit for python docstrings in a 4-layer attention-only transformer}.
\newblock \emph{AI Alignment Forum}, February 2023.

\bibitem[Hendel et~al.(2023)Hendel, Geva, and Globerson]{hendel_incontext_2023}
Roee Hendel, Mor Geva, and Amir Globerson.
\newblock \href{http://arxiv.org/abs/2310.15916}{In-context learning creates task vectors}.
\newblock \emph{EMNLP}, October 2023.

\bibitem[Hendrycks(2023{\natexlab{a}})]{hendrycks_introduction_2023}
Dan Hendrycks.
\newblock \href{https://www.aisafetybook.com/}{\emph{Introduction to AI Safety, Ethics, and Society}}.
\newblock Self-published, 2023{\natexlab{a}}.

\bibitem[Hendrycks(2023{\natexlab{b}})]{hendrycks_natural_2023}
Dan Hendrycks.
\newblock \href{http://arxiv.org/abs/2303.16200}{Natural selection favors ais over humans}.
\newblock \emph{CoRR}, July 2023{\natexlab{b}}.

\bibitem[Hendrycks \& Mazeika(2022)Hendrycks and Mazeika]{hendrycks_xrisk_2022}
Dan Hendrycks and Mantas Mazeika.
\newblock \href{https://arxiv.org/abs/2206.05862v7}{X-risk analysis for ai research}.
\newblock \emph{CoRR}, June 2022.

\bibitem[Hendrycks et~al.(2022)Hendrycks, Carlini, Schulman, and Steinhardt]{hendrycks_unsolved_2022}
Dan Hendrycks, Nicholas Carlini, John Schulman, and Jacob Steinhardt.
\newblock \href{http://arxiv.org/abs/2109.13916}{Unsolved problems in ml safety}.
\newblock \emph{CoRR}, June 2022.

\bibitem[Hendrycks et~al.(2023)Hendrycks, Mazeika, and Woodside]{hendrycks_overview_2023}
Dan Hendrycks, Mantas Mazeika, and Thomas Woodside.
\newblock \href{http://arxiv.org/abs/2306.12001}{An overview of catastrophic ai risks}.
\newblock \emph{CoRR}, October 2023.

\bibitem[Henighan et~al.(2023)Henighan, Carter, Hume, Elhage, Lasenby, Fort, Schiefer, and Olah]{henighan_superposition_2023}
Tom Henighan, Shan Carter, Tristan Hume, Nelson Elhage, Robert Lasenby, Stanislav Fort, Nicholas Schiefer, and Christopher Olah.
\newblock \href{https://transformer-circuits.pub/2023/toy-double-descent/index.html}{Superposition, memorization, and double descent}.
\newblock \emph{Transformer Circuits Thread}, 2023.

\bibitem[Hernandez et~al.(2022)Hernandez, Brown, Conerly, DasSarma, Drain, {El-Showk}, Elhage, {Hatfield-Dodds}, Henighan, Hume, Johnston, Mann, Olah, Olsson, Amodei, Joseph, Kaplan, and McCandlish]{hernandez_scaling_2022}
Danny Hernandez, Tom Brown, Tom Conerly, Nova DasSarma, Dawn Drain, Sheer {El-Showk}, Nelson Elhage, Zac {Hatfield-Dodds}, Tom Henighan, Tristan Hume, Scott Johnston, Ben Mann, Chris Olah, Catherine Olsson, Dario Amodei, Nicholas Joseph, Jared Kaplan, and Sam McCandlish.
\newblock \href{https://arxiv.org/abs/2205.10487}{Scaling laws and interpretability of learning from repeated data}.
\newblock \emph{CoRR}, 2022.

\bibitem[Hernandez et~al.(2023)Hernandez, Sharma, Haklay, Meng, Wattenberg, Andreas, Belinkov, and Bau]{hernandez_linearity_2023}
Evan Hernandez, Arnab~Sen Sharma, Tal Haklay, Kevin Meng, Martin Wattenberg, Jacob Andreas, Yonatan Belinkov, and David Bau.
\newblock \href{http://arxiv.org/abs/2308.09124}{Linearity of relation decoding in transformer language models}.
\newblock \emph{CoRR}, August 2023.

\bibitem[Hewitt \& Manning(2019)Hewitt and Manning]{hewitt_structural_2019}
John Hewitt and Christopher~D. Manning.
\newblock \href{https://aclanthology.org/N19-1419}{A structural probe for finding syntax in word representations}.
\newblock \emph{NAACL HLT}, June 2019.

\bibitem[Higgins et~al.(2018)Higgins, Amos, Pfau, Racaniere, Matthey, Rezende, and Lerchner]{higgins_definition_2018}
Irina Higgins, David Amos, David Pfau, Sebastien Racaniere, Loic Matthey, Danilo Rezende, and Alexander Lerchner.
\newblock \href{http://arxiv.org/abs/1812.02230}{Towards a definition of disentangled representations}.
\newblock \emph{CoRR}, December 2018.

\bibitem[Hilton et~al.(2020)Hilton, Cammarata, Carter, Goh, and Olah]{hilton_understanding_2020}
Jacob Hilton, Nick Cammarata, Shan Carter, Gabriel Goh, and Chris Olah.
\newblock \href{https://distill.pub/2020/understanding-rl-vision/}{Understanding rl vision}.
\newblock \emph{Distill}, 2020.

\bibitem[Hinton(1984)]{hinton_distributed_1984}
Geoffrey~E Hinton.
\newblock \href{https://www.cs.toronto.edu/~hinton/absps/pdp3.pdf}{Distributed representations}.
\newblock \emph{Carnegie Mellon University}, 1984.

\bibitem[Hobbhahn(2022)]{hobbhahn_marius_2022}
Marius Hobbhahn.
\newblock \href{https://docs.google.com/document/d/1AyuTphQ31rLHDtpZoEwEPb4fWbZna1H3hGx_YUACxk4}{Marius' alignment agenda}, 2022.

\bibitem[Hobbhahn \& Chan(2023)Hobbhahn and Chan]{hobbhahn_should_2023}
Marius Hobbhahn and Lawrence Chan.
\newblock \href{https://www.alignmentforum.org/posts/iDNEjbdHhjzvLLAmm/should-we-publish-mechanistic-interpretability-research}{Should we publish mechanistic interpretability research?}
\newblock \emph{AI Alignment Forum}, April 2023.

\bibitem[Hoffmann et~al.(2022)Hoffmann, Borgeaud, Mensch, Buchatskaya, Cai, Rutherford, Casas, Hendricks, Welbl, Clark, Hennigan, Noland, Millican, van~den Driessche, Damoc, Guy, Osindero, Simonyan, Elsen, Rae, Vinyals, and Sifre]{hoffmann_training_2022}
Jordan Hoffmann, Sebastian Borgeaud, Arthur Mensch, Elena Buchatskaya, Trevor Cai, Eliza Rutherford, Diego de~Las Casas, Lisa~Anne Hendricks, Johannes Welbl, Aidan Clark, Tom Hennigan, Eric Noland, Katie Millican, George van~den Driessche, Bogdan Damoc, Aurelia Guy, Simon Osindero, Karen Simonyan, Erich Elsen, Jack~W. Rae, Oriol Vinyals, and Laurent Sifre.
\newblock \href{http://arxiv.org/abs/2203.15556}{Training compute-optimal large language models}.
\newblock \emph{CoRR}, March 2022.

\bibitem[Hoogland et~al.(2024)Hoogland, Carroll, and Murfet]{hoogland_stagewise_2024}
Jesse Hoogland, Liam Carroll, and Daniel Murfet.
\newblock \href{https://www.lesswrong.com/posts/Zza9MNA7YtHkzAtit/stagewise-development-in-neural-networks}{Stagewise development in neural networks}.
\newblock \emph{AI Alignment Forum}, March 2024.

\bibitem[Huang et~al.(2023)Huang, Geiger, D'Oosterlinck, Wu, and Potts]{huang_rigorously_2023}
Jing Huang, Atticus Geiger, Karel D'Oosterlinck, Zhengxuan Wu, and Christopher Potts.
\newblock \href{http://arxiv.org/abs/2309.10312}{Rigorously assessing natural language explanations of neurons}.
\newblock \emph{CoRR}, September 2023.

\bibitem[Huang et~al.(2024)Huang, Wu, Potts, Geva, and Geiger]{huang_ravel_2024}
Jing Huang, Zhengxuan Wu, Christopher Potts, Mor Geva, and Atticus Geiger.
\newblock \href{https://arxiv.org/abs/2402.17700}{Ravel: Evaluating interpretability methods on disentangling language model representations}.
\newblock \emph{CoRR}, 2024.

\bibitem[Hubinger(2019{\natexlab{a}})]{hubinger_chris_2019}
Evan Hubinger.
\newblock \href{https://www.alignmentforum.org/posts/X2i9dQQK3gETCyqh2/chris-olah-s-views-on-agi-safety}{Chris olah's views on agi safety}.
\newblock \emph{AI Alignment Forum}, November 2019{\natexlab{a}}.

\bibitem[Hubinger(2019{\natexlab{b}})]{hubinger_gradient_2019}
Evan Hubinger.
\newblock \href{https://www.alignmentforum.org/posts/uXH4r6MmKPedk8rMA/gradient-hacking}{Gradient hacking}.
\newblock \emph{AI Alignment Forum}, October 2019{\natexlab{b}}.

\bibitem[Hubinger(2020)]{hubinger_overview_2020}
Evan Hubinger.
\newblock \href{http://arxiv.org/abs/2012.07532}{An overview of 11 proposals for building safe advanced ai}.
\newblock \emph{CoRR}, December 2020.

\bibitem[Hubinger(2022)]{hubinger_transparency_2022}
Evan Hubinger.
\newblock \href{https://www.alignmentforum.org/posts/nbq2bWLcYmSGup9aF/a-transparency-and-interpretability-tech-tree}{A transparency and interpretability tech tree}.
\newblock \emph{AI Alignment Forum}, June 2022.

\bibitem[Hubinger et~al.(2019)Hubinger, {van Merwijk}, Mikulik, Skalse, and Garrabrant]{hubinger_risks_2019}
Evan Hubinger, Chris {van Merwijk}, Vladimir Mikulik, Joar Skalse, and Scott Garrabrant.
\newblock \href{http://arxiv.org/abs/1906.01820}{Risks from learned optimization in advanced machine learning systems}.
\newblock \emph{CoRR}, May 2019.

\bibitem[Hubinger et~al.(2023)Hubinger, Jermyn, Treutlein, Hudson, and Woolverton]{hubinger_conditioning_2023}
Evan Hubinger, Adam Jermyn, Johannes Treutlein, Rubi Hudson, and Kate Woolverton.
\newblock \href{http://arxiv.org/abs/2302.00805}{Conditioning predictive models: Risks and strategies}.
\newblock \emph{CoRR}, February 2023.

\bibitem[Hubinger et~al.(2024)Hubinger, Denison, Mu, Lambert, Tong, MacDiarmid, Lanham, Ziegler, Maxwell, Cheng, Jermyn, Askell, Radhakrishnan, Anil, Duvenaud, Ganguli, Barez, Clark, Ndousse, Sachan, Sellitto, Sharma, DasSarma, Grosse, Kravec, Bai, Witten, Favaro, Brauner, Karnofsky, Christiano, Bowman, Graham, Kaplan, Mindermann, Greenblatt, Shlegeris, Schiefer, and Perez]{hubinger_sleeper_2024}
Evan Hubinger, Carson Denison, Jesse Mu, Mike Lambert, Meg Tong, Monte MacDiarmid, Tamera Lanham, Daniel~M. Ziegler, Tim Maxwell, Newton Cheng, Adam Jermyn, Amanda Askell, Ansh Radhakrishnan, Cem Anil, David Duvenaud, Deep Ganguli, Fazl Barez, Jack Clark, Kamal Ndousse, Kshitij Sachan, Michael Sellitto, Mrinank Sharma, Nova DasSarma, Roger Grosse, Shauna Kravec, Yuntao Bai, Zachary Witten, Marina Favaro, Jan Brauner, Holden Karnofsky, Paul Christiano, Samuel~R. Bowman, Logan Graham, Jared Kaplan, S{\"o}ren Mindermann, Ryan Greenblatt, Buck Shlegeris, Nicholas Schiefer, and Ethan Perez.
\newblock \href{https://arxiv.org/abs/2401.05566}{Sleeper agents: Training deceptive llms that persist through safety training}.
\newblock \emph{CoRR}, 2024.

\bibitem[Ilyas et~al.(2019)Ilyas, Santurkar, Tsipras, Engstrom, Tran, and Madry]{ilyas_adversarial_2019}
Andrew Ilyas, Shibani Santurkar, Dimitris Tsipras, Logan Engstrom, Brandon Tran, and Aleksander Madry.
\newblock \href{http://arxiv.org/abs/1905.02175}{Adversarial examples are not bugs, they are features}.
\newblock \emph{NeurIPS}, August 2019.

\bibitem[Ivanitskiy et~al.(2023)Ivanitskiy, Spies, Rauker, Corlouer, Mathwin, Quirke, Rager, Shah, Valentine, Behn, Inoue, and Fung]{ivanitskiy_structured_2023}
M.~Ivanitskiy, Alexander~F. Spies, Tilman Rauker, Guillaume Corlouer, Chris Mathwin, Lucia Quirke, Can Rager, Rusheb Shah, Dan Valentine, Cecilia~Diniz Behn, Katsumi Inoue, and Samy~Wu Fung.
\newblock \href{https://arxiv.org/abs/2312.02566}{Structured world representations in maze-solving transformers}.
\newblock \emph{CoRR}, December 2023.

\bibitem[Jacovi \& Goldberg(2020)Jacovi and Goldberg]{jacovi_faithfully_2020}
Alon Jacovi and Yoav Goldberg.
\newblock \href{http://arxiv.org/abs/2004.03685}{Towards faithfully interpretable nlp systems: How should we define and evaluate faithfulness?}
\newblock \emph{CoRR}, April 2020.

\bibitem[Jain et~al.(2023)Jain, Kirk, Lubana, Dick, Tanaka, Grefenstette, Rockt{\"a}schel, and Krueger]{jain_mechanistically_2023}
Samyak Jain, Robert Kirk, Ekdeep~Singh Lubana, Robert~P. Dick, Hidenori Tanaka, Edward Grefenstette, Tim Rockt{\"a}schel, and David~Scott Krueger.
\newblock \href{http://arxiv.org/abs/2311.12786}{Mechanistically analyzing the effects of fine-tuning on procedurally defined tasks}.
\newblock \emph{CoRR}, November 2023.

\bibitem[Jain et~al.(2024)Jain, Lubana, Oksuz, Joy, Torr, Sanyal, and Dokania]{jain_what_2024}
Samyak Jain, Ekdeep~Singh Lubana, Kemal Oksuz, Tom Joy, Philip Torr, Amartya Sanyal, and Puneet~K. Dokania.
\newblock \href{https://openreview.net/forum?id=BS2CbUkJpy}{What makes and breaks safety fine-tuning? a mechanistic study}.
\newblock \emph{ICML MI Workshop}, June 2024.

\bibitem[{janus}(2022)]{janus_simulators_2022}
{janus}.
\newblock \href{https://www.lesswrong.com/posts/vJFdjigzmcXMhNTsx/simulators}{Simulators}.
\newblock \emph{LessWrong}, September 2022.

\bibitem[Jenner et~al.(2023)Jenner, {Garriga-alonso}, and Zverev]{jenner_comparison_2023}
Erik Jenner, Adri{\`a} {Garriga-alonso}, and Egor Zverev.
\newblock \href{https://www.alignmentforum.org/posts/uLMWMeBG3ruoBRhMW/a-comparison-of-causal-scrubbing-causal-abstractions-and}{A comparison of causal scrubbing, causal abstractions, and related methods}.
\newblock \emph{AI Alignment Forum}, June 2023.

\bibitem[Jenner et~al.(2024)Jenner, Kapur, Georgiev, Allen, Emmons, and Russell]{jenner_evidence_2024}
Erik Jenner, Shreyas Kapur, Vasil Georgiev, Cameron Allen, Scott Emmons, and Stuart Russell.
\newblock \href{http://arxiv.org/abs/2406.00877}{Evidence of learned look-ahead in a chess-playing neural network}.
\newblock \emph{CoRR}, June 2024.

\bibitem[Jermyn et~al.(2023)Jermyn, Olah, and Henighan]{jermyn_circuits_2023}
Adam Jermyn, Chris Olah, and T~Henighan.
\newblock \href{https://transformer-circuits.pub/2023/may-update/index.html}{Circuits updates - may 2023: Attention head superposition}.
\newblock \emph{Transformer Circuits Thread}, 2023.

\bibitem[Jermyn et~al.(2022)Jermyn, Schiefer, and Hubinger]{jermyn_engineering_2022}
Adam~S. Jermyn, Nicholas Schiefer, and Evan Hubinger.
\newblock \href{http://arxiv.org/abs/2211.09169}{Engineering monosemanticity in toy models}.
\newblock \emph{CoRR}, November 2022.

\bibitem[Ji et~al.(2024)Ji, Qiu, Chen, Zhang, Lou, Wang, Duan, He, Zhou, Zhang, Zeng, Ng, Dai, Pan, O'Gara, Lei, Xu, Tse, Fu, McAleer, Yang, Wang, Zhu, Guo, and Gao]{ji_ai_2024}
Jiaming Ji, Tianyi Qiu, Boyuan Chen, Borong Zhang, Hantao Lou, Kaile Wang, Yawen Duan, Zhonghao He, Jiayi Zhou, Zhaowei Zhang, Fanzhi Zeng, Kwan~Yee Ng, Juntao Dai, Xuehai Pan, Aidan O'Gara, Yingshan Lei, Hua Xu, Brian Tse, Jie Fu, Stephen McAleer, Yaodong Yang, Yizhou Wang, Song-Chun Zhu, Yike Guo, and Wen Gao.
\newblock \href{http://arxiv.org/abs/2310.19852}{Ai alignment: A comprehensive survey}.
\newblock \emph{CoRR}, January 2024.

\bibitem[Jozdien(2022)]{jozdien_conditioning_2022}
Jozdien.
\newblock \href{https://www.alignmentforum.org/posts/JqnkeqaPseTgxLgEL/conditioning-generative-models-for-alignment}{Conditioning generative models for alignment}.
\newblock \emph{AI Alignment Forum}, July 2022.

\bibitem[Jumelet(2023)]{jumelet_evaluating_2023}
Jaap Jumelet.
\newblock \href{https://cl-illc.github.io/nlp1-2023/resources/slides/NLP1-Lecture-Interpretability.pdf}{Evaluating and interpreting language models}.
\newblock \emph{NLP Lecture}, November 2023.

\bibitem[Jyoti et~al.(2022)Jyoti, Ganesh, Gayala, Tunuguntla, Kamath, and Balasubramanian]{jyoti_robustness_2022}
Amlan Jyoti, Karthik~Balaji Ganesh, Manoj Gayala, Nandita~Lakshmi Tunuguntla, Sandesh Kamath, and Vineeth~N. Balasubramanian.
\newblock \href{http://arxiv.org/abs/2211.04780}{On the robustness of explanations of deep neural network models: A survey}.
\newblock \emph{CoRR}, November 2022.

\bibitem[Karvonen(2024)]{karvonen_emergent_2024}
Adam Karvonen.
\newblock \href{http://arxiv.org/abs/2403.15498}{Emergent world models and latent variable estimation in chess-playing language models}.
\newblock \emph{COLM}, July 2024.

\bibitem[Karvonen et~al.(2024)Karvonen, Wright, Rager, Angell, Brinkmann, Smith, Verdun, Bau, and Marks]{karvonen_measuring_2024}
Adam Karvonen, Benjamin Wright, Can Rager, Rico Angell, Jannik Brinkmann, Logan Smith, C.~M. Verdun, David Bau, and Samuel Marks.
\newblock \href{https://www.semanticscholar.org/paper/Measuring-Progress-in-Dictionary-Learning-for-Model-Karvonen-Wright/b244b80d0a2d36412b56c0156532d9cbeb298ffa}{Measuring progress in dictionary learning for language model interpretability with board game models}.
\newblock \emph{ICML MI Workshop (Oral)}, July 2024.

\bibitem[Kasioumis et~al.(2021)Kasioumis, Townsend, and Inakoshi]{kasioumis_elite_2021}
Theodoros Kasioumis, Joe Townsend, and Hiroya Inakoshi.
\newblock \href{https://ceur-ws.org/Vol-2986/paper6.pdf}{Elite backprop: Training sparse interpretable neurons}.
\newblock \emph{International Workshop on Neuro-Symbolic Learning and Reasoning}, 2021.

\bibitem[Ke et~al.(2021)Ke, Didolkar, Mittal, Goyal, Lajoie, Bauer, Rezende, Bengio, Mozer, and Pal]{ke_systematic_2021}
Nan~Rosemary Ke, Aniket Didolkar, Sarthak Mittal, Anirudh Goyal, Guillaume Lajoie, Stefan Bauer, Danilo Rezende, Yoshua Bengio, Michael Mozer, and Christopher Pal.
\newblock \href{http://arxiv.org/abs/2107.00848}{Systematic evaluation of causal discovery in visual model based reinforcement learning}.
\newblock \emph{CoRR}, July 2021.

\bibitem[Kirchner(2023)]{kirchner_neuroscience_2023}
Jan Kirchner.
\newblock \href{https://www.lesswrong.com/posts/WGFtgFKuLFMvLuET3/jan-s-shortform}{Neuroscience and natural abstractions}.
\newblock \emph{LessWrong}, March 2023.

\bibitem[Kissane et~al.(2024)Kissane, Krzyzanowski, Bloom, Conmy, and Nanda]{kissane_interpreting_2024}
Connor Kissane, Robert Krzyzanowski, Joseph~Isaac Bloom, Arthur Conmy, and Neel Nanda.
\newblock \href{https://openreview.net/forum?id=fewUBDwjji}{Interpreting attention layer outputs with sparse autoencoders}.
\newblock \emph{ICML MI Workshop}, June 2024.

\bibitem[Kornblith et~al.(2019)Kornblith, Norouzi, Lee, and Hinton]{kornblith_similarity_2019}
Simon Kornblith, Mohammad Norouzi, Honglak Lee, and Geoffrey Hinton.
\newblock \href{http://arxiv.org/abs/1905.00414}{Similarity of neural network representations revisited}.
\newblock \emph{ICML}, July 2019.

\bibitem[Kram{\'a}r et~al.(2024)Kram{\'a}r, Lieberum, Shah, and Nanda]{kramar_atp_2024}
J{\'a}nos Kram{\'a}r, Tom Lieberum, Rohin Shah, and Neel Nanda.
\newblock \href{http://arxiv.org/abs/2403.00745}{Atp*: An efficient and scalable method for localizing llm behaviour to components}.
\newblock \emph{CoRR}, March 2024.

\bibitem[Kross(2023)]{nicholaskross_why_2023}
Nicholas Kross.
\newblock \href{https://www.lesswrong.com/posts/uhMRgEXabYbWeLc6T/why-and-when-interpretability-work-is-dangerous}{Why and when interpretability work is dangerous}.
\newblock \emph{LessWrong}, May 2023.

\bibitem[Kulveit(2024)]{kulveit_risks_2024}
Jan Kulveit.
\newblock Risks from ai misalignment at different scales, July 2024.

\bibitem[Kulveit et~al.(2023)Kulveit, {von Stengel}, and Leventov]{kulveit_predictive_2023}
Jan Kulveit, Clem {von Stengel}, and Roman Leventov.
\newblock \href{http://arxiv.org/abs/2311.10215}{Predictive minds: Llms as atypical active inference agents}.
\newblock \emph{CoRR}, November 2023.

\bibitem[Lamparth \& Reuel(2023)Lamparth and Reuel]{lamparth_analyzing_2023}
Max Lamparth and Anka Reuel.
\newblock \href{https://arxiv.org/abs/2302.12461}{Analyzing and editing inner mechanisms of backdoored language models}.
\newblock \emph{CoRR}, 2023.

\bibitem[Lan \& Barez(2023)Lan and Barez]{lan_locating_2023}
Michael Lan and Fazl Barez.
\newblock \href{https://www.semanticscholar.org/paper/Locating-Cross-Task-Sequence-Continuation-Circuits-Lan-Barez/458642a7695013cd128bb3070540970be814e50d?citedSort=relevance&citedPage=2}{Locating cross-task sequence continuation circuits in transformers}.
\newblock \emph{CoRR}, November 2023.

\bibitem[Lang et~al.(2024)Lang, Foote, Russell, Dragan, Jenner, and Emmons]{lang_when_2024}
Leon Lang, Davis Foote, Stuart Russell, Anca Dragan, Erik Jenner, and Scott Emmons.
\newblock \href{http://arxiv.org/abs/2402.17747}{When your ais deceive you: Challenges with partial observability of human evaluators in reward learning}.
\newblock \emph{CoRR}, March 2024.

\bibitem[Lange et~al.(2023)Lange, Makelov, and Nanda]{lange_interpretability_2023}
Georg Lange, Alex Makelov, and Neel Nanda.
\newblock \href{https://www.alignmentforum.org/posts/RFtkRXHebkwxygDe2/an-interpretability-illusion-for-activation-patching-of}{An interpretability illusion for activation patching of arbitrary subspaces}.
\newblock \emph{AI Alignment Forum}, August 2023.

\bibitem[Langedijk et~al.(2023)Langedijk, Mohebbi, Sarti, Zuidema, and Jumelet]{langedijk_decoderlens_2023}
Anna Langedijk, Hosein Mohebbi, Gabriele Sarti, Willem Zuidema, and Jaap Jumelet.
\newblock \href{https://arxiv.org/abs/2310.03686}{Decoderlens: Layerwise interpretation of encoder-decoder transformers}.
\newblock \emph{CoRR}, 2023.

\bibitem[Lau et~al.(2023)Lau, Murfet, and Wei]{lau_quantifying_2023}
Edmund Lau, Daniel Murfet, and Susan Wei.
\newblock \href{http://arxiv.org/abs/2308.12108}{Quantifying degeneracy in singular models via the learning coefficient}.
\newblock \emph{CoRR}, August 2023.

\bibitem[Leahy(2023)]{leahy_barriers_2023}
Connor Leahy.
\newblock \href{https://www.lesswrong.com/posts/KRDo2afKJtD7bzSM8/barriers-to-mechanistic-interpretability-for-agi-safety}{Barriers to mechanistic interpretability for agi safety}.
\newblock \emph{AI Alignment Forum}, 2023.

\bibitem[Leavitt \& Morcos(2020)Leavitt and Morcos]{leavitt_falsifiable_2020}
Matthew~L. Leavitt and Ari Morcos.
\newblock \href{http://arxiv.org/abs/2010.12016}{Towards falsifiable interpretability research}.
\newblock \emph{CoRR}, October 2020.

\bibitem[Lecomte et~al.(2023)Lecomte, Thaman, Chow, Schaeffer, and Koyejo]{lecomte_incidental_2023}
Victor Lecomte, Kushal Thaman, Trevor Chow, Rylan Schaeffer, and Sanmi Koyejo.
\newblock \href{https://arxiv.org/abs/2312.03096}{Incidental polysemanticity}.
\newblock \emph{CoRR}, 2023.

\bibitem[Lee et~al.(2024)Lee, Bai, Pres, Wattenberg, Kummerfeld, and Mihalcea]{lee_mechanistic_2024}
Andrew Lee, Xiaoyan Bai, Itamar Pres, Martin Wattenberg, Jonathan~K. Kummerfeld, and Rada Mihalcea.
\newblock \href{https://arxiv.org/abs/2401.01967}{A mechanistic understanding of alignment algorithms: A case study on dpo and toxicity}.
\newblock \emph{CoRR}, 2024.

\bibitem[Li et~al.(2021)Li, Nye, and Andreas]{li_implicit_2021}
Belinda~Z. Li, Maxwell Nye, and Jacob Andreas.
\newblock \href{https://aclanthology.org/2021.acl-long.143}{Implicit representations of meaning in neural language models}.
\newblock \emph{ACL-IJCNLP}, August 2021.

\bibitem[Li et~al.(2023{\natexlab{a}})Li, Hopkins, Bau, Vi{\'e}gas, Pfister, and Wattenberg]{li_emergent_2023}
Kenneth Li, Aspen~K. Hopkins, David Bau, Fernanda Vi{\'e}gas, Hanspeter Pfister, and Martin Wattenberg.
\newblock \href{https://arxiv.org/abs/2210.13382}{Emergent world representations: Exploring a sequence model trained on a synthetic task}.
\newblock \emph{ICLR}, 2023{\natexlab{a}}.

\bibitem[Li et~al.(2023{\natexlab{b}})Li, Patel, Vi{\'e}gas, Pfister, and Wattenberg]{li_inferencetime_2023}
Kenneth Li, Oam Patel, Fernanda Vi{\'e}gas, Hanspeter Pfister, and Martin Wattenberg.
\newblock \href{http://arxiv.org/abs/2306.03341}{Inference-time intervention: Eliciting truthful answers from a language model}.
\newblock \emph{NeurIPS Spotlight}, July 2023{\natexlab{b}}.

\bibitem[Li et~al.(2015)Li, Yosinski, Clune, Lipson, and Hopcroft]{li_convergent_2015}
Yixuan Li, Jason Yosinski, Jeff Clune, Hod Lipson, and John Hopcroft.
\newblock \href{https://proceedings.mlr.press/v44/li15convergent.html}{Convergent learning: Do different neural networks learn the same representations?}
\newblock \emph{NIPS Workshop on Feature Extraction}, December 2015.

\bibitem[Lieberum et~al.(2023)Lieberum, Rahtz, Kram{\'a}r, Nanda, Irving, Shah, and Mikulik]{lieberum_does_2023}
Tom Lieberum, Matthew Rahtz, J{\'a}nos Kram{\'a}r, Neel Nanda, Geoffrey Irving, Rohin Shah, and Vladimir Mikulik.
\newblock \href{http://arxiv.org/abs/2307.09458}{Does circuit analysis interpretability scale? evidence from multiple choice capabilities in chinchilla}.
\newblock \emph{CoRR}, July 2023.

\bibitem[Lindner et~al.(2023)Lindner, Kram{\'a}r, Farquhar, Rahtz, McGrath, and Mikulik]{lindner_tracr_2023}
David Lindner, J{\'a}nos Kram{\'a}r, Sebastian Farquhar, Matthew Rahtz, Thomas McGrath, and Vladimir Mikulik.
\newblock \href{https://arxiv.org/abs/2301.05062}{Tracr: Compiled transformers as a laboratory for interpretability}.
\newblock \emph{CoRR}, 2023.

\bibitem[Liu et~al.(2021)Liu, Wang, Kasai, Hajishirzi, and Smith]{liu_probing_2021}
Leo~Z. Liu, Yizhong Wang, Jungo Kasai, Hannaneh Hajishirzi, and Noah~A. Smith.
\newblock \href{http://arxiv.org/abs/2104.07885}{Probing across time: What does roberta know and when?}
\newblock \emph{EMNLP}, September 2021.

\bibitem[Liu \& Tegmark(2023)Liu and Tegmark]{liu_neural_2023}
Ziming Liu and Max Tegmark.
\newblock \href{https://arxiv.org/abs/2310.02258}{A neural scaling law from lottery ticket ensembling}.
\newblock \emph{CoRR}, 2023.

\bibitem[Liu et~al.(2022{\natexlab{a}})Liu, Kitouni, Nolte, Michaud, Tegmark, and Williams]{liu_understanding_2022}
Ziming Liu, Ouail Kitouni, Niklas Nolte, Eric~J. Michaud, Max Tegmark, and Mike Williams.
\newblock \href{https://arxiv.org/abs/2205.10343}{Towards understanding grokking: An effective theory of representation learning}.
\newblock \emph{NeurIPS}, 2022{\natexlab{a}}.

\bibitem[Liu et~al.(2022{\natexlab{b}})Liu, Michaud, and Tegmark]{liu_omnigrok_2022}
Ziming Liu, Eric~J. Michaud, and Max Tegmark.
\newblock \href{https://arxiv.org/abs/2210.01117}{Omnigrok: Grokking beyond algorithmic data}.
\newblock \emph{ICML}, 2022{\natexlab{b}}.

\bibitem[Liu et~al.(2023{\natexlab{a}})Liu, Gan, and Tegmark]{liu_seeing_2023}
Ziming Liu, Eric Gan, and Max Tegmark.
\newblock \href{http://arxiv.org/abs/2305.08746}{Seeing is believing: Brain-inspired modular training for mechanistic interpretability}.
\newblock \emph{Entropy}, June 2023{\natexlab{a}}.

\bibitem[Liu et~al.(2023{\natexlab{b}})Liu, Khona, Fiete, and Tegmark]{liu_growing_2023}
Ziming Liu, Mikail Khona, Ila~R. Fiete, and Max Tegmark.
\newblock \href{https://arxiv.org/abs/2310.07711}{Growing brains: Co-emergence of anatomical and functional modularity in recurrent neural networks}.
\newblock \emph{CoRR}, 2023{\natexlab{b}}.

\bibitem[Liu et~al.(2023{\natexlab{c}})Liu, Zhong, and Tegmark]{liu_grokking_2023}
Ziming Liu, Ziqian Zhong, and Max Tegmark.
\newblock \href{https://www.semanticscholar.org/paper/8cd06e9959151ea852f2842a87ed65ff0a47182f}{Grokking as compression: A nonlinear complexity perspective}.
\newblock \emph{CoRR}, 2023{\natexlab{c}}.

\bibitem[Locatello et~al.(2019)Locatello, Bauer, Lucic, R{\"a}tsch, Gelly, Sch{\"o}lkopf, and Bachem]{locatello_challenging_2019}
Francesco Locatello, Stefan Bauer, Mario Lucic, Gunnar R{\"a}tsch, Sylvain Gelly, Bernhard Sch{\"o}lkopf, and Olivier Bachem.
\newblock \href{http://arxiv.org/abs/1811.12359}{Challenging common assumptions in the unsupervised learning of disentangled representations}.
\newblock \emph{CoRR}, June 2019.

\bibitem[Lu et~al.(2024)Lu, Lu, Lange, Foerster, Clune, and Ha]{lu_ai_2024}
Chris Lu, Cong Lu, Robert~Tjarko Lange, Jakob Foerster, Jeff Clune, and David Ha.
\newblock \href{http://arxiv.org/abs/2408.06292}{The ai scientist: Towards fully automated open-ended scientific discovery}.
\newblock \emph{CoRR}, August 2024.

\bibitem[Lv et~al.(2024)Lv, Chen, Zhang, Wang, Liu, Wen, Xie, and Yan]{lv_interpreting_2024}
Ang Lv, Yuhan Chen, Kaiyi Zhang, Yulong Wang, Lifeng Liu, Ji-Rong Wen, Jian Xie, and Rui Yan.
\newblock \href{https://arxiv.org/abs/2403.19521}{Interpreting key mechanisms of factual recall in transformer-based language models}.
\newblock \emph{CoRR}, 2024.

\bibitem[Magliacane et~al.(2018)Magliacane, {van Ommen}, Claassen, Bongers, Versteeg, and Mooij]{magliacane_domain_2018}
Sara Magliacane, Thijs {van Ommen}, Tom Claassen, Stephan Bongers, Philip Versteeg, and Joris~M. Mooij.
\newblock \href{http://arxiv.org/abs/1707.06422}{Domain adaptation by using causal inference to predict invariant conditional distributions}.
\newblock \emph{NeurIPS}, October 2018.

\bibitem[Makelov(2024)]{makelov_sparse_2024}
Aleksandar Makelov.
\newblock \href{https://openreview.net/forum?id=JdrVuEQih5}{Sparse autoencoders match supervised features for model steering on the ioi task}.
\newblock \emph{ICML MI Workshop}, June 2024.

\bibitem[Makelov et~al.(2024)Makelov, Lange, and Nanda]{makelov_principled_2024}
Aleksandar Makelov, George Lange, and Neel Nanda.
\newblock \href{http://arxiv.org/abs/2405.08366}{Towards principled evaluations of sparse autoencoders for interpretability and control}.
\newblock \emph{CoRR}, May 2024.

\bibitem[Mallen \& Belrose(2023)Mallen and Belrose]{mallen_eliciting_2023}
Alex Mallen and Nora Belrose.
\newblock \href{http://arxiv.org/abs/2312.01037}{Eliciting latent knowledge from quirky language models}.
\newblock \emph{CoRR}, December 2023.

\bibitem[Maloyan et~al.(2024)Maloyan, Verma, Nutfullin, and Ashinov]{maloyan_trojan_2024}
Narek Maloyan, Ekansh Verma, Bulat Nutfullin, and Bislan Ashinov.
\newblock \href{http://arxiv.org/abs/2404.13660}{Trojan detection in large language models: Insights from the trojan detection challenge}.
\newblock \emph{CoRR}, April 2024.

\bibitem[Marchetti et~al.(2023)Marchetti, Hillar, Kragic, and Sanborn]{marchetti_harmonics_2023}
Giovanni~Luca Marchetti, Christopher Hillar, Danica Kragic, and Sophia Sanborn.
\newblock \href{http://arxiv.org/abs/2312.08550}{Harmonics of learning: Universal fourier features emerge in invariant networks}.
\newblock \emph{CoRR}, December 2023.

\bibitem[Marks et~al.(2023{\natexlab{a}})Marks, Abdullah, Mendez, Arike, Torr, and Barez]{marks_interpreting_2023}
Luke Marks, Amir Abdullah, Luna Mendez, Rauno Arike, Philip Torr, and Fazl Barez.
\newblock \href{https://www.semanticscholar.org/paper/Interpreting-Reward-Models-in-RLHF-Tuned-Language-Marks-Abdullah/53b9f8c62837b12d3955778f737678aabea39a83}{Interpreting reward models in rlhf-tuned language models using sparse autoencoders}.
\newblock \emph{CoRR}, October 2023{\natexlab{a}}.

\bibitem[Marks et~al.(2023{\natexlab{b}})Marks, Abdullah, Neo, Arike, Torr, and Barez]{marks_training_2023}
Luke Marks, Amir Abdullah, Clement Neo, Rauno Arike, Philip Torr, and Fazl Barez.
\newblock \href{https://arxiv.org/abs/2310.08164}{Beyond training objectives: Interpreting reward model divergence in large language models}.
\newblock \emph{CoRR}, October 2023{\natexlab{b}}.

\bibitem[Marks et~al.(2024)Marks, Rager, Michaud, Belinkov, Bau, and Mueller]{marks_sparse_2024}
Samuel Marks, Can Rager, Eric~J. Michaud, Yonatan Belinkov, David Bau, and Aaron Mueller.
\newblock \href{http://arxiv.org/abs/2403.19647}{Sparse feature circuits: Discovering and editing interpretable causal graphs in language models}.
\newblock \emph{CoRR}, March 2024.

\bibitem[Marshall \& Kirchner(2024)Marshall and Kirchner]{marshall_understanding_2024}
Simon~C. Marshall and Jan~H. Kirchner.
\newblock \href{http://arxiv.org/abs/2401.17975}{Understanding polysemanticity in neural networks through coding theory}.
\newblock \emph{CoRR}, January 2024.

\bibitem[Mazeika et~al.(2022)Mazeika, Zou, Arora, Pleskov, Song, Hendrycks, Li, and Forsyth]{mazeika_how_2022}
Mantas Mazeika, Andy Zou, Akul Arora, Pavel Pleskov, Dawn Song, Dan Hendrycks, Bo~Li, and David Forsyth.
\newblock \href{https://openreview.net/forum?id=V-RDBWYf0go}{How hard is trojan detection in dnns? fooling detectors with evasive trojans}.
\newblock \emph{CoRR}, September 2022.

\bibitem[McDougall et~al.(2023)McDougall, Conmy, Rushing, McGrath, and Nanda]{mcdougall_copy_2023}
Callum McDougall, Arthur Conmy, Cody Rushing, Thomas McGrath, and Neel Nanda.
\newblock \href{http://arxiv.org/abs/2310.04625}{Copy suppression: Comprehensively understanding an attention head}.
\newblock \emph{CoRR}, October 2023.

\bibitem[McGrath et~al.(2022)McGrath, Kapishnikov, Toma{\v s}ev, Pearce, Wattenberg, Hassabis, Kim, Paquet, and Kramnik]{mcgrath_acquisition_2022}
Thomas McGrath, Andrei Kapishnikov, Nenad Toma{\v s}ev, Adam Pearce, Martin Wattenberg, Demis Hassabis, Been Kim, Ulrich Paquet, and Vladimir Kramnik.
\newblock \href{https://www.pnas.org/doi/abs/10.1073/pnas.2206625119}{Acquisition of chess knowledge in alphazero}.
\newblock \emph{PNAS}, November 2022.

\bibitem[McGrath et~al.(2023)McGrath, Rahtz, Kramar, Mikulik, and Legg]{mcgrath_hydra_2023}
Thomas McGrath, Matthew Rahtz, Janos Kramar, Vladimir Mikulik, and Shane Legg.
\newblock \href{http://arxiv.org/abs/2307.15771}{The hydra effect: Emergent self-repair in language model computations}.
\newblock \emph{CoRR}, July 2023.

\bibitem[Meng et~al.(2022{\natexlab{a}})Meng, Bau, Andonian, and Belinkov]{meng_locating_2022}
Kevin Meng, David Bau, Alex Andonian, and Yonatan Belinkov.
\newblock \href{http://arxiv.org/abs/2202.05262}{Locating and editing factual associations in gpt}.
\newblock \emph{NeurIPS}, 2022{\natexlab{a}}.

\bibitem[Meng et~al.(2022{\natexlab{b}})Meng, Sharma, Andonian, Belinkov, and Bau]{meng_massediting_2022}
Kevin Meng, Arnab~Sen Sharma, Alex Andonian, Yonatan Belinkov, and David Bau.
\newblock \href{http://arxiv.org/abs/2210.07229}{Mass-editing memory in a transformer}.
\newblock \emph{ICLR}, 2022{\natexlab{b}}.

\bibitem[Merrill et~al.(2023)Merrill, Tsilivis, and Shukla]{merrill_tale_2023}
William Merrill, Nikolaos Tsilivis, and Aman Shukla.
\newblock \href{https://arxiv.org/abs/2303.11873}{A tale of two circuits: Grokking as competition of sparse and dense subnetworks}.
\newblock \emph{CoRR}, 2023.

\bibitem[Merullo et~al.(2023)Merullo, Eickhoff, and Pavlick]{merullo_mechanism_2023}
Jack Merullo, Carsten Eickhoff, and Ellie Pavlick.
\newblock \href{https://www.semanticscholar.org/paper/A-Mechanism-for-Solving-Relational-Tasks-in-Models-Merullo-Eickhoff/823fb3163a5600b5be957fc8337a9f7cdd177fef}{A mechanism for solving relational tasks in transformer language models}.
\newblock \emph{CoRR}, May 2023.

\bibitem[Michaud et~al.(2023)Michaud, Liu, Girit, and Tegmark]{michaud_quantization_2023}
Eric~J. Michaud, Ziming Liu, Uzay Girit, and Max Tegmark.
\newblock \href{http://arxiv.org/abs/2303.13506}{The quantization model of neural scaling}.
\newblock \emph{CoRR}, March 2023.

\bibitem[Michaud et~al.(2024)Michaud, Liao, Lad, Liu, Mudide, Loughridge, Guo, Kheirkhah, Vukeli{\'c}, and Tegmark]{michaud_opening_2024}
Eric~J. Michaud, Isaac Liao, Vedang Lad, Ziming Liu, Anish Mudide, Chloe Loughridge, Zifan~Carl Guo, Tara~Rezaei Kheirkhah, Mateja Vukeli{\'c}, and Max Tegmark.
\newblock \href{http://arxiv.org/abs/2402.05110}{Opening the ai black box: program synthesis via mechanistic interpretability}.
\newblock \emph{CoRR}, February 2024.

\bibitem[Mikolov et~al.(2013)Mikolov, Sutskever, Chen, Corrado, and Dean]{mikolov_distributed_2013}
Tomas Mikolov, Ilya Sutskever, Kai Chen, Greg Corrado, and Jeffrey Dean.
\newblock \href{http://arxiv.org/abs/1310.4546}{Distributed representations of words and phrases and their compositionality}.
\newblock \emph{NeurIPS}, October 2013.

\bibitem[Miller \& Neo(2023)Miller and Neo]{miller_we_2023}
Joseph Miller and Clement Neo.
\newblock \href{https://www.alignmentforum.org/posts/cgqh99SHsCv3jJYDS/we-found-an-neuron-in-gpt-2}{We found an neuron in gpt-2}.
\newblock \emph{AI Alignment Forum}, February 2023.

\bibitem[Miller(2019)]{miller_explanation_2019}
Tim Miller.
\newblock \href{https://www.sciencedirect.com/science/article/pii/S0004370218305988}{Explanation in artificial intelligence: Insights from the social sciences}.
\newblock \emph{Artificial Intelligence}, February 2019.

\bibitem[Mini et~al.(2023)Mini, Grietzer, Sharma, Meek, MacDiarmid, and Turner]{mini_understanding_2023}
Ulisse Mini, Peli Grietzer, Mrinank Sharma, Austin Meek, Monte MacDiarmid, and Alexander~Matt Turner.
\newblock \href{http://arxiv.org/abs/2310.08043}{Understanding and controlling a maze-solving policy network}.
\newblock \emph{CoRR}, October 2023.

\bibitem[Monea et~al.(2023)Monea, Peyrard, Josifoski, Chaudhary, Eisner, K{\i}c{\i}man, Palangi, Patra, and West]{monea_glitch_2023}
Giovanni Monea, Maxime Peyrard, Martin Josifoski, Vishrav Chaudhary, Jason Eisner, Emre K{\i}c{\i}man, Hamid Palangi, Barun Patra, and Robert West.
\newblock \href{https://arxiv.org/abs/2312.02073}{A glitch in the matrix? locating and detecting language model grounding with fakepedia}.
\newblock \emph{CoRR}, 2023.

\bibitem[Mousi et~al.(2023)Mousi, Durrani, and Dalvi]{mousi_can_2023}
Basel Mousi, Nadir Durrani, and Fahim Dalvi.
\newblock \href{https://arxiv.org/abs/2305.13386}{Can llms facilitate interpretation of pre-trained language models?}
\newblock \emph{EMNLP}, 2023.

\bibitem[Mu \& Andreas(2020)Mu and Andreas]{mu_compositional_2020}
Jesse Mu and Jacob Andreas.
\newblock \href{http://arxiv.org/abs/2006.14032}{Compositional explanations of neurons}.
\newblock \emph{NeurIPS}, June 2020.

\bibitem[Mueller et~al.(2024)Mueller, Brinkmann, Li, Marks, Pal, Prakash, Rager, Sankaranarayanan, Sharma, Sun, Todd, Bau, and Belinkov]{mueller_quest_2024}
Aaron Mueller, Jannik Brinkmann, Millicent Li, Samuel Marks, Koyena Pal, Nikhil Prakash, Can Rager, Aruna Sankaranarayanan, Arnab~Sen Sharma, Jiuding Sun, Eric Todd, David Bau, and Yonatan Belinkov.
\newblock \href{https://arxiv.org/abs/2408.01416}{The quest for the right mediator: A history, survey, and theoretical grounding of causal interpretability}.
\newblock \emph{CoRR}, August 2024.

\bibitem[Nainani(2024)]{nainani_evaluating_2024}
Jatin Nainani.
\newblock \href{http://arxiv.org/abs/2401.03646}{Evaluating brain-inspired modular training in automated circuit discovery for mechanistic interpretability}.
\newblock \emph{CoRR}, January 2024.

\bibitem[Nakkiran et~al.(2019)Nakkiran, Kaplun, Kalimeris, Yang, Edelman, Zhang, and Barak]{nakkiran_sgd_2019}
Preetum Nakkiran, Gal Kaplun, Dimitris Kalimeris, Tristan Yang, Benjamin~L. Edelman, Fred Zhang, and Boaz Barak.
\newblock \href{http://arxiv.org/abs/1905.11604}{Sgd on neural networks learns functions of increasing complexity}.
\newblock \emph{NeurIPS}, May 2019.

\bibitem[Nanda(2022{\natexlab{a}})]{nanda_200circuits_2022}
Neel Nanda.
\newblock \href{https://www.lesswrong.com/posts/XNjRwEX9kxbpzWFWd/200-cop-in-mi-looking-for-circuits-in-the-wild}{200 cop in mi: Looking for circuits in the wild}.
\newblock \emph{Neel Nanda's Blog}, 2022{\natexlab{a}}.

\bibitem[Nanda(2022{\natexlab{b}})]{nanda_200dynamics_2022}
Neel Nanda.
\newblock \href{https://www.lesswrong.com/posts/hHaXzJQi6SKkeXzbg/200-cop-in-mi-analysing-training-dynamics}{200 cop in mi: Analysing training dynamics}.
\newblock \emph{Neel Nanda's Blog}, 2022{\natexlab{b}}.

\bibitem[Nanda(2022{\natexlab{c}})]{nanda_200features_2022}
Neel Nanda.
\newblock \href{https://www.lesswrong.com/posts/Qup9gorqpd9qKAEav/200-cop-in-mi-studying-learned-features-in-language-models}{200 cop in mi: Studying learned features in language models}.
\newblock \emph{Neel Nanda's Blog}, 2022{\natexlab{c}}.

\bibitem[Nanda(2022{\natexlab{d}})]{nanda_comprehensive_2022}
Neel Nanda.
\newblock \href{https://www.neelnanda.io/mechanistic-interpretability/glossary}{A comprehensive mechanistic interpretability explainer \& glossary}.
\newblock \emph{Neel Nanda's Blog}, December 2022{\natexlab{d}}.

\bibitem[Nanda(2022{\natexlab{e}})]{nanda_longlist_2022}
Neel Nanda.
\newblock \href{https://www.alignmentforum.org/posts/uK6sQCNMw8WKzJeCQ/a-longlist-of-theories-of-impact-for-interpretability}{A longlist of theories of impact for interpretability}.
\newblock \emph{AI Alignment Forum}, March 2022{\natexlab{e}}.

\bibitem[Nanda(2023{\natexlab{a}})]{nanda_200superposition_2023}
Neel Nanda.
\newblock \href{https://www.lesswrong.com/posts/o6ptPu7arZrqRCxyz/200-cop-in-mi-exploring-polysemanticity-and-superposition}{200 cop in mi: Exploring polysemanticity and superposition}.
\newblock \emph{Neel Nanda's Blog}, January 2023{\natexlab{a}}.

\bibitem[Nanda(2023{\natexlab{b}})]{nanda_200tool_2023}
Neel Nanda.
\newblock \href{https://www.lesswrong.com/posts/btasQF7wiCYPsr5qw/200-cop-in-mi-techniques-tooling-and-automation}{200 cop in mi: Techniques, tooling and automation}.
\newblock \emph{Neel Nanda's Blog}, 2023{\natexlab{b}}.

\bibitem[Nanda(2023{\natexlab{c}})]{nanda_actually_2023}
Neel Nanda.
\newblock \href{https://neelnanda.io/mechanistic-interpretability/othello}{Actually, othello-gpt has a linear emergent world representation}.
\newblock \emph{Neel Nanda's Blog}, March 2023{\natexlab{c}}.

\bibitem[Nanda(2023{\natexlab{d}})]{nanda_attribution_2023}
Neel Nanda.
\newblock \href{https://www.neelnanda.io/mechanistic-interpretability/attribution-patching}{Attribution patching: Activation patching at industrial scale}.
\newblock \emph{Neel Nanda's Blog}, February 2023{\natexlab{d}}.

\bibitem[Nanda(2023{\natexlab{e}})]{nanda_how_2023}
Neel Nanda.
\newblock \href{https://www.lesswrong.com/posts/xh85KbTFhbCz7taD4/how-to-think-about-activation-patching}{How to think about activation patching}.
\newblock \emph{AI Alignment Forum}, April 2023{\natexlab{e}}.

\bibitem[Nanda(2023{\natexlab{f}})]{nanda_mechanistic_2023}
Neel Nanda.
\newblock \href{https://www.neelnanda.io/mechanistic-interpretability/quickstart}{Mechanistic interpretability quickstart guide}.
\newblock \emph{Neel Nanda's Blog}, January 2023{\natexlab{f}}.

\bibitem[Nanda(2024)]{nanda_extremely_2024}
Neel Nanda.
\newblock \href{https://www.alignmentforum.org/posts/NfFST5Mio7BCAQHPA/an-extremely-opinionated-annotated-list-of-my-favourite}{An extremely opinionated annotated list of my favourite mechanistic interpretability papers v2}.
\newblock \emph{AI Alignment Forum}, July 2024.

\bibitem[Nanda et~al.(2023{\natexlab{a}})Nanda, Chan, Lieberum, Smith, and Steinhardt]{nanda_progress_2023}
Neel Nanda, Lawrence Chan, Tom Lieberum, Jess Smith, and Jacob Steinhardt.
\newblock \href{http://arxiv.org/abs/2301.05217}{Progress measures for grokking via mechanistic interpretability}.
\newblock \emph{ICLR}, January 2023{\natexlab{a}}.

\bibitem[Nanda et~al.(2023{\natexlab{b}})Nanda, Lee, and Wattenberg]{nanda_emergent_2023}
Neel Nanda, Andrew Lee, and Martin Wattenberg.
\newblock \href{http://arxiv.org/abs/2309.00941}{Emergent linear representations in world models of self-supervised sequence models}.
\newblock \emph{BlackboxNLP Workshop on Analyzing and Interpreting Neural Networks for NLP}, September 2023{\natexlab{b}}.

\bibitem[Nanda et~al.(2023{\natexlab{c}})Nanda, Rajamanoharan, Kram{\'a}r, and Shah]{nanda_fact_2023}
Neel Nanda, S.~Rajamanoharan, J.~Kram{\'a}r, and R.~Shah.
\newblock \href{https://scholar.google.com/scholar?cluster=9546235215420404405&hl=en&oi=scholarr}{Fact finding: Attempting to reverse-engineer factual recall on the neuron level}.
\newblock \emph{AI Alignment Forum, 2023c. URL https://www. alignmentforum. org/posts/iGuwZTHWb6DFY3sKB/fact-finding-attempting-to-reverse-engineer-factual-recall}, 2023{\natexlab{c}}.

\bibitem[Nanfack et~al.(2024)Nanfack, Fulleringer, Marty, Eickenberg, and Belilovsky]{nanfack_adversarial_2024}
Geraldin Nanfack, Alexander Fulleringer, Jonathan Marty, Michael Eickenberg, and Eugene Belilovsky.
\newblock \href{https://ojs.aaai.org/index.php/AAAI/article/view/28228}{Adversarial attacks on the interpretation of neuron activation maximization}.
\newblock \emph{Proceedings of the AAAI Conference on Artificial Intelligence}, March 2024.

\bibitem[Ngo et~al.(2022)Ngo, Chan, and Mindermann]{ngo_alignment_2022}
Richard Ngo, Lawrence Chan, and S{\"o}ren Mindermann.
\newblock \href{http://arxiv.org/abs/2209.00626}{The alignment problem from a deep learning perspective}.
\newblock \emph{CoRR}, December 2022.

\bibitem[Nguyen et~al.(2022)Nguyen, Huynh, Nguyen, Liew, Yin, and Nguyen]{nguyen_survey_2022}
Thanh~Tam Nguyen, Thanh~Trung Huynh, Phi~Le Nguyen, Alan Wee-Chung Liew, Hongzhi Yin, and Quoc Viet~Hung Nguyen.
\newblock \href{http://arxiv.org/abs/2209.02299}{A survey of machine unlearning}.
\newblock \emph{CoRR}, October 2022.

\bibitem[Nichani et~al.(2024)Nichani, Damian, and Lee]{nichani_how_2024}
Eshaan Nichani, Alex Damian, and Jason~D. Lee.
\newblock \href{http://arxiv.org/abs/2402.14735}{How transformers learn causal structure with gradient descent}.
\newblock \emph{CoRR}, February 2024.

\bibitem[NicholasKees \& {janus}(2022)NicholasKees and {janus}]{nicholaskees_searching_2022}
NicholasKees and {janus}.
\newblock \href{https://www.alignmentforum.org/posts/FDjTgDcGPc7B98AES/searching-for-search-4}{Searching for search}.
\newblock \emph{AI Alignment Forum}, November 2022.

\bibitem[{nostalgebraist}(2020)]{nostalgebraist_interpreting_2020}
{nostalgebraist}.
\newblock \href{https://www.alignmentforum.org/posts/AcKRB8wDpdaN6v6ru/interpreting-gpt-the-logit-lens}{interpreting gpt: the logit lens}.
\newblock \emph{AI Alignment Forum}, August 2020.

\bibitem[Olah(2023)]{olah_distributed_2023}
Chris Olah.
\newblock \href{https://transformer-circuits.pub/2023/superposition-composition/index.html}{Distributed representations: Composition \& superposition}.
\newblock \emph{Transformer Circuits Thread}, 2023.

\bibitem[Olah \& Carter(2017)Olah and Carter]{olah_research_2017}
Chris Olah and Shan Carter.
\newblock \href{https://distill.pub/2017/research-debt}{Research debt}.
\newblock \emph{Distill}, March 2017.

\bibitem[Olah et~al.(2017)Olah, Mordvintsev, and Schubert]{olah_feature_2017}
Chris Olah, Alexander Mordvintsev, and Ludwig Schubert.
\newblock \href{https://distill.pub/2017/feature-visualization}{Feature visualization}.
\newblock \emph{Distill}, November 2017.

\bibitem[Olah et~al.(2018)Olah, Satyanarayan, Johnson, Carter, Schubert, Ye, and Mordvintsev]{olah_building_2018}
Chris Olah, Arvind Satyanarayan, Ian Johnson, Shan Carter, Ludwig Schubert, Katherine Ye, and Alexander Mordvintsev.
\newblock \href{https://distill.pub/2018/building-blocks}{The building blocks of interpretability}.
\newblock \emph{Distill}, March 2018.

\bibitem[Olah et~al.(2020)Olah, Cammarata, Schubert, Goh, Petrov, and Carter]{olah_zoom_2020}
Chris Olah, Nick Cammarata, Ludwig Schubert, Gabriel Goh, Michael Petrov, and Shan Carter.
\newblock \href{https://distill.pub/2020/circuits/zoom-in}{Zoom in: An introduction to circuits}.
\newblock \emph{Distill}, March 2020.

\bibitem[Olah(2022)]{olah_mechanistic_2022}
Christopher Olah.
\newblock \href{https://transformer-circuits.pub/2022/mech-interp-essay/index.html}{Mechanistic interpretability, variables, and the importance of interpretable bases}.
\newblock \emph{Transformer Circuits Thread}, 2022.

\bibitem[Olshausen \& Field(1997)Olshausen and Field]{olshausen_sparse_1997}
B.~A. Olshausen and D.~J. Field.
\newblock \href{https://pubmed.ncbi.nlm.nih.gov/9425546/}{Sparse coding with an overcomplete basis set: a strategy employed by v1?}
\newblock \emph{Vision Res}, December 1997.

\bibitem[Olsson et~al.(2022)Olsson, Elhage, Nanda, Joseph, DasSarma, Henighan, Mann, Askell, Bai, Chen, Conerly, Drain, Ganguli, {Hatfield-Dodds}, Hernandez, Johnston, Jones, Kernion, Lovitt, Ndousse, Amodei, Brown, Clark, Kaplan, McCandlish, and Olah]{olsson_incontext_2022}
Catherine Olsson, Nelson Elhage, Neel Nanda, Nicholas Joseph, Nova DasSarma, Tom Henighan, Ben Mann, Amanda Askell, Yuntao Bai, Anna Chen, Tom Conerly, Dawn Drain, Deep Ganguli, Zac {Hatfield-Dodds}, Danny Hernandez, Scott Johnston, Andy Jones, Jackson Kernion, Liane Lovitt, Kamal Ndousse, Dario Amodei, Tom Brown, Jack Clark, Jared Kaplan, Sam McCandlish, and Chris Olah.
\newblock \href{https://transformer-circuits.pub/2022/in-context-learning-and-induction-heads/index.html}{In-context learning and induction heads}.
\newblock \emph{Transformer Circuits Thread}, 2022.

\bibitem[O'Mahony et~al.(2023)O'Mahony, Andrearczyk, Muller, and Graziani]{omahony_disentangling_2023}
Laura O'Mahony, Vincent Andrearczyk, Henning Muller, and Mara Graziani.
\newblock \href{http://arxiv.org/abs/2304.09707}{Disentangling neuron representations with concept vectors}.
\newblock \emph{CVPR Workshops}, April 2023.

\bibitem[O'Neill \& Bui(2024)O'Neill and Bui]{oneill_sparse_2024}
Charles O'Neill and Thang Bui.
\newblock \href{http://arxiv.org/abs/2405.12522}{Sparse autoencoders enable scalable and reliable circuit identification in language models}.
\newblock \emph{CoRR}, May 2024.

\bibitem[Ortu et~al.(2024)Ortu, Jin, Doimo, Sachan, Cazzaniga, and Sch{\"o}lkopf]{ortu_competition_2024}
Francesco Ortu, Zhijing Jin, Diego Doimo, Mrinmaya Sachan, Alberto Cazzaniga, and Bernhard Sch{\"o}lkopf.
\newblock \href{http://arxiv.org/abs/2402.11655}{Competition of mechanisms: Tracing how language models handle facts and counterfactuals}.
\newblock \emph{ACL 2024}, June 2024.

\bibitem[Ouyang et~al.(2022)Ouyang, Wu, Jiang, Almeida, Wainwright, Mishkin, Zhang, Agarwal, Slama, Ray, Schulman, Hilton, Kelton, Miller, Simens, Askell, Welinder, Christiano, Leike, and Lowe]{ouyang_training_2022}
Long Ouyang, Jeff Wu, Xu~Jiang, Diogo Almeida, Carroll~L. Wainwright, Pamela Mishkin, Chong Zhang, Sandhini Agarwal, Katarina Slama, Alex Ray, John Schulman, Jacob Hilton, Fraser Kelton, Luke Miller, Maddie Simens, Amanda Askell, Peter Welinder, Paul Christiano, Jan Leike, and Ryan Lowe.
\newblock \href{https://arxiv.org/abs/2203.02155}{Training language models to follow instructions with human feedback}.
\newblock \emph{CoRR}, 2022.

\bibitem[Pal et~al.(2023)Pal, Sun, Yuan, Wallace, and Bau]{pal_future_2023}
Koyena Pal, Jiuding Sun, Andrew Yuan, Byron~C. Wallace, and David Bau.
\newblock \href{https://arxiv.org/abs/2311.04897}{Future lens: Anticipating subsequent tokens from a single hidden state}.
\newblock \emph{CoNLL}, 2023.

\bibitem[Palit et~al.(2023)Palit, Pandey, Arora, and Liang]{palit_visionlanguage_2023}
Vedant Palit, Rohan Pandey, Aryaman Arora, and P.~Liang.
\newblock \href{https://www.semanticscholar.org/paper/d494727306a375e524c4c4c8cc1a2dc1845cc4b7}{Towards vision-language mechanistic interpretability: A causal tracing tool for blip}.
\newblock \emph{ICCVW}, 2023.

\bibitem[Pan et~al.(2024)Pan, Philip, Xie, and Schwartz]{pan_dissecting_2024}
Xu~Pan, Aaron Philip, Ziqian Xie, and Odelia Schwartz.
\newblock \href{https://openreview.net/forum?id=CsF3PwBN6N}{Dissecting query-key interaction in vision transformers}.
\newblock \emph{ICML MI Workshop}, June 2024.

\bibitem[Park et~al.(2023{\natexlab{a}})Park, Choe, and Veitch]{park_linear_2023}
Kiho Park, Yo~Joong Choe, and Victor Veitch.
\newblock \href{http://arxiv.org/abs/2311.03658}{The linear representation hypothesis and the geometry of large language models}.
\newblock \emph{NeurIPS Workshop on Causal Representation Learning}, November 2023{\natexlab{a}}.

\bibitem[Park et~al.(2024)Park, Choe, Jiang, and Veitch]{park_geometry_2024}
Kiho Park, Yo~Joong Choe, Yibo Jiang, and Victor Veitch.
\newblock \href{https://openreview.net/forum?id=KXuYjuBzKo}{The geometry of categorical and hierarchical concepts in large language models}.
\newblock \emph{ICML MI Workshop (Oral)}, June 2024.

\bibitem[Park et~al.(2023{\natexlab{b}})Park, Goldstein, O'Gara, Chen, and Hendrycks]{park_ai_2023}
Peter~S. Park, Simon Goldstein, Aidan O'Gara, Michael Chen, and Dan Hendrycks.
\newblock \href{http://arxiv.org/abs/2308.14752}{Ai deception: A survey of examples, risks, and potential solutions}.
\newblock \emph{CoRR}, August 2023{\natexlab{b}}.

\bibitem[Patel \& Pavlick(2022)Patel and Pavlick]{patel_mapping_2022}
Roma Patel and Ellie Pavlick.
\newblock \href{https://openreview.net/forum?id=gJcEM8sxHK}{Mapping language models to grounded conceptual spaces}.
\newblock \emph{ICLR}, 2022.

\bibitem[Pearl(2009)]{pearl_causality_2009}
Judea Pearl.
\newblock \href{https://www.cambridge.org/core/books/causality/B0046844FAE10CBF274D4ACBDAEB5F5B}{\emph{Causality}}.
\newblock Cambridge University Press, 2009.

\bibitem[Pearl \& Mackenzie(2018)Pearl and Mackenzie]{pearl_book_2018}
Judea Pearl and Dana Mackenzie.
\newblock \emph{The Book of Why: The New Science of Cause and Effect}.
\newblock Penguin Books Limited, May 2018.

\bibitem[Peters et~al.(2015)Peters, B{\"u}hlmann, and Meinshausen]{peters_causal_2015}
Jonas Peters, Peter B{\"u}hlmann, and Nicolai Meinshausen.
\newblock \href{http://arxiv.org/abs/1501.01332}{Causal inference using invariant prediction: identification and confidence intervals}.
\newblock \emph{Journal of the Royal Statistical Society Series B: Statistical Methodology}, November 2015.

\bibitem[Peters et~al.(2017)Peters, Janzing, and Schlkopf]{peters_elements_2017}
Jonas Peters, Dominik Janzing, and Bernhard Schlkopf.
\newblock \emph{Elements of Causal Inference: Foundations and Learning Algorithms}.
\newblock The MIT Press, October 2017.

\bibitem[Pochinkov \& ~(2023)Pochinkov and ~]{pochinkov_machine_2023}
Nicky Pochinkov and Nandi ~.
\newblock \href{https://www.alignmentforum.org/posts/mTi8TQEyP5Pr7oczd/machine-unlearning-evaluations-as-interpretability}{Machine unlearning evaluations as interpretability benchmarks}.
\newblock \emph{AI Alignment Forum}, August 2023.

\bibitem[Poli et~al.(2023)Poli, Massaroli, Nguyen, Fu, Dao, Baccus, Bengio, Ermon, and R{\'e}]{poli_hyena_2023}
Michael Poli, Stefano Massaroli, Eric Nguyen, Daniel~Y. Fu, Tri Dao, Stephen Baccus, Yoshua Bengio, Stefano Ermon, and Christopher R{\'e}.
\newblock \href{http://arxiv.org/abs/2302.10866}{Hyena hierarchy: Towards larger convolutional language models}.
\newblock \emph{ICML}, April 2023.

\bibitem[Power et~al.(2022)Power, Burda, Edwards, Babuschkin, and Misra]{power_grokking_2022}
Alethea Power, Yuri Burda, Harri Edwards, Igor Babuschkin, and Vedant Misra.
\newblock \href{http://arxiv.org/abs/2201.02177}{Grokking: Generalization beyond overfitting on small algorithmic datasets}.
\newblock \emph{CoRR}, January 2022.

\bibitem[Prakash et~al.(2024)Prakash, Shaham, Haklay, Belinkov, and Bau]{prakash_finetuning_2024}
Nikhil Prakash, Tamar~Rott Shaham, Tal Haklay, Yonatan Belinkov, and David Bau.
\newblock \href{http://arxiv.org/abs/2402.14811}{Fine-tuning enhances existing mechanisms: A case study on entity tracking}.
\newblock \emph{ICLR}, February 2024.

\bibitem[Quirke \& Barez(2023)Quirke and Barez]{quirke_understanding_2023}
Philip Quirke and Fazl Barez.
\newblock \href{http://arxiv.org/abs/2310.13121}{Understanding addition in transformers}.
\newblock \emph{CoRR}, October 2023.

\bibitem[Quirke et~al.(2024)Quirke, Neo, and Barez]{quirke_increasing_2024}
Philip Quirke, Clement Neo, and Fazl Barez.
\newblock \href{http://arxiv.org/abs/2402.02619}{Increasing trust in language models through the reuse of verified circuits}.
\newblock \emph{CoRR}, February 2024.

\bibitem[Rai et~al.(2024)Rai, Zhou, Feng, Saparov, and Yao]{rai_practical_2024}
Daking Rai, Yilun Zhou, Shi Feng, Abulhair Saparov, and Ziyu Yao.
\newblock \href{http://arxiv.org/abs/2407.02646}{A practical review of mechanistic interpretability for transformer-based language models}.
\newblock \emph{CoRR}, July 2024.

\bibitem[Rajamanoharan et~al.(2024)Rajamanoharan, Conmy, Smith, Lieberum, Varma, Kram{\'a}r, Shah, and Nanda]{rajamanoharan_improving_2024}
Senthooran Rajamanoharan, Arthur Conmy, Lewis Smith, Tom Lieberum, Vikrant Varma, J{\'a}nos Kram{\'a}r, Rohin Shah, and Neel Nanda.
\newblock \href{http://arxiv.org/abs/2404.16014}{Improving dictionary learning with gated sparse autoencoders}.
\newblock \emph{CoRR}, April 2024.

\bibitem[R{\"a}uker et~al.(2023)R{\"a}uker, Ho, Casper, and {Hadfield-Menell}]{rauker_transparent_2023}
Tilman R{\"a}uker, Anson Ho, Stephen Casper, and Dylan {Hadfield-Menell}.
\newblock \href{http://arxiv.org/abs/2207.13243}{Toward transparent ai: A survey on interpreting the inner structures of deep neural networks}.
\newblock \emph{TMLR}, August 2023.

\bibitem[Ravfogel et~al.(2020)Ravfogel, Elazar, Gonen, Twiton, and Goldberg]{ravfogel_null_2020}
Shauli Ravfogel, Yanai Elazar, Hila Gonen, Michael Twiton, and Yoav Goldberg.
\newblock \href{https://aclanthology.org/2020.acl-main.647}{Null it out: Guarding protected attributes by iterative nullspace projection}.
\newblock \emph{ACL}, July 2020.

\bibitem[Ren et~al.(2023)Ren, Li, Chen, Deng, and Zhang]{ren_defining_2023}
Jie Ren, Mingjie Li, Qirui Chen, Huiqi Deng, and Quanshi Zhang.
\newblock \href{http://arxiv.org/abs/2111.06206}{Defining and quantifying the emergence of sparse concepts in dnns}.
\newblock \emph{CoRR}, April 2023.

\bibitem[Ribeiro et~al.(2016)Ribeiro, Singh, and Guestrin]{ribeiro_why_2016}
Marco~Tulio Ribeiro, Sameer Singh, and Carlos Guestrin.
\newblock \href{http://arxiv.org/abs/1602.04938}{"why should i trust you?": Explaining the predictions of any classifier}.
\newblock \emph{NAACL}, August 2016.

\bibitem[RicG(2023)]{__ricg___agiautomated_2023}
RicG.
\newblock \href{https://www.lesswrong.com/posts/pQqoTTAnEePRDmZN4/agi-automated-interpretability-is-suicide}{Agi-automated interpretability is suicide}.
\newblock \emph{LessWrong}, May 2023.

\bibitem[Richens \& Everitt(2024)Richens and Everitt]{richens_robust_2024}
Jonathan Richens and Tom Everitt.
\newblock \href{http://arxiv.org/abs/2402.10877}{Robust agents learn causal world models}.
\newblock \emph{ICLR Oral}, February 2024.

\bibitem[Riegel et~al.(2020)Riegel, Gray, Luus, Khan, Makondo, Akhalwaya, Qian, Fagin, Barahona, Sharma, Ikbal, Karanam, Neelam, Likhyani, and Srivastava]{riegel_logical_2020}
Ryan Riegel, Alexander Gray, Francois Luus, Naweed Khan, Ndivhuwo Makondo, Ismail~Yunus Akhalwaya, Haifeng Qian, Ronald Fagin, Francisco Barahona, Udit Sharma, Shajith Ikbal, Hima Karanam, Sumit Neelam, Ankita Likhyani, and Santosh Srivastava.
\newblock \href{http://arxiv.org/abs/2006.13155}{Logical neural networks}.
\newblock \emph{NeurIPS}, June 2020.

\bibitem[{Rojas-Carulla} et~al.(2018){Rojas-Carulla}, Scholkopf, Turner, and Peters]{rojas-carulla_invariant_2018}
Mateo {Rojas-Carulla}, Bernhard Scholkopf, Richard Turner, and Jonas Peters.
\newblock Invariant models for causal transfer learning.
\newblock \emph{JMLR}, 2018.

\bibitem[Ross \& {Doshi-Velez}(2017)Ross and {Doshi-Velez}]{ross_improving_2017}
Andrew~Slavin Ross and Finale {Doshi-Velez}.
\newblock \href{http://arxiv.org/abs/1711.09404}{Improving the adversarial robustness and interpretability of deep neural networks by regularizing their input gradients}.
\newblock \emph{AAAI}, November 2017.

\bibitem[Rudin(2019)]{rudin_stop_2019}
Cynthia Rudin.
\newblock \href{https://www.nature.com/articles/s42256-019-0048-x}{Stop explaining black box machine learning models for high stakes decisions and use interpretable models instead}.
\newblock \emph{Nat Mach Intell}, May 2019.

\bibitem[Rushing \& Nanda(2024)Rushing and Nanda]{rushing_explorations_2024}
Cody Rushing and Neel Nanda.
\newblock \href{http://arxiv.org/abs/2402.15390}{Explorations of self-repair in language models}.
\newblock \emph{ICML}, May 2024.

\bibitem[Ruthenis(2022)]{ruthenis_internal_2022}
Thane Ruthenis.
\newblock \href{https://www.lesswrong.com/posts/nwLQt4e7bstCyPEXs/internal-interfaces-are-a-high-priority-interpretability}{Internal interfaces are a high-priority interpretability target}.
\newblock \emph{AI Alignment Forum}, December 2022.

\bibitem[Ruthenis(2023)]{ruthenis_worldmodel_2023}
Thane Ruthenis.
\newblock \href{https://www.alignmentforum.org/posts/HaHcsrDSZ3ZC2b4fK/world-model-interpretability-is-all-we-need}{World-model interpretability is all we need}.
\newblock \emph{AI Alignment Forum}, January 2023.

\bibitem[Sajjad et~al.(2022)Sajjad, Durrani, and Dalvi]{sajjad_neuronlevel_2022}
Hassan Sajjad, Nadir Durrani, and Fahim Dalvi.
\newblock \href{https://direct.mit.edu/tacl/article/doi/10.1162/tacl_a_00519/113852/Neuron-level-Interpretation-of-Deep-NLP-Models-A}{Neuron-level interpretation of deep nlp models: A survey}.
\newblock \emph{TACL}, November 2022.

\bibitem[Sakarvadia et~al.(2023{\natexlab{a}})Sakarvadia, Ajith, Khan, Grzenda, Hudson, Bauer, Chard, and Foster]{sakarvadia_memory_2023}
Mansi Sakarvadia, Aswathy Ajith, Arham Khan, Daniel Grzenda, Nathaniel Hudson, Andr{\'e} Bauer, Kyle Chard, and Ian Foster.
\newblock \href{http://arxiv.org/abs/2309.05605}{Memory injections: Correcting multi-hop reasoning failures during inference in transformer-based language models}.
\newblock \emph{CoRR}, September 2023{\natexlab{a}}.

\bibitem[Sakarvadia et~al.(2023{\natexlab{b}})Sakarvadia, Khan, Ajith, Grzenda, Hudson, Bauer, Chard, and Foster]{sakarvadia_attention_2023}
Mansi Sakarvadia, Arham Khan, Aswathy Ajith, Daniel Grzenda, Nathaniel Hudson, Andr{\'e} Bauer, Kyle Chard, and Ian Foster.
\newblock \href{http://arxiv.org/abs/2310.16270}{Attention lens: A tool for mechanistically interpreting the attention head information retrieval mechanism}.
\newblock \emph{CoRR}, October 2023{\natexlab{b}}.

\bibitem[Salin et~al.(2022)Salin, Farah, Ayache, and Favre]{salin_are_2022}
Emmanuelle Salin, Badreddine Farah, St{\'e}phane Ayache, and Benoit Favre.
\newblock \href{https://ojs.aaai.org/index.php/AAAI/article/view/21375}{Are vision-language transformers learning multimodal representations? a probing perspective}.
\newblock \emph{AAAI}, June 2022.

\bibitem[Salvatori et~al.(2023)Salvatori, Mali, Buckley, Lukasiewicz, Rao, Friston, and Ororbia]{salvatori_braininspired_2023}
Tommaso Salvatori, Ankur Mali, Christopher~L. Buckley, Thomas Lukasiewicz, Rajesh P.~N. Rao, Karl Friston, and Alexander Ororbia.
\newblock \href{https://arxiv.org/abs/2308.07870}{Brain-inspired computational intelligence via predictive coding}.
\newblock \emph{CoRR}, 2023.

\bibitem[Saphra(2023)]{saphra_interpretability_2023}
Naomi Saphra.
\newblock \href{https://thegradient.pub/interpretability-creationism}{Interpretability creationism}.
\newblock \emph{The Gradient}, 2023.

\bibitem[Schaeffer et~al.(2023)Schaeffer, Miranda, and Koyejo]{schaeffer_are_2023}
Rylan Schaeffer, Brando Miranda, and Sanmi Koyejo.
\newblock \href{http://arxiv.org/abs/2304.15004}{Are emergent abilities of large language models a mirage?}
\newblock \emph{CoRR}, May 2023.

\bibitem[Scherlis et~al.(2023)Scherlis, Sachan, Jermyn, Benton, and Shlegeris]{scherlis_polysemanticity_2023}
Adam Scherlis, Kshitij Sachan, Adam~S. Jermyn, Joe Benton, and Buck Shlegeris.
\newblock \href{http://arxiv.org/abs/2210.01892}{Polysemanticity and capacity in neural networks}.
\newblock \emph{CoRR}, July 2023.

\bibitem[Sch{\"o}lkopf et~al.(2021)Sch{\"o}lkopf, Locatello, Bauer, Ke, Kalchbrenner, Goyal, and Bengio]{scholkopf_causal_2021}
Bernhard Sch{\"o}lkopf, Francesco Locatello, Stefan Bauer, Nan~Rosemary Ke, Nal Kalchbrenner, Anirudh Goyal, and Yoshua Bengio.
\newblock \href{http://arxiv.org/abs/2102.11107}{Towards causal representation learning}.
\newblock \emph{Special Issue of Proceedings of the IEEE - Advances in Machine Learning and Deep Neural Networks}, February 2021.

\bibitem[Schuster et~al.(2022)Schuster, Fisch, Gupta, Dehghani, Bahri, Tran, Tay, and Metzler]{schuster_confident_2022}
Tal Schuster, Adam Fisch, Jai Gupta, Mostafa Dehghani, Dara Bahri, Vinh~Q. Tran, Yi~Tay, and Donald Metzler.
\newblock \href{http://arxiv.org/abs/2207.07061}{Confident adaptive language modeling}.
\newblock \emph{NeurIPS Oral}, October 2022.

\bibitem[Selvaraju et~al.(2016)Selvaraju, Das, Vedantam, Cogswell, Parikh, and Batra]{selvaraju_gradcam_2016}
Ramprasaath~R. Selvaraju, Abhishek Das, Ramakrishna Vedantam, Michael Cogswell, Devi Parikh, and Dhruv Batra.
\newblock \href{https://arxiv.org/abs/1610.02391}{Grad-cam: Why did you say that? visual explanations from deep networks via gradient-based localization}.
\newblock \emph{ICCV}, 2016.

\bibitem[Shanahan et~al.(2023)Shanahan, McDonell, and Reynolds]{shanahan_role_2023}
Murray Shanahan, Kyle McDonell, and Laria Reynolds.
\newblock \href{https://www.nature.com/articles/s41586-023-06647-8}{Role play with large language models}.
\newblock \emph{Nature}, November 2023.

\bibitem[Shapley(1988)]{shapley_value_1988}
Lloyd~S. Shapley.
\newblock \href{https://www.cambridge.org/core/product/identifier/CBO9780511528446A008/type/book_part}{A value for {\emph{n}} -person games}.
\newblock \emph{Cambridge University Press}, October 1988.

\bibitem[Sharkey(2022)]{sharkey_circumventing_2022}
Lee Sharkey.
\newblock \href{https://arxiv.org/abs/2212.11415}{Circumventing interpretability: How to defeat mind-readers}.
\newblock \emph{CoRR}, December 2022.

\bibitem[Sharkey(2023)]{sharkey_technical_2023}
Lee Sharkey.
\newblock \href{https://arxiv.org/abs/2305.03452}{A technical note on bilinear layers for interpretability}.
\newblock \emph{CoRR}, May 2023.

\bibitem[Sharkey et~al.(2022{\natexlab{a}})Sharkey, Black, and {beren}]{sharkey_current_2022}
Lee Sharkey, Sid Black, and {beren}.
\newblock \href{https://www.alignmentforum.org/posts/Jgs7LQwmvErxR9BCC/current-themes-in-mechanistic-interpretability-research}{Current themes in mechanistic interpretability research}.
\newblock \emph{AI Alignment Forum}, November 2022{\natexlab{a}}.

\bibitem[Sharkey et~al.(2022{\natexlab{b}})Sharkey, Braun, and Millidge]{sharkey_taking_2022}
Lee Sharkey, Dan Braun, and Beren Millidge.
\newblock \href{https://www.alignmentforum.org/posts/z6QQJbtpkEAX3Aojj/interim-research-report-taking-features-out-of-superposition}{Taking features out of superposition with sparse autoencoders}.
\newblock \emph{AI Alignment Forum}, 2022{\natexlab{b}}.

\bibitem[Sharma et~al.(2023)Sharma, Tong, Korbak, Duvenaud, Askell, Bowman, Cheng, Durmus, {Hatfield-Dodds}, Johnston, Kravec, Maxwell, McCandlish, Ndousse, Rausch, Schiefer, Yan, Zhang, and Perez]{sharma_understanding_2023}
Mrinank Sharma, Meg Tong, Tomasz Korbak, David Duvenaud, Amanda Askell, Samuel~R. Bowman, Newton Cheng, Esin Durmus, Zac {Hatfield-Dodds}, Scott~R. Johnston, Shauna Kravec, Timothy Maxwell, Sam McCandlish, Kamal Ndousse, Oliver Rausch, Nicholas Schiefer, Da~Yan, Miranda Zhang, and Ethan Perez.
\newblock \href{http://arxiv.org/abs/2310.13548}{Towards understanding sycophancy in language models}.
\newblock \emph{CoRR}, October 2023.

\bibitem[Shovelain \& McKernon(2023)Shovelain and McKernon]{shovelain_riskreward_2023}
Justin Shovelain and Elliot McKernon.
\newblock \href{https://www.lesswrong.com/posts/HdqdqNC3MyABHzSqf/the-risk-reward-tradeoff-of-interpretability-research}{The risk-reward tradeoff of interpretability research}.
\newblock \emph{LessWrong}, July 2023.

\bibitem[Shrikumar et~al.(2017)Shrikumar, Greenside, and Kundaje]{shrikumar_learning_2017}
Avanti Shrikumar, Peyton Greenside, and Anshul Kundaje.
\newblock \href{http://arxiv.org/abs/1704.02685}{Learning important features through propagating activation differences}.
\newblock \emph{ICML}, 2017.

\bibitem[Simon et~al.(2023)Simon, Knutins, Ziyin, Geisz, Fetterman, and Albrecht]{simon_stepwise_2023}
James~B. Simon, Maksis Knutins, Liu Ziyin, Daniel Geisz, Abraham~J. Fetterman, and Joshua Albrecht.
\newblock \href{http://arxiv.org/abs/2303.15438}{On the stepwise nature of self-supervised learning}.
\newblock \emph{ICML}, May 2023.

\bibitem[{slavachalnev}(2024)]{slavachalnev_sparse_2024}
{slavachalnev}.
\newblock \href{https://www.lesswrong.com/posts/MXabwqMwo3rkGqEW8/sparse-mlp-distillation}{Sparse mlp distillation}.
\newblock \emph{LessWrong}, 2024.

\bibitem[Smilkov et~al.(2017)Smilkov, Thorat, Kim, Vi{\'e}gas, and Wattenberg]{smilkov_smoothgrad_2017}
Daniel Smilkov, Nikhil Thorat, Been Kim, Fernanda Vi{\'e}gas, and Martin Wattenberg.
\newblock \href{http://arxiv.org/abs/1706.03825}{Smoothgrad: removing noise by adding noise}.
\newblock \emph{CoRR}, June 2017.

\bibitem[Soares(2023)]{soares_if_2023}
Nate Soares.
\newblock \href{https://www.lesswrong.com/posts/BinkknLBYxskMXuME/if-interpretability-research-goes-well-it-may-get-dangerous}{If interpretability research goes well, it may get dangerous}.
\newblock \emph{LessWrong}, April 2023.

\bibitem[Srivastava et~al.(2014)Srivastava, Hinton, Krizhevsky, Sutskever, and Salakhutdinov]{srivastava_dropout_2014}
Nitish Srivastava, Geoffrey Hinton, Alex Krizhevsky, Ilya Sutskever, and Ruslan Salakhutdinov.
\newblock \href{http://jmlr.org/papers/v15/srivastava14a.html}{Dropout: A simple way to prevent neural networks from overfitting}.
\newblock \emph{JMLR}, 2014.

\bibitem[Stander et~al.(2023)Stander, Yu, Fan, and Biderman]{stander_grokking_2023}
Dashiell Stander, Qinan Yu, Honglu Fan, and Stella Biderman.
\newblock \href{https://arxiv.org/abs/2312.06581}{Grokking group multiplication with cosets}.
\newblock \emph{CoRR}, 2023.

\bibitem[Steinhardt(2023)]{jacob_emergent_2023}
Jacob Steinhardt.
\newblock \href{https://bounded-regret.ghost.io/emergent-deception-optimization/}{Emergent deception and emergent optimization}.
\newblock \emph{Bounded Regret}, February 2023.

\bibitem[Stolfo et~al.(2023)Stolfo, Belinkov, and Sachan]{stolfo_mechanistic_2023}
Alessandro Stolfo, Yonatan Belinkov, and Mrinmaya Sachan.
\newblock \href{http://arxiv.org/abs/2305.15054}{A mechanistic interpretation of arithmetic reasoning in language models using causal mediation analysis}.
\newblock \emph{EMNLP}, October 2023.

\bibitem[Sucholutsky et~al.(2023)Sucholutsky, Muttenthaler, Weller, Peng, Bobu, Kim, Love, Grant, Groen, Achterberg, Tenenbaum, Collins, Hermann, Oktar, Greff, Hebart, Jacoby, Zhang, Marjieh, Geirhos, Chen, Kornblith, Rane, Konkle, O'Connell, Unterthiner, Lampinen, M{\"u}ller, Toneva, and Griffiths]{sucholutsky_getting_2023}
Ilia Sucholutsky, Lukas Muttenthaler, Adrian Weller, Andi Peng, Andreea Bobu, Been Kim, Bradley~C. Love, Erin Grant, Iris Groen, Jascha Achterberg, Joshua~B. Tenenbaum, Katherine~M. Collins, Katherine~L. Hermann, Kerem Oktar, Klaus Greff, Martin~N. Hebart, Nori Jacoby, Qiuyi Zhang, Raja Marjieh, Robert Geirhos, Sherol Chen, Simon Kornblith, Sunayana Rane, Talia Konkle, Thomas~P. O'Connell, Thomas Unterthiner, Andrew~K. Lampinen, Klaus-Robert M{\"u}ller, Mariya Toneva, and Thomas~L. Griffiths.
\newblock \href{http://arxiv.org/abs/2310.13018}{Getting aligned on representational alignment}.
\newblock \emph{CoRR}, November 2023.

\bibitem[Sun et~al.(2024)Sun, Miller, Zhmoginov, Vladymyrov, and Sandler]{sun_learning_2024}
Chen Sun, Nolan~Andrew Miller, Andrey Zhmoginov, Max Vladymyrov, and Mark Sandler.
\newblock \href{https://openreview.net/forum?id=R5Q5lANcjY}{Learning and unlearning of fabricated knowledge in language models}.
\newblock \emph{ICML MI Workshop}, June 2024.

\bibitem[Sundararajan et~al.(2017)Sundararajan, Taly, and Yan]{sundararajan_axiomatic_2017}
Mukund Sundararajan, Ankur Taly, and Qiqi Yan.
\newblock \href{http://arxiv.org/abs/1703.01365}{Axiomatic attribution for deep networks}.
\newblock \emph{ICML}, June 2017.

\bibitem[Syed et~al.(2023)Syed, Rager, and Conmy]{syed_attribution_2023}
Aaquib Syed, Can Rager, and Arthur Conmy.
\newblock \href{http://arxiv.org/abs/2310.10348}{Attribution patching outperforms automated circuit discovery}.
\newblock \emph{CoRR}, October 2023.

\bibitem[{technicalities} \& Stag(2023){technicalities} and Stag]{technicalities_shallow_2023}
{technicalities} and Stag.
\newblock \href{https://www.lesswrong.com/posts/zaaGsFBeDTpCsYHef/shallow-review-of-live-agendas-in-alignment-and-safety}{Shallow review of live agendas in alignment \& safety}.
\newblock \emph{LessWrong}, 2023.

\bibitem[Tegmark \& Omohundro(2023)Tegmark and Omohundro]{tegmark_provably_2023}
Max Tegmark and Steve Omohundro.
\newblock \href{http://arxiv.org/abs/2309.01933}{Provably safe systems: the only path to controllable agi}.
\newblock \emph{CoRR}, September 2023.

\bibitem[Templeton et~al.(2024)Templeton, Conerly, Marcus, Lindsey, Bricken, and Chen]{templeton_scaling_2024}
Adly Templeton, Tom Conerly, Jonathan Marcus, Jack Lindsey, Trenton Bricken, and Brian Chen.
\newblock \href{https://transformer-circuits.pub/2024/scaling-monosemanticity/index.html}{Scaling monosemanticity: Extracting interpretable features from claude 3 sonnet}.
\newblock \emph{Transformer Circuits Thread}, 2024.

\bibitem[Tenney et~al.(2019)Tenney, Das, and Pavlick]{tenney_bert_2019}
Ian Tenney, Dipanjan Das, and Ellie Pavlick.
\newblock \href{http://arxiv.org/abs/1905.05950}{Bert rediscovers the classical nlp pipeline}.
\newblock \emph{ACL}, August 2019.

\bibitem[Thilak et~al.(2022)Thilak, Littwin, Zhai, Saremi, Paiss, and Susskind]{thilak_slingshot_2022}
Vimal Thilak, Etai Littwin, Shuangfei Zhai, Omid Saremi, Roni Paiss, and Joshua Susskind.
\newblock \href{https://arxiv.org/abs/2206.04817}{The slingshot mechanism: An empirical study of adaptive optimizers and the grokking phenomenon}.
\newblock \emph{CoRR}, 2022.

\bibitem[Thurnherr \& Scheurer(2024)Thurnherr and Scheurer]{thurnherr_tracrbench_2024}
Hannes Thurnherr and J{\'e}r{\'e}my Scheurer.
\newblock \href{https://openreview.net/forum?id=vNubZ5zK8h}{Tracrbench: Generating interpretability testbeds with large language models}.
\newblock \emph{ICML MI Workshop}, June 2024.

\bibitem[Tigges et~al.(2024)Tigges, Hollinsworth, Geiger, and Nanda]{tigges_language_2024}
Curt Tigges, Oskar~John Hollinsworth, Atticus Geiger, and Neel Nanda.
\newblock \href{https://openreview.net/forum?id=Xsf6dOOMMc}{Language models linearly represent sentiment}.
\newblock \emph{ICML MI Workshop}, June 2024.

\bibitem[Todd et~al.(2023)Todd, Li, Sharma, Mueller, Wallace, and Bau]{todd_function_2023}
Eric Todd, Millicent~L. Li, Arnab~Sen Sharma, Aaron Mueller, Byron~C. Wallace, and David Bau.
\newblock \href{https://arxiv.org/abs/2310.15213}{Function vectors in large language models}.
\newblock \emph{CoRR}, 2023.

\bibitem[Trivedi et~al.(2021)Trivedi, Zhang, Sun, and Lim]{trivedi_learning_2021}
Dweep Trivedi, Jesse Zhang, Shao-Hua Sun, and Joseph~J. Lim.
\newblock \href{http://arxiv.org/abs/2108.13643}{Learning to synthesize programs as interpretable and generalizable policies}.
\newblock \emph{NeurIPS}, 2021.

\bibitem[Turner et~al.(2023)Turner, Thiergart, Udell, Leech, Mini, and MacDiarmid]{turner_activation_2023}
Alexander~Matt Turner, Lisa Thiergart, David Udell, Gavin Leech, Ulisse Mini, and Monte MacDiarmid.
\newblock \href{http://arxiv.org/abs/2308.10248}{Activation addition: Steering language models without optimization}.
\newblock \emph{CoRR}, September 2023.

\bibitem[Vaintrob et~al.(2024)Vaintrob, {jake\_mendel}, and Kaarel]{vaintrob_mathematical_2024}
Dmitry Vaintrob, {jake\_mendel}, and Kaarel.
\newblock \href{https://www.lesswrong.com/posts/2roZtSr5TGmLjXMnT/toward-a-mathematical-framework-for-computation-in}{Toward a mathematical framework for computation in superposition}.
\newblock \emph{AI Alignment Forum}, 2024.

\bibitem[Variengien \& Winsor(2023)Variengien and Winsor]{variengien_look_2023}
Alexandre Variengien and Eric Winsor.
\newblock \href{http://arxiv.org/abs/2312.10091}{Look before you leap: A universal emergent decomposition of retrieval tasks in language models}.
\newblock \emph{ICML MI Workshop}, December 2023.

\bibitem[Varma et~al.(2023)Varma, Shah, Kenton, Kram{\'a}r, and Kumar]{varma_explaining_2023}
Vikrant Varma, Rohin Shah, Zachary Kenton, J{\'a}nos Kram{\'a}r, and Ramana Kumar.
\newblock \href{http://arxiv.org/abs/2309.02390}{Explaining grokking through circuit efficiency}.
\newblock \emph{CoRR}, September 2023.

\bibitem[Verma et~al.(2019{\natexlab{a}})Verma, Le, Yue, and Chaudhuri]{verma_imitation-projected_2019}
Abhinav Verma, Hoang~M. Le, Yisong Yue, and Swarat Chaudhuri.
\newblock \href{http://arxiv.org/abs/1907.05431}{Imitation-projected programmatic reinforcement learning}.
\newblock \emph{NeurIPS}, 2019{\natexlab{a}}.

\bibitem[Verma et~al.(2019{\natexlab{b}})Verma, Murali, Singh, Kohli, and Chaudhuri]{verma_programmatically_2019}
Abhinav Verma, Vijayaraghavan Murali, Rishabh Singh, Pushmeet Kohli, and Swarat Chaudhuri.
\newblock \href{http://arxiv.org/abs/1804.02477}{Programmatically interpretable reinforcement learning}.
\newblock \emph{CoRR}, April 2019{\natexlab{b}}.

\bibitem[Vig et~al.(2020)Vig, Gehrmann, Belinkov, Qian, Nevo, Singer, and Shieber]{vig_investigating_2020}
Jesse Vig, Sebastian Gehrmann, Yonatan Belinkov, Sharon Qian, Daniel Nevo, Yaron Singer, and Stuart Shieber.
\newblock \href{https://proceedings.neurips.cc/paper/2020/hash/92650b2e92217715fe312e6fa7b90d82-Abstract.html}{Investigating gender bias in language models using causal mediation analysis}.
\newblock \emph{NeurIPS}, 2020.

\bibitem[Vilas et~al.(2023)Vilas, Schauml{\"o}ffel, and Roig]{vilas_analyzing_2023}
M.~Vilas, Timothy Schauml{\"o}ffel, and Gemma Roig.
\newblock \href{https://www.semanticscholar.org/paper/7c57530318ae6c6e49c835f23e75affd9cf0827d}{Analyzing vision transformers for image classification in class embedding space}.
\newblock \emph{CoRR}, 2023.

\bibitem[Voita \& Titov(2020)Voita and Titov]{voita_informationtheoretic_2020}
Elena Voita and Ivan Titov.
\newblock \href{http://arxiv.org/abs/2003.12298}{Information-theoretic probing with minimum description length}.
\newblock \emph{EMNLP}, March 2020.

\bibitem[Voita et~al.(2023)Voita, Ferrando, and Nalmpantis]{voita_neurons_2023}
Elena Voita, Javier Ferrando, and Christoforos Nalmpantis.
\newblock \href{http://arxiv.org/abs/2309.04827}{Neurons in large language models: Dead, n-gram, positional}.
\newblock \emph{CoRR}, September 2023.

\bibitem[von K{\"u}gelgen et~al.(2019)von K{\"u}gelgen, Loog, Mey, and Scholkopf]{kugelgen_semisupervised_2019}
Julius von K{\"u}gelgen, M.~Loog, A.~Mey, and B.~Scholkopf.
\newblock \href{https://arxiv.org/abs/1905.12081}{Semi-supervised learning, causality, and the conditional cluster assumption}.
\newblock \emph{Conference on Uncertainty in Artificial Intelligence}, May 2019.

\bibitem[{von Oswald} et~al.(2023){von Oswald}, Niklasson, Schlegel, Kobayashi, Zucchet, Scherrer, Miller, Sandler, y~Arcas, Vladymyrov, Pascanu, and Sacramento]{vonoswald_uncovering_2023}
Johannes {von Oswald}, Eyvind Niklasson, Maximilian Schlegel, Seijin Kobayashi, Nicolas Zucchet, Nino Scherrer, Nolan Miller, Mark Sandler, Blaise~Ag{\"u}era y~Arcas, Max Vladymyrov, Razvan Pascanu, and Jo{\~a}o Sacramento.
\newblock \href{http://arxiv.org/abs/2309.05858}{Uncovering mesa-optimization algorithms in transformers}.
\newblock \emph{CoRR}, September 2023.

\bibitem[Voss et~al.(2021)Voss, Goh, Cammarata, Petrov, Schubert, and Olah]{voss_branch_2021}
Chelsea Voss, Gabriel Goh, Nick Cammarata, Michael Petrov, Ludwig Schubert, and Chris Olah.
\newblock \href{https://distill.pub/2020/circuits/branch-specialization}{Branch specialization}.
\newblock \emph{Distill}, April 2021.

\bibitem[Wang et~al.(2024)Wang, Yue, Su, and Sun]{wang_grokked_2024}
Boshi Wang, Xiang Yue, Yu~Su, and Huan Sun.
\newblock \href{https://openreview.net/forum?id=ns8IH5Sn5y}{Grokked transformers are implicit reasoners: A mechanistic journey to the edge of generalization}.
\newblock \emph{ICML MI Workshop}, June 2024.

\bibitem[Wang et~al.(2023)Wang, Variengien, Conmy, Shlegeris, and Steinhardt]{wang_interpretability_2023}
Kevin Wang, Alexandre Variengien, Arthur Conmy, Buck Shlegeris, and Jacob Steinhardt.
\newblock \href{http://arxiv.org/abs/2211.00593}{Interpretability in the wild: a circuit for indirect object identification in gpt-2 small}.
\newblock \emph{ICLR}, 2023.

\bibitem[Warstadt et~al.(2020)Warstadt, Parrish, Liu, Mohananey, Peng, Wang, and Bowman]{warstadt_blimp_2020}
Alex Warstadt, Alicia Parrish, Haokun Liu, Anhad Mohananey, Wei Peng, Sheng-Fu Wang, and Samuel~R. Bowman.
\newblock \href{https://aclanthology.org/2020.tacl-1.25}{Blimp: The benchmark of linguistic minimal pairs for english}.
\newblock \emph{Transactions of the Association for Computational Linguistics}, 2020.

\bibitem[Watanabe(2009)]{watanabe_algebraic_2009}
Sumio Watanabe.
\newblock \href{https://www.cambridge.org/core/books/algebraic-geometry-and-statistical-learning-theory/9C8FD1BDC817E2FC79117C7F41544A3A}{\emph{Algebraic Geometry and Statistical Learning Theory}}.
\newblock Cambridge Monographs on Applied and Computational Mathematics. Cambridge University Press, 2009.

\bibitem[Watanabe(2018)]{watanabe_mathematical_2018}
Sumio Watanabe.
\newblock \href{https://www.taylorfrancis.com/books/9781482238082}{\emph{Mathematical Theory of Bayesian Statistics}}.
\newblock {Chapman and Hall}, 1 edition, April 2018.

\bibitem[Wei et~al.(2022)Wei, Tay, Bommasani, Raffel, Zoph, Borgeaud, Yogatama, Bosma, Zhou, Metzler, Chi, Hashimoto, Vinyals, Liang, Dean, and Fedus]{wei_emergent_2022}
Jason Wei, Yi~Tay, Rishi Bommasani, Colin Raffel, Barret Zoph, Sebastian Borgeaud, Dani Yogatama, Maarten Bosma, Denny Zhou, Donald Metzler, Ed~H. Chi, Tatsunori Hashimoto, Oriol Vinyals, Percy Liang, Jeff Dean, and William Fedus.
\newblock \href{http://arxiv.org/abs/2206.07682}{Emergent abilities of large language models}.
\newblock \emph{TMLR}, October 2022.

\bibitem[Wentworth(2022)]{wentworth_how_2022}
John Wentworth.
\newblock \href{https://www.alignmentforum.org/posts/w4aeAFzSAguvqA5qu/how-to-go-from-interpretability-to-alignment-just-retarget}{How to go from interpretability to alignment: Just retarget the search}.
\newblock \emph{AI Alignment Forum}, August 2022.

\bibitem[Whittington et~al.(2022)Whittington, Dorrell, Ganguli, and Behrens]{whittington_disentangling_2022}
James C.~R. Whittington, Will Dorrell, Surya Ganguli, and Timothy E.~J. Behrens.
\newblock \href{http://arxiv.org/abs/2210.01768}{Disentangling with biological constraints: A theory of functional cell types}.
\newblock \emph{CoRR}, September 2022.

\bibitem[Wu et~al.(2022{\natexlab{a}})Wu, Chen, Zhang, Zhu, Wei, Yuan, and Shen]{wu_backdoorbench_2022}
Baoyuan Wu, Hongrui Chen, Mingda Zhang, Zihao Zhu, Shaokui Wei, Danni Yuan, and Chao Shen.
\newblock \href{http://arxiv.org/abs/2206.12654}{Backdoorbench: A comprehensive benchmark of backdoor learning}.
\newblock \emph{NeurIPS Datasets and Benchmarks}, October 2022{\natexlab{a}}.

\bibitem[Wu et~al.(2022{\natexlab{b}})Wu, Geiger, Rozner, Kreiss, Lu, Icard, Potts, and Goodman]{wu_causal_2022}
Zhengxuan Wu, Atticus Geiger, Joshua Rozner, Elisa Kreiss, Hanson Lu, Thomas Icard, Christopher Potts, and Noah Goodman.
\newblock \href{https://aclanthology.org/2022.naacl-main.318}{Causal distillation for language models}.
\newblock \emph{NAACL-HLT}, July 2022{\natexlab{b}}.

\bibitem[Wu et~al.(2023{\natexlab{a}})Wu, D'Oosterlinck, Geiger, Zur, and Potts]{wu_causal_2023}
Zhengxuan Wu, Karel D'Oosterlinck, Atticus Geiger, Amir Zur, and Christopher Potts.
\newblock \href{http://arxiv.org/abs/2209.14279}{Causal proxy models for concept-based model explanations}.
\newblock \emph{ICML}, 2023{\natexlab{a}}.

\bibitem[Wu et~al.(2023{\natexlab{b}})Wu, Geiger, Potts, and Goodman]{wu_interpretability_2023}
Zhengxuan Wu, Atticus Geiger, Christopher Potts, and Noah~D. Goodman.
\newblock \href{http://arxiv.org/abs/2305.08809}{Interpretability at scale: Identifying causal mechanisms in alpaca}.
\newblock \emph{CoRR}, May 2023{\natexlab{b}}.

\bibitem[Wu et~al.(2024)Wu, Arora, Wang, Geiger, Jurafsky, Manning, and Potts]{wu_reft_2024}
Zhengxuan Wu, Aryaman Arora, Zheng Wang, Atticus Geiger, Dan Jurafsky, Christopher~D. Manning, and Christopher Potts.
\newblock \href{http://arxiv.org/abs/2404.03592}{Reft: Representation finetuning for language models}.
\newblock \emph{CoRR}, May 2024.

\bibitem[Yu et~al.(2023)Yu, Merullo, and Pavlick]{yu_characterizing_2023}
Qinan Yu, Jack Merullo, and Ellie Pavlick.
\newblock \href{http://arxiv.org/abs/2310.15910}{Characterizing mechanisms for factual recall in language models}.
\newblock \emph{CoRR}, October 2023.

\bibitem[Zeiler \& Fergus(2014)Zeiler and Fergus]{zeiler_visualizing_2014}
Matthew~D. Zeiler and Rob Fergus.
\newblock \href{http://arxiv.org/abs/1311.2901}{Visualizing and understanding convolutional networks}.
\newblock \emph{ECCV}, 2014.

\bibitem[Zhang et~al.(2021)Zhang, Titov, and Sennrich]{zhang_sparse_2021}
Biao Zhang, Ivan Titov, and Rico Sennrich.
\newblock \href{http://arxiv.org/abs/2104.07012}{Sparse attention with linear units}.
\newblock \emph{EMNLP}, October 2021.

\bibitem[Zhang et~al.(2017)Zhang, Bengio, Hardt, Recht, and Vinyals]{zhang_understanding_2017}
Chiyuan Zhang, Samy Bengio, Moritz Hardt, Benjamin Recht, and Oriol Vinyals.
\newblock \href{http://arxiv.org/abs/1611.03530}{Understanding deep learning requires rethinking generalization}.
\newblock \emph{ICLR}, February 2017.

\bibitem[Zhang \& Nanda(2023)Zhang and Nanda]{zhang_best_2023}
Fred Zhang and Neel Nanda.
\newblock \href{https://arxiv.org/abs/2309.16042}{Towards best practices of activation patching in language models: Metrics and methods}.
\newblock \emph{ICLR}, 2023.

\bibitem[Zhang et~al.(2019)Zhang, Yang, Ma, and Wu]{zhang_interpreting_2019}
Quanshi Zhang, Yu~Yang, Haotian Ma, and Ying~Nian Wu.
\newblock \href{https://arxiv.org/abs/1802.00121}{Interpreting cnns via decision trees}.
\newblock \emph{CVPR}, 2019.

\bibitem[Zhong et~al.(2023)Zhong, Liu, Tegmark, and Andreas]{zhong_clock_2023}
Ziqian Zhong, Ziming Liu, Max Tegmark, and Jacob Andreas.
\newblock \href{https://arxiv.org/abs/2306.17844}{The clock and the pizza: Two stories in mechanistic explanation of neural networks}.
\newblock \emph{CoRR}, 2023.

\bibitem[Zimmermann et~al.(2021)Zimmermann, Borowski, Geirhos, Bethge, Wallis, and Brendel]{zimmermann_how_2021}
Roland~S. Zimmermann, Judy Borowski, Robert Geirhos, Matthias Bethge, Thomas S.~A. Wallis, and Wieland Brendel.
\newblock \href{http://arxiv.org/abs/2106.12447}{How well do feature visualizations support causal understanding of cnn activations?}
\newblock \emph{NeurIPS}, November 2021.

\bibitem[Zou et~al.(2023)Zou, Phan, Chen, Campbell, Guo, Ren, Pan, Yin, Mazeika, Dombrowski, Goel, Li, Byun, Wang, Mallen, Basart, Koyejo, Song, Fredrikson, Kolter, and Hendrycks]{zou_representation_2023}
Andy Zou, Long Phan, Sarah Chen, James Campbell, Phillip Guo, Richard Ren, Alexander Pan, Xuwang Yin, Mantas Mazeika, Ann-Kathrin Dombrowski, Shashwat Goel, Nathaniel Li, Michael~J. Byun, Zifan Wang, Alex Mallen, Steven Basart, Sanmi Koyejo, Dawn Song, Matt Fredrikson, J.~Zico Kolter, and Dan Hendrycks.
\newblock \href{http://arxiv.org/abs/2310.01405}{Representation engineering: A top-down approach to ai transparency}.
\newblock \emph{CoRR}, October 2023.

\end{thebibliography}
\bibliographystyle{tmlr/mybibstyle}

\end{document}